\documentclass[11pt]{article}
\ifdefined\pdftexversion\pdfoutput=1\fi
\usepackage[utf8]{inputenc}
\usepackage[T1]{fontenc}
\usepackage{lmodern}
\usepackage[a4paper,margin=25mm]{geometry}
\usepackage{microtype}
\usepackage{graphicx}
\usepackage{amsmath,amssymb}
\usepackage{booktabs,longtable,array}
\usepackage{pdflscape}
\usepackage{caption}
\usepackage{hyperref}
\usepackage[numbers,sort&compress]{natbib}
\usepackage{enumitem}
\setlist{nosep,leftmargin=*}
\hypersetup{colorlinks=true,linkcolor=black,citecolor=black,urlcolor=blue,pdfborder={0 0 0}}
\begin{document}

\begin{center}
{\LARGE\bfseries An auditable conditional-strategy framework for open-ended decision-making in complex lung cancer\par}
\vspace{1.2em}
{\small Daoyun Wang\textsuperscript{1\#}, Zhicheng Huang\textsuperscript{1\#}, Huaiyuan Sun\textsuperscript{2\#}, Jiaqi Xu\textsuperscript{3\#}, Xiaowei Xu\textsuperscript{4\#}, Zhibo Zheng\textsuperscript{1}, Zhongxing Bing\textsuperscript{1}, Yuxiao Lin\textsuperscript{1}, Yicheng Liang\textsuperscript{1}, Chao Gao\textsuperscript{1}, Bowen Xue\textsuperscript{1}, Kai Zhang\textsuperscript{1}, Song Xu\textsuperscript{5,6}, Wanpu Yan\textsuperscript{7}, Hui Xia\textsuperscript{8}, Lin Li\textsuperscript{9}, Xiang Yan\textsuperscript{10}, Mu Hu\textsuperscript{11}, Qianli Ma\textsuperscript{12}, Zhiqiang Xue\textsuperscript{13}, Xiaofang Liu\textsuperscript{14}, Zhihai Han\textsuperscript{15}, Nan Zhang\textsuperscript{16}, Chuanhao Tang\textsuperscript{17}, Tongmei Zhang\textsuperscript{18}, Lan Song\textsuperscript{19}, Chaohui Zhu\textsuperscript{20}, Xuan Zeng\textsuperscript{21}, Shafei Wu\textsuperscript{21}, Hui Guan\textsuperscript{22}, Lei Deng\textsuperscript{23}, Huaxia Yang\textsuperscript{24}, Zeliang Lian\textsuperscript{2}, Wubin Sun\textsuperscript{2}, Yongxin Wang\textsuperscript{2}, Xiaohui Shen\textsuperscript{2}, Binlin Wang\textsuperscript{2}, Tiantian Gu\textsuperscript{2}, Yu Cui\textsuperscript{2}, Li Zhang\textsuperscript{25*}, Shirui Wang\textsuperscript{2*}, Naixin Liang\textsuperscript{1*}\par}
\end{center}
\vspace{0.4em}
{\small\centering\interlinepenalty=10000
\textsuperscript{1} Department of Thoracic Surgery, Peking Union Medical College Hospital, Chinese Academy of Medical Sciences and Peking Union Medical College, Beijing, China.

\textsuperscript{2} Medlinker Intelligent and Digital Technology Co. Ltd., Beijing, China.

\textsuperscript{3} Department of Rheumatology and Clinical Immunology, National Clinical Research Center for Dermatologic and Immunologic Diseases, the Ministry of Education Key Laboratory, Peking Union Medical College Hospital, Chinese Academy of Medical Sciences and Peking Union Medical College.

\textsuperscript{4} Institute of Intelligent Medicine, Chinese Academy of Medical Sciences.

\textsuperscript{5} Department of Lung Cancer Surgery, Tianjin Medical University General Hospital, Tianjin, China.

\textsuperscript{6} Tianjin Key Laboratory of Lung Cancer Metastasis and Tumor Microenvironment, Lung Cancer Institute, Tianjin Medical University General Hospital, Tianjin, China.

\textsuperscript{7} Key laboratory of Carcinogenesis and Translational Research (Ministry of Education), Department of Thoracic Surgery I, Peking University Cancer Hospital \& Institute, Beijing, China.

\textsuperscript{8} Department of Thoracic Surgery, the Fourth Medical Center of PLA General Hospital, Beijing, China.

\textsuperscript{9} Oncology Department, Beijing Hospital, National Center for Gerontology; National Clinical Research Center for Gerontology; The Key Laboratory of Geriatrics of NHC; Institute of Geriatric Medicine, Chinese Academy of Medical Sciences.

\textsuperscript{10} Department of Thoracic Surgery, Peking University People's Hospital, Beijing, China.

\textsuperscript{11} Department of Thoracic Surgery, Beijing Friendship Hospital, Capital Medical University, Beijing, China.

\textsuperscript{12} Department of Thoracic Surgery, China-Japan Friendship Hospital, Beijing, China.

\textsuperscript{13} Department of Thoracic Surgery, The First Medical Center of Chinese PLA General Hospital, Beijing, China.

\textsuperscript{14} Department of Pulmonary and Critical Care Medicine, Beijing Tongren Hospital, Capital Medical University, Beijing, China.

\textsuperscript{15} Department of Pulmonary and Critical Care Medicine, the Sixth Medical Center of PLA General Hospital, Beijing, China.

\textsuperscript{16} Department of Pulmonary and Critical Care Medicine 2, Emergency General Hospital, Beijing, China.

\textsuperscript{17} Department of Oncology, Peking University Shougang Hospital, Beijing, China.

\textsuperscript{18} Medical Oncology, Beijing Chest Hospital, Capital Medical University, Beijing Tuberculosis and Thoracic Tumor Research Institute, Beijing, China.

\textsuperscript{19} Department of Radiology, Peking Union Medical College Hospital, Chinese Academy of Medical Sciences and Peking Union Medical College, Beijing, China.

\textsuperscript{20} Department of Nuclear Medicine, State Key Laboratory of Complex Severe and Rare Diseases, Beijing Key Laboratory of Targeted Radiopharmaceutical Development and Translational nuclear medicine, Peking Union Medical College Hospital, Chinese Academy of Medical Sciences \& Peking Union Medical College, Beijing, China.

\textsuperscript{21} Department of Pathology, Peking Union Medical College Hospital, Chinese Academy of Medical Sciences and Peking Union Medical College, Beijing, China.

\textsuperscript{22} Department of Radiation Oncology, Peking Union Medical College Hospital, Chinese Academy of Medical Sciences and Peking Union Medical College, Beijing, China.

\textsuperscript{23} Department of Radiation Oncology, National Cancer Center / National Clinical Research Center for Cancer / Cancer Hospital, Chinese Academy of Medical Sciences and Peking Union Medical College, Beijing, China.

\textsuperscript{24} Department of Rheumatology and Clinical Immunology, Peking Union Medical College Hospital, Chinese Academy of Medical Sciences and Peking Union Medical College.

\textsuperscript{25} Department of Respiratory and Critical Care Medicine, Peking Union Medical College Hospital, Chinese Academy of Medical Sciences and Peking Union Medical College, Beijing, China.

\textsuperscript{\#} These authors contributed equally to this work.

\textsuperscript{*} Corresponding authors: Li Zhang (zhanglipumch@aliyun.com), Shirui Wang (wsr@medlinker.com) and Naixin Liang (pumchnelson@163.com).\par}
\vspace{1em}

\begin{abstract}
\textbf{Background} Complex lung cancer requires clinicians to compare treatment pathways, connect comorbidity and organ reserve to treatment eligibility, intensity and sequence, and identify missing information that could change the plan. We developed MedGPT Clinical Explorer (MCE) to support this work through reviewable conditional strategies informed by multidisciplinary reasoning.

\textbf{Methods} We conducted an offline evaluation of 100 purposively selected cases. Twenty chief physicians from 15 institutions and six disciplines constructed clinical references for 40 unpublished internal cases. We compared 2,250 final strategies from 250 physicians at 98 institutions under unaided, retrieval-reference and MCE-assisted conditions. The primary outcome was the Admissible Pathway Attainment Score (APAS; 0--100), measuring attainment of applicable case-specific clinical requirements. Retrieval reference and MCE shared a knowledge base. Supporting analyses examined format-aligned outputs, clinical acceptability and connections between pathways, prerequisites and subsequent actions.

\textbf{Findings} MCE-assisted physician strategies had higher APAS than unaided strategies (adjusted difference 12.87 points, 95\% CI 11.18--14.55) and retrieval-reference strategies (5.22, 3.52--6.93). Differences were largest for candidate pathways and their eligibility or redirection conditions. In a supporting case-paired comparison, format-aligned MCE outputs scored 19.11 points (15.58--22.33) higher than the mean physician APAS for the same case under MCE support, despite being shorter. APAS correlated with physicians' whole-strategy acceptability judgments (Spearman's \ensuremath{\rho} = 0.671, 95\% CI 0.581--0.752).

\textbf{Interpretation} MCE supports physicians in comparing reasonable treatment options, establishing treatment prerequisites and planning how to revise a strategy when information or risks change. In the offline study, MCE-assisted physicians expressed applicable clinical requirements more fully; prospective studies should evaluate MCE's effects on treatment decisions, workload and patient care.

\textbf{Funding} Peking Union Medical College Hospital Talent Cultivation Program (Category B; UGG10521) and National High Level Hospital Clinical Research Funding (2025-PUMCH-A-035).

\end{abstract}

\section{Introduction}

Complex lung cancer decisions often require clinicians to compare several reasonable treatment pathways whose suitability depends on stage, molecular findings, previous treatment, comorbidity and organ reserve\cite{ref1,ref2,ref3}. Unresolved information can change treatment intent, eligibility, intensity or sequence; competing goals and risks can also prevent immediate commitment to a single plan\cite{ref4,ref5}. We use \emph{complex lung cancer} to describe this decision state: available information and clinical constraints leave substantive alternatives or decision-changing questions unresolved at a particular decision point. The term describes decision structure rather than disease severity or prognosis.

Multidisciplinary team (MDT) discussion is a standard approach to these interdependent decisions, bringing complementary expertise together to assess diagnosis, treatment eligibility, risk and sequencing\cite{ref6,ref7,ref8,ref9}. Access to this input requires case preparation, team coordination and participation by relevant specialists. These demands consume clinical time and personnel resources and affect timely access to comprehensive multidisciplinary review\cite{ref6,ref10}. Clinicians therefore need support that brings key elements of multidisciplinary reasoning into individual strategy formulation. While awaiting team discussion, or when a full team cannot be convened promptly, such support should help clinicians compare reasonable pathways, identify questions that must be resolved before treatment and specify disagreements requiring further specialist judgment.

Medical large language model research has examined reliability and limitations in clinical decision-making, patient interaction and human evaluation\cite{ref11,ref12,ref13}. Agentic systems have extended support to rare-disease diagnosis with traceable reasoning\cite{ref14}, investigation and treatment actions in simulated electronic health record environments\cite{ref15}, and disease management across repeated consultations\cite{ref16}. In oncology, studies have evaluated guideline-based decision support within MDTs and its implementation\cite{ref17,ref18}, as well as large language model recommendations for complex cases\cite{ref19}. Other work has examined support for neuro-oncology radiotherapy\cite{ref20} and virtual tumor boards for patients with non-small-cell lung cancer receiving immunotherapy\cite{ref21}. These studies provide a foundation for AI participation in individual and multidisciplinary decision-making.

When tumor-control goals conflict with comorbidity, organ reserve or treatment toxicity, clinicians need to consider treatment options alongside the conditions the patient can tolerate. The support addressed here should help clinicians retain reasonable primary and alternative pathways, identify diagnostic or tolerance-related questions that must be resolved before each can proceed, and connect patient risks to treatment eligibility, intensity and sequence. The strategy should also specify advancement, deferral, redirection or reassessment when key premises are supported, refuted or remain unresolved, with a fallback when the preferred option is infeasible. Such support is intended to help individual physicians formulate a plan for clinical discussion and focus specialist consultation on questions that could change it.

We developed MedGPT Clinical Explorer (MCE), a medical-agent system built on MedGPT\cite{ref22}, to help individual physicians compare reasonable treatment pathways, identify what must be resolved before each can proceed, and plan subsequent actions when key premises or patient risks change. MCE organizes treatment options, eligibility conditions and contingent next steps informed by multidisciplinary reasoning into a reviewable \emph{conditional strategy} that clinicians can question, revise and bring to focused specialist discussion. We evaluated MCE within a purposive 100-case corpus, with case-specific references constructed by a multidisciplinary expert panel for 40 cases. In an offline task, 250 physicians from 98 institutions submitted 2,250 final strategies under unaided, retrieval-reference and MCE-assisted conditions. The primary question was whether physicians' final strategies expressed applicable clinical requirements more fully under MCE support than under unaided or retrieval-reference conditions. Complementary evaluations examined the clinical meaning of content scores and the coherence of pathway-condition-action relationships in complete MCE outputs and purposive case variants.

\section{Results}

\subsection{Overlapping clinical constraints and case-specific references defined the conditional decision task}

At a clinical decision point, known facts, missing or conflicting information, current goals, applicable evidence and risk constraints jointly define the treatment options available for comparison. An open-ended cancer decision therefore requires clinically distinct candidate strategies with explicit enabling, limiting and exit conditions, together with unknowns that could change pathway eligibility, sequence or intensity. Candidates with different benefit, risk and implementation burdens can remain under review together. They also require arrangements for advancement, modification, reassessment or fallback; inclusion indicates that an option warrants comparison, while its ultimate acceptability remains a case-level clinical judgment. Figure 1a illustrates this structure using generalized decision elements: support, refutation or continuing uncertainty about premises such as disease extent, treatment tolerance and patient goals should connect candidates to appropriate subsequent actions. Evaluation covered five prespecified relationships: candidate-pathway differentiation, action-premise links, decision-changing unknowns, verification-candidate links and reassessment or fallback.

The study included 100 purposively selected complex lung cancer cases organized as fixed input snapshots: 47 internal cases and 53 published cases. Five non-exclusive clinical gray-zone attributes formed 13 exact combinations. Guideline recommendations were unclear in 74 cases, and interacting factors impeded direct use of a standard pathway in 63. Comorbidity or treatment constraints conflicted in 47 cases, while pathway timing or strategy choice could materially affect subsequent outcomes in 45. One case recorded a real clinical disagreement. Seventy-eight cases met at least two attributes; the largest exact intersection comprised 31 cases with unclear guideline recommendations, interacting factors and pathway-timing constraints (Table 1; Fig. 1b). These overlaps describe co-occurring decision constraints in this corpus, not a validated scale of complexity, severity, prognosis or support need.

\textbf{Table 1 | Case corpus and prespecified attributes of complex clinical decisions}

\begingroup
\footnotesize
\setlength{\tabcolsep}{3pt}
\renewcommand{\arraystretch}{1.12}
\begin{longtable}{@{}>{\raggedright\arraybackslash}p{0.347\linewidth}>{\raggedright\arraybackslash}p{0.161\linewidth}>{\raggedright\arraybackslash}p{0.182\linewidth}>{\raggedright\arraybackslash}p{0.250\linewidth}@{}}
\toprule
\textbf{Characteristic} & \textbf{Overall (N = 100)} & \textbf{Expert-reference set (N = 40)} & \textbf{Source-masked content-evaluation set (N = 60)} \\
\midrule
\endfirsthead
\toprule
\textbf{Characteristic} & \textbf{Overall (N = 100)} & \textbf{Expert-reference set (N = 40)} & \textbf{Source-masked content-evaluation set (N = 60)} \\
\midrule
\endhead
\midrule
\multicolumn{4}{r}{\footnotesize Continued on next page} \\
\endfoot
\bottomrule
\endlastfoot
\textbf{Case source} &  &  &  \\
Internal case & 47 (47.0\%) & 40 (100.0\%) & 7 (11.7\%) \\
Published case & 53 (53.0\%) & 0 & 53 (88.3\%) \\
\textbf{Demographic information} &  &  &  \\
Age, median (IQR), years & 64 (57-69); n = 99 & 65.5 (58.5-69.5); n = 40 & 63 (56.5-68); n = 59 \\
Male & 51 (51.0\%) & 28 (70.0\%) & 23 (38.3\%) \\
Female & 30 (30.0\%) & 12 (30.0\%) & 18 (30.0\%) \\
Sex not reported & 19 (19.0\%) & 0 & 19 (31.7\%) \\
\textbf{Prespecified gray-zone attributes, non-exclusive} &  &  &  \\
Guideline recommendation unclear & 74 (74.0\%) & 26 (65.0\%) & 48 (80.0\%) \\
Major conflict from comorbidity or treatment constraints & 47 (47.0\%) & 26 (65.0\%) & 21 (35.0\%) \\
Standard pathway not directly applicable because of interacting factors & 63 (63.0\%) & 11 (27.5\%) & 52 (86.7\%) \\
At least two prespecified gray-zone criteria & 78 (78.0\%) & 24 (60.0\%) & 54 (90.0\%) \\
\end{longtable}
\endgroup

Data are n (\%) unless otherwise stated. The expert-reference and source-masked content-evaluation sets were purposively constructed for different study functions; no baseline balance test was performed between them. Operational definitions, complete attribute counts and exploratory primary decision scenarios are in Supplementary Table 10.

To establish case-specific clinical references, a multidisciplinary panel independently constructed the Expert-Admissible Strategy Space (EASS) from fixed materials for 40 unpublished internal cases. The panel comprised 20 chief physicians from 15 institutions and six disciplines: thoracic surgery, respiratory medicine, medical oncology, radiology, pathology and radiation oncology. MCE, comparator and physician outputs were not shown during reference construction. EASS organized requirements into five clinical-content domains: clinical problem and treatment goals (M1), decision-critical information (M2), candidate clinical pathways (M3), safety boundaries and risk governance (M4), and reassessment and governance (M5). The final reference contained 240 rules, six per case: 139 confirmed acceptable and 101 conditionally acceptable. Rules also recorded inadmissible conditions, fallback pathways, reasonable disagreement and unresolved content (Fig. 1c; Supplementary Methods 1; Supplementary Table 1).

The Admissible Pathway Attainment Score (APAS) measured the extent to which a system or physician strategy satisfied applicable clinical requirements for each case. The Complete-and-Safe Pathway Rate (CSPR) used a separate nine-item joint gate to assess whether a text simultaneously met completeness and key-constraint requirements; \emph{safe} refers to satisfaction of prespecified scoring constraints, not measured clinical safety. EASS rules were not provided to MCE, comparators or physicians during strategy generation. The study therefore had two complementary evaluation routes. One compared expression of required clinical content in 2,250 final strategies from 250 physicians under unaided, retrieval-reference and MCE-assisted conditions. The other examined connections among candidates, premises, unknowns, verification results and subsequent actions in 40 complete MCE outputs (Fig. 1c). Expression of clinical requirements and correct connections between them are distinct evaluation dimensions; potentially reasonable pathways not encoded in the reference remained subject to clinical judgment.

\subsection{MCE linked evidence boundaries and decision-changing unknowns to pathway advancement, de-escalation and fallback}

An archived run illustrated how MCE linked evidence applicability and decision-changing unknowns to candidate pathways and fallback arrangements. The case concerned a 68-year-old patient with suspected postoperative recurrence and possible chest-wall involvement, alongside interstitial lung disease, continuous oxygen dependence and an Eastern Cooperative Oncology Group performance status of 3. The central conflict was whether an opportunity for local control justified the risk of irreversible pulmonary injury. MCE jointly represented previous treatment, suspected recurrence, limited functional and respiratory reserve and pulmonary-toxicity risk. Pathological confirmation of recurrence, thoracic and distant restaging, pulmonary tolerance, interstitial-lung-disease activity and radiotherapy feasibility remained unresolved because they could change pathway eligibility or treatment intensity. This single archived run was the illustrative analysis unit (Fig. 2).

The output organized evidence propositions around local-treatment eligibility and pulmonary-toxicity risk, recording whether evidence supported, limited or left candidate applicability unresolved. Conservative local radiotherapy ranked first in the system's heuristic ordering in this run. More intensive local control remained a conditional option, while low-burden palliative radiotherapy or best supportive care formed the fallback.

Imaging and treatment-planning assessments, low-risk pathological confirmation and assessment of respiratory and interstitial-lung-disease tolerance were linked to verification tasks that could change candidate pathways. When a key premise was supported, the output proposed retaining or advancing local treatment; refutation prompted a lower-burden palliative or supportive pathway. Continuing uncertainty retained the more conservative pathway with bridging, delayed escalation and reassessment. The output thus explicitly connected possible verification results to advancement, de-escalation and fallback. The subsequent physician analysis examined whether access to this support was associated with fuller expression of applicable case-level clinical requirements in final strategies.

\subsection{MCE-assisted physicians expressed more clinical requirements, especially candidate pathways}

The offline task included 250 physicians from 98 institutions, each completing nine different cases: three under each of the unaided (U), retrieval-reference (R) and MCE-assisted (M) conditions. Their 2,250 final strategies had adjusted APAS of 53.40 (95\% CI, 51.32 to 55.48) under U, 61.05 (58.95 to 63.15) under R and 66.27 (64.18 to 68.36) under M. The adjusted M--U difference was 12.87 points (95\% CI, 11.18 to 14.55; Holm P = 8.49 \ensuremath{\times} \ensuremath{10^{-51}}), and the M--R difference was 5.22 points (3.52 to 6.93; Holm P = 1.98 \ensuremath{\times} \ensuremath{10^{-9}}; Fig. 3a). Under MCE support, physicians' final strategies expressed applicable clinical requirements more fully than under either comparison condition.

All five prespecified clinical-content domains had positive M--U and M--R differences, with Holm P < 0.001 for all ten comparisons. Candidate clinical pathways (M3) had the largest descriptive differences: 20.64 points (95\% CI, 18.15--23.13) for M--U and 8.59 points (6.07--11.12) for M--R. The increment principally concerned primary and alternative pathways and their eligibility or redirection conditions, with fuller expression of decision-critical information and risk constraints as well (Fig. 3b; Supplementary Table 9b).

In a supporting comparison, one additional MCE output for each of the 40 expert-reference cases used the physicians' five-part response structure. Physician APAS was averaged separately within each case and condition, then paired with the system output for that case. System-output-minus-physician-mean differences under U, R and M were 32.03 points (95\% CI, 26.48 to 37.09), 24.29 points (19.94 to 28.60) and 19.11 points (15.58 to 22.33), respectively; all three supplementary paired tests had Holm P < 0.001. Relative to M-condition physician responses, the paired difference in non-whitespace characters was \ensuremath{-}247.5 (95\% CI, \ensuremath{-}313.4 to \ensuremath{-}185.2; Holm P < 0.001; Fig. 3c). All five domain differences were positive, with the largest for M3: 29.65 points (23.54--35.30; Supplementary Table 9e). This comparison described within-case differences between system outputs and final physician strategies, distinct from the principal comparison between physician-support conditions.

To characterize support in physicians' final strategies across clinical tasks, a post hoc analysis used 2,133 responses from 237 physicians with classifiable hospital backgrounds. The 40 cases were complex on the comorbidity axis only, the lung-cancer axis only or both axes (9, 13 and 18 cases, respectively). Candidate pathways (M3) had the largest M--U point estimate among the five domains in all six hospital-by-category combinations, ranging from 13.33 to 25.77 points; all six unadjusted 95\% confidence intervals excluded zero (Fig. 3d).

Relative to retrieval reference, the M3 differences in general hospitals were 10.12 points (95\% CI, 4.42 to 15.81), 9.04 (3.40 to 14.67) and 8.79 (4.20 to 13.37) in the comorbidity-only, lung-cancer-only and both-complex categories, respectively. Corresponding estimates for physicians from cancer or thoracic/pulmonary-specialty hospitals were 4.93 (\ensuremath{-}3.05 to 12.91), 0.44 (\ensuremath{-}13.22 to 14.10) and 9.07 (\ensuremath{-}0.005 to 18.14). In both-complex cases, the M--U point estimates were 23.91 and 13.33 points in general and specialty hospitals, respectively, whereas their M--R point estimates were close. Four of the six M--U comparisons and two of the six M--R comparisons retained Holm-adjusted support within the complete 60-comparison family. None of the 10 joint tests of hospital-by-category increment heterogeneity retained Holm-adjusted support (Extended Data Figure 9; Supplementary Table 9k).

Decision-critical information (M2) showed a further content-specific pattern in the comorbidity-only category. M--R differences were 10.65 points (95\% CI, 5.88 to 15.42) in general hospitals and \ensuremath{-}3.23 (\ensuremath{-}10.39 to 3.92) in specialty hospitals. The direct specialty-minus-general increment difference was \ensuremath{-}13.88 points (\ensuremath{-}22.47 to \ensuremath{-}5.29; Holm P = 0.075). Existing EASS requirements containing explicit links between accompanying constraints and treatment pathways accounted for 15/54 rules (27.8\%) in the comorbidity-only category, 4/78 (5.1\%) in the lung-cancer-only category and 26/108 (24.1\%) in the both-complex category. These requirements included characterization of accompanying lesions before systemic therapy, changes to local-treatment sequencing when infection was indicated, and safety-prioritized pathway selection after pulmonary toxicity (Supplementary Tables 9k and 13).

Across the three professional-title groups, all six M--U and M--R APAS CIs excluded zero; the exploratory interaction test gave P = 0.548. Secondary CSPR was 70.2\% under M, 61.3\% under R and 44.1\% under U, with 13 indeterminate responses (Extended Data Figure 1; Supplementary Tables 9a and 9c). A separate subsequent-event exploration, excluding six cases without follow-up and three with date or linkage concerns, retained positive APAS contrasts in both the 12 event-recorded and 19 non-recorded cases. Full estimates and interactions are in Supplementary Tables 9f--9h; these text-score comparisons do not estimate effects on subsequent events.

\subsection{Conditional clinical-content increment of complete MCE outputs and support from the functional architecture}

A supportive comparison using the same 40 expert-reference cases showed that complete MCE outputs had higher APAS than strategies generated by GPT-5.4, Claude Opus 4.7 and Gemini 3.1 Pro across all nine model--judge combinations (all nine supplementary paired-test Holm P < 0.001). Mean APAS ranged from 94.71 to 97.72 for complete MCE and from 53.16 to 74.34 for the other-model outputs (Fig. 4a; Supplementary Table 9d). Across applicable cases, complete MCE also had higher estimates in all five clinical-content domains, with all 45 domain contrasts retaining Holm P < 0.01. Differences were largest descriptively for candidate clinical pathways (M3) in seven of nine combinations, with within-case M3 differences of 28.75--56.25 points (Fig. 4b). Thus, the increment centered on primary and alternative pathways and their eligibility and redirection conditions, rather than only restating the clinical problem.

Independent source-masked clinical evaluation supported this content difference. Experts preferred MCE text in 100 of 120 within-case A/B modules (83.3\%). Among 60 modules selected for independent re-review, MCE was preferred in 54 (90.0\%) in the first round and 57 (95.0\%) at re-review. The three-category judgment was identical across rounds for 55 of 60 modules (91.7\%; Gwet AC1 = 0.910; Extended Data Figure 4a; Supplementary Table 6). Dimension-specific exact agreement was 66.7\% for key clinical conflicts and treatment goals, 73.3\% for patient state and treatment fit, 80.0\% for safety constraints and unacceptable risk, 93.3\% for decision-changing verification steps and 80.0\% for subsequent pathways and alternatives (Extended Data Figure 4b). These values describe the reproducibility of judgments on the same fixed modules, not preference rates for MCE within each dimension.

The functional-configuration analysis further associated the MCE architecture with higher clinical-content coverage. Under GPT-5.5 scoring, removing the multidisciplinary-prior function reduced APAS by 16.62 points (95\% CI, 11.84--21.40), removing the simulation--critique--repair block by 3.60 points (1.72--5.49), and jointly removing pathway and qualifying-evidence construction together with simulation, critique and repair by 7.72 points (4.87--10.57). Complete MCE also exceeded MedGPT base direct generation by 23.82 points (15.95--31.70) and MedGPT-LightRAG by 14.34 points (9.11--19.57; Fig. 4c; Supplementary Table 7). All five supplementary paired comparisons retained Holm P < 0.001. Configuration rankings were similar across the three judges (Spearman's \ensuremath{\rho} = 0.943--1.000). These results indicate that higher coverage was associated with the realized functional architecture rather than orchestration alone, although the grouped contrasts do not identify individual-node effects.

Together, these comparisons associate the complete MCE architecture with fuller expression of applicable case-level clinical requirements, particularly candidate pathways and their eligibility or redirection conditions. To interpret the clinical meaning of these differences, subsequent analyses examined how the scores related to whole-strategy clinical acceptability and whether treatment choices were coherently connected to their prerequisites and subsequent actions.

\subsection{APAS has a clinical-interpretation basis, with limits set by relationship closure}

The principal comparisons used APAS to measure expression of applicable case-level requirements from the EASS reference. We examined its alignment with physicians' whole-strategy clinical judgments, reproducibility across physician raters and response to controlled text changes. In parallel, relationship audits directly assessed whether candidate treatments, enabling premises, verification results and subsequent actions were correctly connected, complementing the content score.

Clinical anchoring included 120 outputs from 40 case clusters. Two attending thoracic surgeons independently rated overall acceptability for clinical discussion and major clinical defects. Using the prespecified binary grouping, they classified 74/120 and 81/120 outputs as acceptable, with 71/120 accepted by both. Their original five-category judgments agreed exactly for 89/120 outputs; the four-category ordinal ratings had 97.5\% adjacent agreement and a linearly weighted \ensuremath{\kappa} of 0.745 (95\% CI, 0.671--0.815). Across the same outputs, APAS correlated with the mean ordinal acceptability score (Spearman's \ensuremath{\rho} = 0.671, 95\% CI, 0.581--0.752; Fig. 5a,b; Extended Data Figure 2; Supplementary Table 9i). Thus, higher APAS aligned with movement toward more acceptable whole-strategy judgments. Complete category distributions and major-defect agreement are reported in Supplementary Table 9i.

In a separate re-rating of 240 rules in 40 complete MCE reports, physician agreement was 78.33\% at the rule level, with linearly weighted Gwet AC2 of 0.842 (95\% CI, 0.785--0.893). Case-level ICC(A,1) was 0.694 (0.468--0.835), indicating that the reference was reproducible at the rule level while absolute scores for individual reports retained rater-related uncertainty (Fig. 5b; Supplementary Table 9i).

Controlled perturbations assessed whether APAS responded to presentation order or removal of targeted content. Across 36 original--variant triplets from 12 cases, section reordering changed overall APAS by +0.85 points (95\% CI, \ensuremath{-}0.44 to 2.31), whereas targeted deletion changed it by \ensuremath{-}0.66 points (\ensuremath{-}1.69 to 0.36). After deletion, APAS decreased in 20/36 triplets, increased in 15 and was unchanged in one; domain-specific responses were also heterogeneous. Thus, limited reordering did not produce a clear systematic shift, but domain-selective sensitivity to content deletion was not established (Extended Data Figure 3; Supplementary Table 9j). Together, these analyses support APAS as a measure of encoded case-level content in fixed cases, but not as a substitute for clinical judgment or an assessment of how strategy elements are connected.

We therefore directly audited five prespecified relationships in 40 frozen complete MCE outputs. Three AI judges independently classified 36/40, 30/40 and 40/40 outputs as complete across all five relationships. Differences were concentrated in action--premise and verification--candidate links. At least one AI judge flagged a conflict, conditional misconnection or omission in 12/40 outputs; targeted review of ten unresolved branches confirmed nine subsequent actions and one omission (Extended Data Figure 5; Supplementary Table 11). Detailed relationship-specific counts and verification-task states are reported in Supplementary Table 11.

Purposive case variants tested these relationships under predefined condition changes. Among 48 offline scenarios derived from 16 source-case templates, 13 met all six requirements, 23 required clinical review and 12 contained at least one clear relationship gap, most often involving pathway prerequisites, sequence or fallback (Extended Data Figures 6 and 7; Supplementary Table 8). In independent physician re-review, exact agreement was higher for the overall scenario classification than for individual requirements (93.75\% versus 67.01\%), whereas each physician agreed with AI consensus on 39.93\% of requirements (Extended Data Figure 8). These findings indicate that structural audits can localize material requiring review but cannot replace clinical adjudication.

APAS and relationship auditing therefore address complementary properties. APAS compares expression of case-level clinical requirements, whereas the relationship analyses assess whether treatment intent, pathway eligibility, action sequence and fallback are coherently connected.

\section{Discussion}

Complex lung cancer requires integration of multidisciplinary judgment at the point of strategy formulation. MCE provides individual physicians with reviewable support by organizing candidate pathways, decision-critical information and treatment constraints into a conditional strategy. In the offline physician task, MCE-assisted final strategies expressed expert-defined clinical requirements more fully than unaided and retrieval-reference strategies, with adjusted APAS differences of 12.87 and 5.22 points, respectively. The largest content increment concerned candidate pathways and the conditions governing their eligibility and redirection.

Multidisciplinary expert participation and cross-institutional physician participation served different purposes in this study. Twenty chief physicians from 15 institutions and six disciplines independently constructed case-specific references, allowing treatment goals, diagnostic information, local and systemic options, and patient-tolerance constraints to be reviewed within a common evaluation framework. The 250 physicians from 98 institutions provided a large-scale, multicenter evaluation based on the final strategies they formulated themselves, examining whether provision of support was reflected in those strategies. These complementary forms of participation linked multidisciplinary clinical requirements to individual strategy formulation, enabling evaluation of both the generated support and how fully physicians' final strategies addressed the same clinical requirements under different support conditions.

These content differences concern requirements that can change treatment decisions. Staging and resectability determine treatment intent and local-treatment eligibility; pathology and molecular findings alter systemic options; and organ reserve, comorbidity and toxicity risk constrain intensity and influence monitoring and stopping conditions\cite{ref1,ref2,ref3}. MCE-assisted physician strategies expressed more of the requirements governing when treatment can proceed, which alternatives remain reasonable and when reassessment is needed. Format-aligned MCE outputs also had higher APAS than the mean physician APAS for the same case under MCE support despite being shorter, showing that these requirements could be organized compactly. Their clinical value is to provide a basis for comparing reasonable options and make potentially consequential omissions in treatment eligibility, risk control or pathway redirection visible and reviewable before action. Differences between the support content and final physician strategies also motivate recording how clinicians use that content in subsequent studies.

The exploratory patterns raise a clinical question about how physicians translate reference information into candidate pathways. In the lung-cancer-only category, the specialty group's near-zero M--R point estimate contrasted with its positive M--U estimate. One possible explanation is that physicians familiar with lung-cancer treatment can formulate relevant alternatives from retrieval material, whereas explicitly assembled pathways and their eligibility conditions provide additional support in general-hospital settings. The wide specialty-group interval leaves the magnitude uncertain, but the pattern motivates examining how treatment familiarity influences the use of different forms of support.

In the comorbidity-only category, the information-support pattern directs attention to how accompanying findings become treatment prerequisites. Recognizing a coexisting condition does not by itself specify whether it requires further characterization, changes treatment eligibility or alters the sequence of cancer care. The reference requirements made these questions concrete through accompanying lesions, infection and pulmonary-toxicity constraints. A useful hypothesis is that MCE supports the formulation of decision-changing prerequisites, with the contribution depending on how clinicians already incorporate these findings into their plans.

When both complexity axes were present, both hospital groups had M--R pathway point estimates of approximately nine points despite different M--U increments. A possible shared task is to retain competing cancer-treatment alternatives while incorporating accompanying constraints on eligibility, intensity and sequence. Requirements explicitly linking constraints to pathways give clinical content to this interpretation. Together, the patterns motivate evaluating support according to the decision work it enables: translating reference information into alternatives, identifying treatment prerequisites and combining pathways with patient constraints. These functions connect individual strategy formulation with focused specialist review.

Conditional strategies bring distributed evidence, patient constraints and reasonable alternatives into clinicians' current treatment decisions while identifying questions requiring specialist input or collective judgment. Agents for diagnosis, sequential action and longitudinal management provide relevant capabilities\cite{ref14,ref15,ref16}; oncology research has also applied guideline support, complex-case recommendations and virtual tumor boards to individual or MDT decisions\cite{ref17,ref18,ref19,ref20,ref21}. Building on this work, MCE focuses on organizing concurrent candidate pathways and their eligibility, exit and redirection conditions, and evaluates how this support relates to physicians' final strategies. APAS was higher under M than R despite a shared knowledge base; together with source-masked evaluation and configuration comparisons, these findings support the value of organizing information as conditional strategies. Removing multidisciplinary priors produced the largest decrease among the targeted ablations, although grouped configuration comparisons do not identify an individual component's independent contribution.

System-generated support and final physician strategies require separate evaluation. Prior case-based studies found that model assistance did not necessarily improve physicians' diagnostic reasoning and could increase time per case despite higher management scores\cite{ref23,ref24}. How clinicians review, select and modify support is therefore an important step in realizing clinical value. APAS aligned with whole-strategy clinical acceptability, and EASS re-rating showed good rule-level reproducibility; acceptability here concerned suitability for clinical discussion. Limited section reordering produced no clear systematic score shift, but deletion responses were heterogeneous and domain-selective sensitivity was not established. Relationship audits complemented these measures by checking connections among treatment intent, eligibility, sequence and fallback: at least one AI judge flagged a conflict, conditional misconnection or omission in 12 of 40 complete outputs. Limited physician-AI agreement on individual relationship states further indicates that these audits should inform clinical judgment.

The main limitations concern case representation, clinician use and measurement. The purposively enriched fixed cases define the empirical target, and all cases had development or evaluation exposure. Offline final texts did not record which MCE elements physicians viewed, adopted, modified or rejected, and the task was not embedded in a real MDT workflow. EASS does not exhaust all reasonable strategies, principal estimates depend on fixed AI judges, and human calibration of U/R/M texts remains incomplete. Hospital/category comparisons were post hoc; hospital background did not directly measure treatment familiarity, resources, MDT access or treatment capability. The study did not directly compare MCE-supported clinicians with a human MDT or measure resource savings or patient benefit. Cross-model and configuration comparisons were affected by differences in version and presentation. Follow-up lacked a uniform window and independent adjudication, so subsequent-event strata provide context for clinical-consequence mapping rather than treatment-effect estimates. Support accuracy and interaction design can also influence reliance and inappropriate adoption\cite{ref25,ref26,ref27,ref28,ref29}.

Prospective studies should examine how conditional strategies enter individual care and multidisciplinary work. Relevant outcomes include case-preparation and decision time, availability and actual use of specialist input, decision delays caused by missing information, plan completeness and unresolved disagreement. Studies should record clinicians' adoption, revision or rejection of candidate pathways, treatment prerequisites and risk constraints, with prespecified analyses of whether case complexity and treatment familiarity modify the use of different support approaches. Evaluation should also capture cognitive burden, propagation of erroneous suggestions, relationship gaps and escalation needs. Staged clinical-AI evaluation and governance principles provide a framework\cite{ref30,ref31,ref32}. A pragmatic cluster-randomized study of generative-AI decision support offers a workflow-level design reference\cite{ref33}, but complex lung cancer requires task-specific outcomes and reporting\cite{ref34,ref35}. Subsequent studies can assess treatment delivery, adherence to risk constraints, reassessment after new information or toxicity, and patient outcomes.

MCE provides individual physicians with multidisciplinary decision support for comparing reasonable treatment pathways, identifying questions that must be resolved before treatment and planning subsequent actions when conditions change. The offline physician study showed fuller expression of applicable clinical requirements under MCE support, particularly candidate pathways and their eligibility or redirection conditions. Its clinical purpose is to help physicians identify consequential omissions before treatment decisions and focus specialist discussion on questions that could change treatment choices; prospective studies should determine its actual effects on treatment decisions, workload and patient care.

\section{Methods}

\subsection{Study design and estimands}

This offline evaluation used fixed inputs from complex lung cancer cases and centered on physicians' strategy formulation. The primary analysis compared final physician strategies under three information-support conditions. Complementary analyses examined whether MCE produced reviewable conditional strategies, whether the content score aligned with clinical judgment, and whether pathways, premises, verification results and subsequent actions were coherently linked in complete outputs and purposive case variants. Subsequent-event records were used only in a post hoc exploration of the clinical context of differences in text scores. Each analysis retained its own unit, denominator and estimand; results were not pooled into an overall performance measure.

The primary estimands for the physician task were differences in attainment of the clinical requirements for each case under M relative to U and R. They describe differences in final texts within the fixed case set and conditions of information support. Post hoc hospital-background and case-category analyses characterized the clinical tasks associated with the domain-level increments. Analyses of subsequent events, comparisons among six configurations sharing a MedGPT foundation, source-masked content evaluation, overall clinical anchoring, controlled text perturbations and case-condition changes were supporting or exploratory analyses.

\subsection{Case sources and analysis sets}

Case material was organized using a common template with a case summary and the treating physician's original plan. Fixed inputs used for system generation and evaluation contained six fields: basic information, presenting complaint, history of the present illness, medical and other history, investigations and diagnosis. They excluded management, the treating physician's recommendation, subsequent plans and outcomes. MCE generation, EASS construction and the U/R/M physician task used only the case summary. The treating physician's original plan was used only as one comparator in the source-masked content evaluation.

The 100 cases comprised 47 internal and 53 published cases. Forty unpublished internal cases formed the expert-reference set used for EASS construction, the six-configuration comparison and the physician task. The remaining 7 internal and 53 published cases formed the 60-case source-masked content-evaluation set. Cases were purposively selected using prespecified gray-zone attributes; no validated continuous scale was used to stratify complexity or decision uncertainty. A case registry recorded source, study set, input version, reuse across branches and exposure during development. All cases had been used in development or evaluation and therefore did not form a development-naive external validation set. Forty-nine publications contributed the 53 public cases: 48 were identified by DOI and one by PMID and a stable public link (Supplementary Table 12).

Demographic fields were extracted from structured case summaries; missing age and sex were retained as missing or not reported. After de-identification and structured abstraction, five non-exclusive gray-zone attributes were recorded from the case summary and available multidisciplinary or treating records. The first three attributes were unclear guideline recommendations, major conflict from comorbidity or treatment constraints, and documented decision disagreement. The other two were pathway timing or strategy choice with substantial potential consequences and interacting factors that prevented direct use of a standard pathway. Each attribute was recorded for each case as present or absent; cases meeting at least two were classified as multi-attribute gray-zone cases. These labels described the purposive corpus and were not outcomes, model covariates or performance strata. Operational definitions and complete counts are in Supplementary Table 10.

\subsection{EASS case-specific clinical reference and scoring rules}

A multidisciplinary panel of 20 experts constructed EASS. Nineteen participated in the first in-person round, and the expert absent from that meeting joined subsequent online rounds. Fourteen treatment-related specialists conducted later case review. EASS was developed independently from fixed material for the 40 expert-reference cases; candidate MCE, comparator and physician outputs were not provided to the panel. The process drew on independent judgment, prespecified anchors, structured feedback and preservation of disagreement from the RAND/UCLA Appropriateness Method, without adopting its panel size, number of rounds or disagreement thresholds\cite{ref36}.

EASS covered five clinical content domains: M1, clinical problem and treatment goals; M2, decision-critical information; M3, candidate clinical pathways; M4, safety boundaries and risk governance; and M5, reassessment and governance. These domains supplied case-level clinical requirements and were distinct from the five structural relationships used to assess whether candidates, premises, unknowns and subsequent actions were connected. Final rules were classified as confirmed acceptable or conditionally acceptable and recorded related inadmissible conditions, fallback pathways, reasonable disagreement and unresolved content.

The final reference contained 240 rules, six per case. Each applicable rule was scored 0, 1 or 2. Domains M1-M4 had weight 3 and M5 had weight 2. APAS was calculated as:

\[
\mathrm{APAS}=100 \times \frac{\sum_i w_i r_i}{2 \times \sum_i w_i}
\]

Only items deemed applicable and included in the denominator under the fixed scoring rules contributed to the sums. Correct use of prespecified fallback, reasonable-disagreement and not-applicable states followed the prespecified rules; not-assessable, technical-missingness and not-applicable-for-clinical-reasons states were retained separately. CSPR was defined by a nine-item joint gate: failure of any applicable gate resulted in non-attainment, and absence of a complete majority was retained as indeterminate. CSPR is a joint result under the judging rules, not a clinical safety rate. The same case-level rule dictionary was used for all scoring, and EASS rules were not provided to MCE, comparators or physicians during generation. Expert procedures and scoring details are in Supplementary Methods 1 and Supplementary Tables 1, 2 and 9; physician-task CSPR results are in Extended Data Figure 1a,b.

\subsection{MCE input and generation of conditional strategies}

MCE received only the fixed case input. Internal cases were managed as limited de-identified data under restricted access. EASS rules for individual cases, APAS and CSPR scoring rules, and Condition-Change Response Rules (CCRRs) used in the case-variant evaluation were not available during generation. MCE used MedGPT as its foundation together with versioned evidence resources and run configurations. It represented the decision state and goals, constructed conditional candidates, linked typed evidence, simulated scenarios, reviewed and repaired candidates, created verification tasks linked to candidates, evaluated stopping criteria and assembled the report (Fig. 2).

For functional reporting, the archived n0-n10 workflow was grouped according to the clinical and computational objects passed between dependent stages rather than treated as 11 independently acting components. Multidisciplinary priors entered at n0. Nodes n1-n2 compiled the case state, goals and constraints; n3-n5 constructed foundational and derived pathways and linked qualifying evidence; n6-n8 performed scenario simulation, critique and candidate repair; n9 performed candidate-strategy reranking and formed candidate-linked verification objects; and n10 controlled stopping assessment and report assembly. This mapping was used to define the grouped functional ablations described below. Because nodes within a block operated on intermediate objects produced by the same block, removing an individual node was not treated as an independently interpretable intervention.

\subsubsection{Decision-state and goal representation}

The decision-state representation organized disease status, previous treatment, comorbidity and organ reserve, timeline, current decision point, known facts, missing and conflicting information and major uncertainty. It did not add case facts from external knowledge. The goal representation recorded clinical goals, the decision window, unacceptable risks, implementation constraints and gaps that could change those goals. Patient-preference fields distinguished explicit patient statements, conservative system inference and unavailable information; an inference was not treated as an original patient statement, informed consent or a record of shared decision-making.

\subsubsection{Versioned evidence resources, conditional candidates and evidence links}

Study runs used a versioned knowledge base and evidence corpus. Candidate construction and dispute checking retrieved evidence for questions that could change pathway eligibility, sequence or intensity. Retrieved information was organized by proposition type into source-linked clinical evidence units (CEUs) that recorded whether evidence supported, limited or left candidate applicability unresolved. Sources were not compressed into a single score across evidence types. The study records support review at the proposition-and-source level; complete PICO fields, formal applicability boundaries, evidence grades, calibrated confidence and broader EvidenceOS functions were not evaluated.

The system generated three clinically distinct foundational candidates and two derived candidates and listed a standard-of-care (SoC) reference separately. Each candidate recorded its primary action, eligibility premises, limitations, relevant evidence, downstream branches, monitoring and transition or fallback. A derived candidate also recorded its source of modification, intended effect, added risks and verification needs. The fixed candidate count was an implementation search budget, not a claim that only those pathways existed, and did not assign clinical priority or evidence level.

\subsubsection{Scenario simulation, candidate review and repair}

For each candidate and the SoC reference, the system represented near-term benefit, near-term risk, intermediate branches and transition signals. It reviewed absolute contraindications, irreversible harm, limits on evidence extrapolation, implementation feasibility and dependencies, and the marginal value of continued search. A candidate could be retained, repaired for a specific gap, or removed because of a hard contraindication, irreparable conflict, conflict with patient constraints or substantive equivalence. Numeric values used within this process supported ranking or checks and were not interpreted as calibrated treatment effects, risks or outcome probabilities for individual patients.

\subsubsection{Candidate-linked verification and report assembly}

Reviewed candidates were placed in a heuristic order comprising one currently top-ranked candidate and a set of candidate-linked verification tasks. The top-ranked candidate was the first candidate in the system ordering, not a clinically optimal pathway. Each verification task recorded the target candidate, purpose, sequence or dependency, and the conditional next action when a key premise was supported, refuted or remained unresolved. General insufficiency of evidence was retained as unresolved and was not equated with refutation.

The stopping assessment checked goals, evidence, candidates, review, repair, verification tasks and residual unresolved items, then either proceeded to report assembly or returned the object to the relevant stage. This was workflow control, not formal convergence, global optimality or clinical clearance. The final Strategy Review Pack summarized the case and decision point, goals and boundaries, candidate pathways, SoC reference, verification tasks, evidence citations, risks, monitoring, reassessment, transitions, fallback and unresolved items. References to multidisciplinary review or governance were proposals and did not indicate that such review had occurred. Further implementation detail and the illustrative run are in Supplementary Methods 2.

\subsection{Audit of strategy relationships in complete MCE outputs}

Forty complete MCE Strategy Review Packs were assessed for five relationships. The first three were candidate-pathway differentiation, links between actions and the premises for proceeding, limitation or exit, and missing information expressed as decision-changing unknowns. The fourth linked verification tasks to a candidate and to subsequent actions after support, refutation or continuing uncertainty. The fifth covered reassessment or fallback when the preferred option was infeasible, risk increased or uncertainty remained. Three AI judges independently rated each output, coding each dimension as complete, partly complete, missing or incorrect, not applicable, or not assessable. Nested objects such as verification tasks did not increase the number of outputs or clinical samples. Criteria and directed checks of the source text are described in Supplementary Methods 2 and Supplementary Table 11.

\subsection{Comparators and output processing}

Condition U presented the case summary alone. Condition R added case-specific reference information produced by a designated model using LightRAG and the same knowledge base available to MCE. Sharing a knowledge base did not imply a shared retriever, index, reranker, prompt, input budget, assembly procedure or output structure. Condition M presented MCE strategy review material in the form of a Strategy Review Pack, which physicians could review before composing their own answer.

Comparators in the source-masked content evaluation were a direct-generation comparator, an independent retrieval-augmented generation output, an MDT-Debate-labelled method and the treating physician's original plan. MedGPT base generation and MedGPT-LightRAG in the six-configuration comparison were separate objects and were not interchanged with these comparators. The cross-model strategy-output comparison was a third set of independent runs. Generation roles and comparison purposes are summarized in Supplementary Table 3.

Explicit source labels were removed for source-masked content evaluation. When the treating physician's original plan was the comparator, MCE text was reorganized into the physician-response structure without rewriting its clinical content; other texts retained their organization. Length, formatting and language could still provide source cues. Structure alignment, length controls and removal of source-identifying labels were assessed separately and did not overwrite the primary stimuli.

\subsection{Physician offline task and allocation}

Question sets were generated before the physician list was imported. Each contained nine different cases, three per U, R and M condition. The initial generation specification used seed 42 and preferentially filled underrepresented case-condition slots. Later expansion continued from existing allocation counts, but the complete extended random state could not be recovered; the study is therefore not described as a fully reproducible randomized trial. After the physician list was imported, physicians received pregenerated sets in sequence, without using specialty, rank, hospital or outcome fields in allocation. Actual case-condition coverage and allocation features are in Supplementary Methods 3 and Supplementary Table 4.

The analysis included 250 physicians who completed all nine tasks, yielding 2,250 responses. Each physician answered multiple cases and each case was answered by multiple physicians, forming an incomplete crossed block structure. Unassigned case-condition combinations were design absences rather than missing outcomes and were not imputed. The final texts did not record item-level viewing, adoption, deletion or revision of MCE content.

\subsection{Exploratory analysis by case complexity and hospital background}

Post hoc analyses characterized the clinical tasks supported by MCE across case categories and hospital backgrounds. The analytic subset comprised 237 physicians, 91 registered institutional codes and 2,133 responses: 167 physicians from general hospitals and 70 from cancer or thoracic/pulmonary-specialty hospitals. Hospital background described organizational context, not measured resources or treatment capability. The primary 250-physician analysis remained unchanged.

Physician-reviewed labels classified the 40 cases as comorbidity-complex only (9), lung-cancer-complex only (13), or complex on both axes (18). The comorbidity axis included other tumors and accompanying disease, organ, nutritional or functional constraints; the lung-cancer axis concerned lung cancer and its treatment. A single-axis label indicated that only that axis was marked complex, not absence of the other disease. Categories were unordered and contained no neither-complex group (Supplementary Tables 4 and 10).

Each domain used a REML linear mixed model with a physician random intercept. Fixed effects retained condition-by-case, condition-by-hospital, condition-by-professional-title group and task position, with additional hospital-by-category baseline and condition-by-hospital-by-category contrasts. Category main effects were absorbed by case fixed effects. Standardization weighted cases equally within category, balanced nine task positions and used the common professional-title distribution. Institutional-code CR2 inference used Satterthwaite single-contrast and HTZ joint tests; 95\% CIs were unadjusted. Complete Holm families and sensitivity specifications are in Supplementary Methods 7 and Supplementary Table 9k.

Three non-exclusive AI-coded markers identified explicit lung-cancer decisions, accompanying constraints and constraint-pathway links in existing EASS requirements. Marker scores descriptively regrouped the original rule ratings; they did not isolate link quality or add clinical ratings. Coverage, examples and composite-rule scores are in Supplementary Table 13.

\subsection{Format-aligned MCE output comparison}

For each of the 40 expert-reference cases, MCE generated one additional output using the five-part structure of the physician task. Each part was written as prose and constrained to the distribution of physician-response length; system-specific pathway labels were translated into concrete clinical actions. This output used clinical content formed within MCE and was not a post hoc summary of the full report by a general-purpose model. Format-aligned MCE outputs and U, R and M physician responses were each scored independently three times by Gemini 3.1 Pro Preview, GPT-5.5 and Claude Opus 4.7 using EASS. This common-case, common-format and common-rule comparison was a supporting analysis of within-case differences between system output and physician responses. It did not re-estimate the U/R/M condition effects from the mixed model for the physician task (Fig. 3c; Supplementary Methods 2; Supplementary Table 9e).

\subsection{Post hoc analysis of subsequent clinical events}

After completion of the primary offline evaluation, available follow-up records for the 40 expert-reference cases were first used to assess whether APAS condition differences persisted in cases with subsequently recorded composite clinical events. The same analysis described the content domains involved. Differences in the condition effect between event groups were secondary explorations. Subsequent outcome fields were not included in the fixed case inputs, EASS construction, MCE generation, physician task or text scoring.

Post hoc text rules identified explicit records of death, disease progression or suspected new metastasis, treatment interruption or non-initiation, and deterioration in health status. Cases were classified as having a record of any composite event, having follow-up without such a record, or having unavailable follow-up. This grouping was not independently clinically adjudicated and was not a standardized overall-survival, progression-free-survival or validated composite endpoint; absence of a record did not establish absence of progression or a favourable prognosis.

The subgroup analysis used case-specific U, R and M estimates derived from the APAS model fitted to all 40 cases and standardized with equal weight across each target case set. Group-specific estimates assessed whether the text difference was retained; event-group-by-condition contrasts explored effect modification. Removing a case from a target set did not refit the model. The main text reports a data-quality sensitivity analysis excluding three records with index-date or record-linkage concerns, with six cases lacking follow-up reported separately; all 34 cases with available follow-up were analysed in supplementary material. M1-M5 domains, death-record subgroups, target structures and primary clinical scenarios were unadjusted post hoc analyses (Supplementary Methods 3; Supplementary Tables 9f--h).

\subsection{Cross-model strategy-output comparison}

The cross-model comparison used the 40 expert-reference cases, each with an existing complete MCE output and one strategy output from GPT-5.4, Claude Opus 4.7 and Gemini 3.1 Pro, for 160 outputs. Four texts per case were presented under randomized A-D labels. Processing removed numeric citation markers but did not normalize length, section structure or style. Complete MCE outputs came from existing runs using the study configuration and versioned knowledge base but were not generated in the same run as the other-model outputs.

Gemini 3.1 Pro Preview, GPT-5.5 and Claude Opus 4.7 independently rated all 160 outputs against the same EASS rules. Each judge-output combination was rated once, producing 480 judge-output rating units. For each judge, APAS means and 95\% CIs and CSPR attainment out of 40 were reported by strategy. Supplementary post hoc two-sided paired t tests assessed complete-MCE-minus-other-output mean differences, with separate Holm families for nine overall APAS and 45 domain contrasts. Gemini M2 used 38 applicable case pairs; the other comparisons used 40. Ratings were not pooled across judges and did not use the nine-rating aggregation from the U/R/M branch. This supporting analysis described differences among complete outputs under fixed cases and requirements and did not attribute them to an individual MCE component (Fig. 4a,b; Supplementary Methods 4; Supplementary Table 9d).

\subsection{Source-masked content evaluation}

Each of the 60 cases formed two A/B modules within the case. The primary module compared MCE with MDT-Debate. The secondary module compared MCE with independent retrieval-augmented generation, Direct generation or the treating physician's original plan. Six clinical experts completed the first-round 120 modules. A separate panel of six experts reassessed 60 modules selected with stratification by case and comparator. Selection did not use first-round preference, domain ratings, safety flags, response time or length. Reviewers did not know the explicit source; source guesses were not collected.

One case-comparator module was the preference-analysis unit, decoded as MCE preferred, no difference or comparator preferred. Overall judgment and five clinical-content dimensions were reported separately. Modules assessed in both rounds were used for cross-tabulation, exact agreement, Gwet AC1 and auxiliary Cohen kappa. A safety flag was a signal for manual review and not a confirmed safety event. Allocation, masking, preference and reassessment agreement are reported in Supplementary Methods 4 and Supplementary Table 6; Extended Data Figure 4 shows the three-category distributions.

\subsection{Functional-module ablation across six MedGPT-based configurations}

Each of the 40 expert-reference cases generated one report under six configurations based on MedGPT. Three were targeted grouped functional ablations of complete MCE. The first removed the multidisciplinary-prior function at n0. The second removed scenario simulation, critique and candidate repair at n6-n8. The third jointly removed pathway and qualifying-evidence construction at n3-n5 and simulation, critique and repair at n6-n8. Both stage-removal configurations retained candidate-strategy reranking at n9; the joint n3-n8 ablation used configuration-limited retrieval. Case-state and goal representation at n1-n2 and stopping and report assembly at n10 remained in the MCE-derived ablation configurations. MedGPT direct generation and MedGPT-LightRAG provided reduced-orchestration reference configurations; the latter used the same knowledge base as MCE. One report was generated per case-configuration combination. The six reports were anonymously rotated and rated using EASS.

GPT-5.5 ratings were used for the principal numerical display. Gemini 3.1 Pro Preview and Claude Opus 4.7 ratings were used with GPT-5.5 only to compare the rank order of case-mean APAS across configurations; they were not pooled into a case-level score. Differences between complete MCE and each configuration were paired by case and reported as mean within-case differences with normal-approximation 95\% CIs. Supplementary post hoc two-sided paired t tests used Holm adjustment across all five contrasts. CSPR was descriptive. The contrasts evaluated functional bundles or reduced-orchestration reference configurations in fixed cases and realized single runs. The n3-n8 contrast was a joint ablation and did not isolate the n3-n5 pathway and evidence block. Output length and organization could change with each configuration, and the design did not estimate individual-node effects, interactions among functional blocks or run-to-run stability. Definitions and results are in Supplementary Methods 4 and Supplementary Table 7; case-level APAS means and intervals and the paired differences are shown in Figure 4c.

\subsection{Clinical anchoring, EASS rule re-rating and controlled text perturbations}

For overall clinical anchoring, existing outputs from the expert-reference set formed 40 case clusters. Each cluster contained one physician strategy, one complete MCE output and one comparator output, for 120 outputs. Two attending thoracic surgeons rated overall clinical acceptability and major clinical defects independently. Explicit source labels were removed while original content and structure were retained. Raters did not have access to source, configuration, condition, EASS rules, APAS or CSPR, sampling strata or the other rater's judgment.

Overall clinical acceptability was an output-level judgment of whether the complete strategy could enter clinical discussion and the extent of revision or additional verification required. The four ordered categories were unacceptable, major revision or additional verification required, broadly acceptable with minor revision, and acceptable for clinical discussion, with not assessable as a special state. Categories were coded 0-3; the mean code across raters was used for rank correlation with APAS. For binary descriptions, the upper two categories were acceptable and the lower two unacceptable. Raw judgments were retained without adjudication. Major clinical defects were reported separately and were not combined with acceptability into a new score. Sampling was balanced and enriched by source, prior score and risk or disagreement; source-stratified results were descriptive (Supplementary Methods 5; Extended Data Figure 2).

For EASS rule re-rating, two physicians independently rated the 240 rules corresponding to 40 complete MCE reports. System configuration, model name, AI rating and the other physician's rating were masked. Ratings were 0, 1, 2, not applicable or not assessable, and analysis used judgments before disagreement resolution. Case-level APAS was calculated only when the rater assigned 0, 1 or 2 to all six rules in that case. Physician ratings were not treated as a human gold standard (Supplementary Methods 5; Supplementary Tables 5, 9a and 9i).

Controlled text perturbations used 12 cases, 36 original-variant triplets and 108 texts. Each triplet included the original, a prespecified section-order variant and a variant deleting rule-linked clinical content. Reordering was intended to preserve clinical content. Deletions targeted M2, M3, M4 or M5, with nine triplets per domain. Physicians checked whether each candidate variant achieved its target and recorded changes outside the target before judging.

Each text was rated three times by each of the three named AI judges. Repeats were aggregated within text and were not independent samples. The triplet was the paired analysis unit and the case the clustering unit. Reordering and deletion were estimated separately and target-domain results were stratified. Texts included in analysis matched the physician-verified versions individually. In the absence of a prespecified equivalence margin, a non-significant reordering difference was not interpreted as equivalence (Supplementary Methods 5; Extended Data Figure 3).

\subsection{Case-condition changes and rater dependence}

Sixteen source-case templates were purposively selected from the expert-reference set. Each produced one original and two modified offline scenarios, for 48 scenarios. One modification changed a safety constraint or decision-critical information and the other changed management sequence, branching, reassessment or escalation. Six Condition-Change Response Rules (CCRRs) were specified for each scenario before its MCE output was reviewed, for 288 rules; rules were paired with reports only after generation.

Three AI judges classified each requirement as pass, clear gap, clinical review, not applicable or not assessable. A strict majority formed consensus; a three-way split was classified as clinical review; when one judgment was missing and the other two agreed, their result was retained and the missing judgment recorded. Requirement- and scenario-level results were reported separately. Complete scenario closure required pass status for every applicable requirement and no clear-gap, clinical-review or not-assessable state.

Two physicians independently rated the 48 scenario cards and 288 requirements before disagreement handling, without seeing the AI consensus or governance mapping. There was no third-rater adjudication and no human gold standard. Agreement between physicians and between each physician and the AI consensus was reported at the requirement and scenario-card levels using exact agreement and Gwet AC1. Intervals were obtained by cluster resampling of the 16 source templates. The records did not establish that the original and changed scenarios used identical model snapshots, prompts, adapters, replicate numbers and random conditions. This branch therefore describes relationship closure and evaluator dependence in purposive case variants; it does not estimate a causal effect of changing case conditions, an overall error rate or real-world safety (Supplementary Methods 6; Extended Data Figures 6--8; Supplementary Table 8).

\subsection{AI judges and rating aggregation}

The fixed judging panel comprised Gemini 3.1 Pro Preview, GPT-5.5 and Claude Opus 4.7. Repetition and aggregation were prespecified separately for each branch. Each judge rated U/R/M physician responses and format-aligned MCE outputs three times. For APAS, response-rule scores were averaged within the three repeats from a judge and then averaged equally across judges. For CSPR, a majority was first formed within judge across repeats and then across judges; absence of a complete majority was indeterminate. Controlled text perturbations also used three ratings per judge but derived APAS and partial gating states from the six selected rules in each request. The cross-model strategy-output comparison, relationship audit of complete MCE outputs and case-condition evaluation used their branch-specific independent judgments. The six-configuration comparison used GPT-5.5 for primary numerical ratings and the other two judges only for configuration-level rank stability. Repeated judgments were measurements of the same object and did not increase the number of physicians, cases, responses or outputs. Model identities, inputs, repeats, parsing and aggregation are detailed in the relevant Supplementary Methods and Supplementary Tables 3 and 5.

\subsection{Statistical analysis}

For the physician task, one response by one physician to one case under one assigned condition was the analysis unit. APAS was analysed with a linear mixed-effects model containing case and question order as categorical fixed effects, physician as a random intercept, and condition differences allowed to vary across fixed cases. Primary comparisons were standardized to an empirical reference grid of the 40 cases and nine balanced question positions, with the physician random effect set to zero. The estimate describes model-adjusted condition differences in the fixed case set; cases were not weighted by clinical frequency or importance, and the interval did not incorporate sampling uncertainty from case selection.

CSPR was analysed using the prespecified common-condition-effect generalized linear mixed model that was ultimately used. Fixed-effect predictions were transformed to probabilities and averaged across the case and question-order reference grid with the physician random effect set to zero rather than integrated over its distribution. The resulting probability difference is not a population-average risk difference across physicians. CSPR was secondary or exploratory.

Unassigned case-condition cells were design absences and were not imputed. Content-domain, professional-title-group interaction, leave-one-judge-out, common-applicability, system-label removal, length and presentation analyses were identified as prespecified or post hoc. Response length occurred after condition assignment and was not adjusted for as a baseline covariate. Subsequent-event analyses used case-specific condition estimates from the APAS model fitted to all 40 cases and equal-weight standardization within target case sets. The main text reports the 31-case data-quality sensitivity analysis after exclusion of three questionable records; all 34 follow-up-available cases are reported separately, and six cases with unavailable follow-up are retained as a separate category. Format-aligned MCE and physician ratings were averaged equally across nine judge ratings and then summarized as system scores and U, R and M physician case means for each of 40 cases. Signed differences in APAS, M1-M5 and character count were paired by case, with 95\% CIs from 30,000 case-cluster percentile bootstrap samples. This supportive paired estimand described system-output-minus-physician-response differences within fixed cases; the primary U/R/M condition effects remained the mixed-model estimands accounting for repeated physician responses, case heterogeneity and task position.

Primary APAS and CSPR comparisons used separate Holm corrections. Supplementary post hoc two-sided paired t tests assessed zero mean differences in the format-aligned, cross-model and functional-configuration branches using the original case pairs. Four separate complete Holm families covered 21 format-aligned contrasts (three overall APAS, 15 domain and three length), nine cross-model overall APAS contrasts, 45 cross-model domain contrasts and five configuration contrasts. Original point estimates and unadjusted 95\% CIs were retained: case-cluster percentile bootstrap intervals for format-aligned comparisons and normal-approximation intervals for cross-model and configuration comparisons. The P values used t distributions with case-pair degrees of freedom, not inversion of these intervals. Existing physician-task and hospital/category model tests and correction families were unchanged. Subsequent-event, death, target-structure and case-scenario strata were unadjusted post hoc analyses. Effect estimates and 95\% CIs were reported with convergence, singular-fit and weak-identification diagnostics. Alternative judges and repeated runs were measurement-sensitivity analyses and did not add clinical samples. Full models, diagnostics and sensitivity analyses are in Supplementary Methods 7 and Supplementary Table 9; CSPR and professional-title-group analyses are in Extended Data Figure 1.

Overall clinical anchoring used 120 outputs in 40 case clusters. Raw five-category judgments were summarized with exact agreement and Cohen kappa; the four-level ordinal outcome also used linearly weighted kappa. Spearman's rank correlation related APAS to the mean ordinal judgment of the two physicians, with intervals from case-cluster resampling. EASS rule re-rating reported exact agreement and linearly weighted Gwet AC2 among paired assessable rules. Agreement in APAS at the case level used ICC(A,1), Lin's concordance correlation coefficient, the mean rater difference and Bland-Altman 95\% limits of agreement. Controlled perturbations were paired within triplet with case-cluster bootstrap. Main-text proportions for case-condition changes used original requirement counts; template-equal estimates first summarized within source template and then weighted the 16 templates equally, with clustering by template. Agreement between physicians and between each physician and the AI consensus used exact agreement and Gwet AC1 at the rule and card levels. The 288 rules were not treated as 288 independent clinical samples.

\subsection{Reproducibility}

Each analysis linked the case or text input to the generated output, rating request, raw response, rule aggregation, analysis data and figures. For generative-model work, recalculation from stored responses was distinguished from a new call to an external model, which can change with model updates and stochastic generation and is not expected to reproduce wording exactly. Input versions, software environments and random seeds were recorded for derived analyses.

Figure 2 presents one traceable run in which the case input, system version, intermediate objects and final report were linked. It illustrates formation and retention of strategy objects in one research run and does not estimate a cohort-wide completion rate or patient outcome. The reviewable scope is described in Supplementary Methods 2.

\subsection{Study governance and generative-model reporting}

Internal clinical cases were de-identified and access was restricted. The scope of data transmitted to external models, data retention and access controls was managed in accordance with the approved study protocol and institutional data-governance requirements.

Generative models used for system generation, comparator generation and AI judging were documented separately by role. Reporting included model identifiers and providers, access dates, inputs and system instructions, retrieval corpora and evidence cutoffs, decoding parameters, repetitions and retries, output processing, anonymization and human review. Reporting followed TRIPOD-LLM\cite{ref35}, with the scope of early clinical evaluation informed by DECIDE-AI\cite{ref31}.

\section{Data and code availability}

Data and code associated with this article will be made publicly available through the \href{https://github.com/Medlinker-MG/MCE}{Medlinker-MG/MCE GitHub repository}. The release will include de-identified aggregate interfaces, data dictionaries, manifests, checksums and analysis code for the reported statistics and figures, together with de-identified and disclosure-reviewed case-level EASS rules, original physician responses, complete model outputs, and the generation and scoring prompts retained during the study. System or user instructions that were not completely retained for the cross-model generation branch will be identified in the release manifest. Original clinical narratives from unpublished internal cases, identifying information, and case-level follow-up or outcome records will not be released. Sources for published cases remain identified in the Supplementary Information.

\section{Ethics statement}

The study protocol and the use of clinical cases were approved by the Ethics Review Committee of Peking Union Medical College Hospital, Chinese Academy of Medical Sciences (approval no. I-26PJ2472).

\section{Generative-AI transparency}

Generative AI was used for language editing during manuscript preparation. All authors reviewed, revised and approved the final content and take responsibility for the manuscript.

\section{Acknowledgements}

We thank all physicians who contributed their time and clinical expertise to this study. Their participation and thoughtful input were essential to this work.

\section{Funding}

This work was supported by the Peking Union Medical College Hospital Talent Cultivation Program (Category B; No. UGG10521) and the National High Level Hospital Clinical Research Funding (No. 2025-PUMCH-A-035).

\section{Author contributions}

D.W., Zhicheng Huang (Z.H.1), H.S., J.X., X.X., L.Z., Shirui Wang (S.W.1) and N.L. conceptualized the study. D.W., Z.H.1, H.S., J.X. and X.X. performed the data analysis and drafted the manuscript. D.W., Z.H.1, Z.Z., Z.B., Yuxiao Lin (Y.L.1), Yicheng Liang (Y.L.2), C.G., B.X., K.Z. and N.L. developed the case collection criteria, collected the clinical cases, provided clinical knowledge and recommendations for system development, and conducted clinician re-evaluations. J.X., S.X., W.Y., H.X., L.L., X.Y., M.H., Q.M., Z.X., X.L., Zhihai Han (Z.H.2), N.Z., C.T., T.Z., L.S., C.Z., X.Z., Shafei Wu (S.W.2), H.G. and L.D. developed the scoring strategy, performed blinded assessments and reviewed the clinician re-evaluations. H.S., Z.L., W.S., Y.W., X.S., B.W., T.G., Y.C. and S.W.1 developed the system and provided technical support. D.W., Z.H.1, H.S., J.X., X.X., L.Z., S.W.1 and N.L. critically reviewed and revised the manuscript. All authors read and approved the final manuscript.

\section{Competing interests}

H.S., Z.L., W.S., Y.W., X.S., B.W., T.G., Y.C. and S.W.1 are employees of Medlinker Intelligent and Digital Technology Co., Ltd., Beijing, China, which developed the MCE system and the MedGPT model evaluated in this study. The clinical evaluation procedures used to compare model and system performance, including development of the scoring strategy, blinded assessments and review of clinician re-evaluations, were conducted independently by clinical experts who were not employed by Medlinker; the Medlinker-employed authors did not participate in these activities. Their respective roles in study conceptualization, data analysis, manuscript preparation, system development and technical support are described in the Author Contributions statement. The remaining authors declare no competing interests.

\clearpage
\section*{Main Figures}

\phantomsection
\label{fig:main-1}
\begin{center}
\centering
\includegraphics[width=\textwidth,height=0.68\textheight,keepaspectratio]{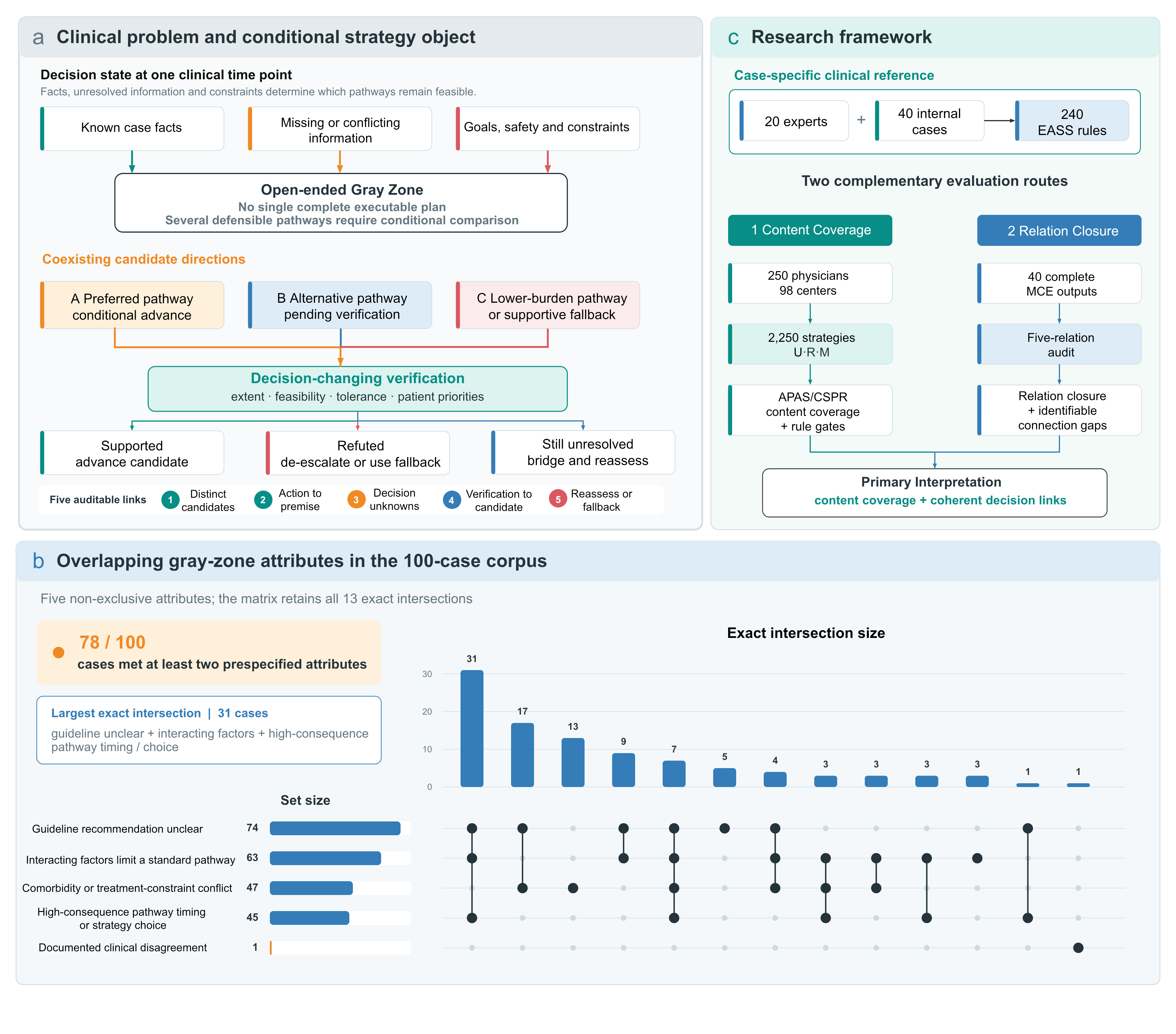}
\end{center}
{\small\noindent\textbf{Figure 1 | Conditional decision object, overlapping gray-zone attributes and principal evidence routes} \textbf{a.} Components and relations of the conditional strategy object. Candidate directions with different benefit, risk and implementation burdens enter review together. Premises such as disease extent, treatment tolerance and patient goals determine whether candidates are supported, refuted or remain unresolved and connect them to advancement, de-escalation or reassessment. \textbf{b.} Set sizes and 13 exact intersections for five non-exclusive gray-zone attributes among 100 purposively selected complex lung cancer cases. The attributes were unclear guideline recommendations, interacting factors, conflict from comorbidity or treatment constraints, pathway timing or strategy choice with substantial potential consequences, and recorded real clinical disagreement; 78 cases met at least two attributes. The intersections describe co-occurring constraints and are not a validated scale of complexity, severity, prognosis or support need. Full definitions and study-set counts are in Table 1 and Supplementary Table 10. \textbf{c.} The principal evidence chain comprised two complementary routes. Twenty specialists, all at chief-physician rank, across six disciplines constructed 240 EASS rules from 40 unpublished internal cases; 250 physicians from 98 centers submitted 2,250 final strategies under U, R and M to assess case-level clinical-content expression and the joint rule gate. Separately, 40 complete MCE outputs were assessed for closure among candidates, premises, decision-changing unknowns, verification tasks and reassessment or fallback. Source-masked content evaluation, clinical anchoring, controlled text perturbations and case-condition changes were supporting or diagnostic branches reported in Extended Data and the Supplementary Information.\par}

\clearpage
\phantomsection
\label{fig:main-2}
\begin{center}
\centering
\includegraphics[width=\textwidth,height=0.68\textheight,keepaspectratio]{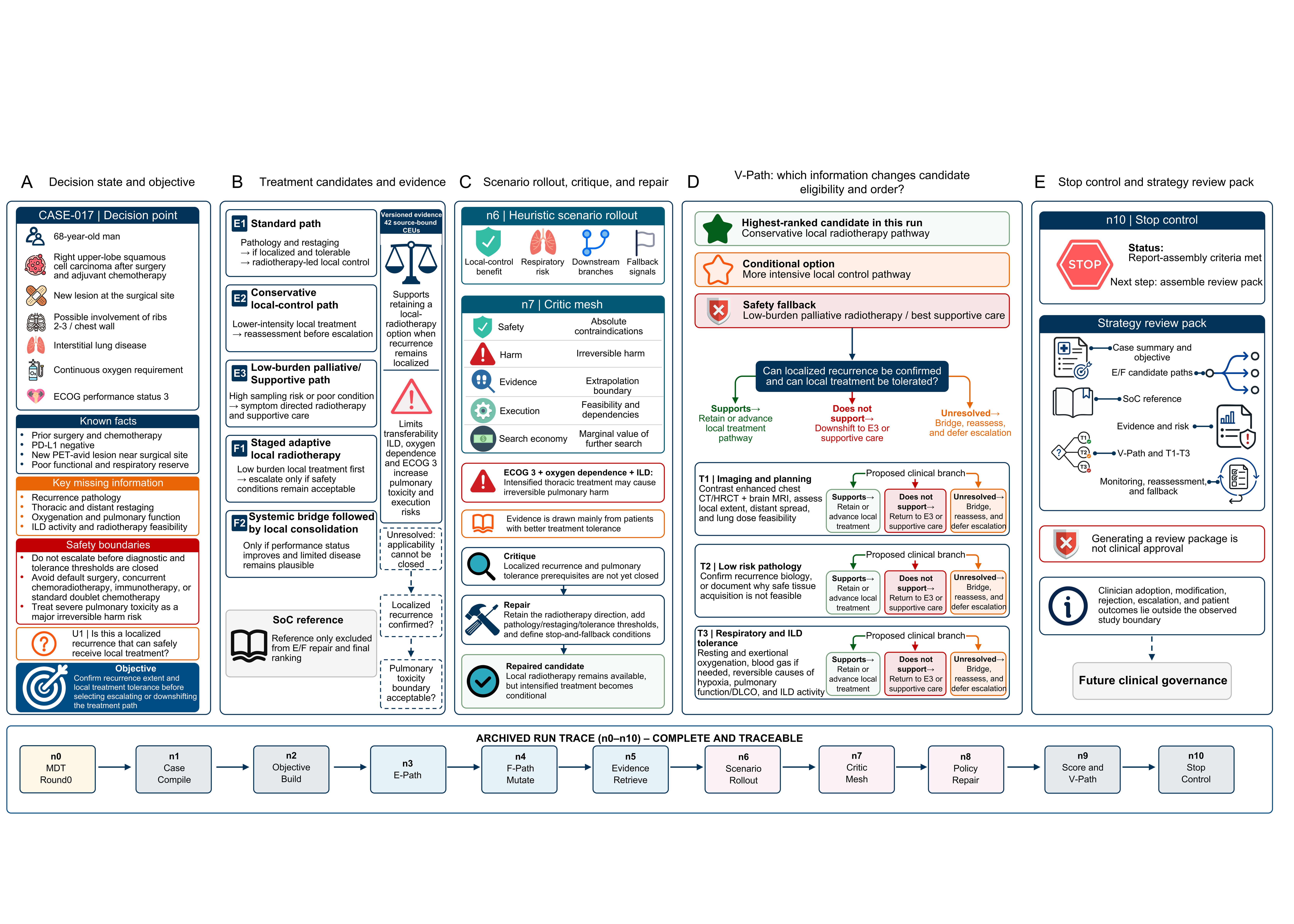}
\end{center}
{\small\noindent\textbf{Figure 2 | MCE links treatment candidates to decision-changing verification in an illustrative run} One archived complete MCE run illustrates the trade-off between local control and irreversible pulmonary injury. \textbf{a.} Previous treatment, suspected recurrence, respiratory reserve and pulmonary-toxicity constraints define the decision; recurrence extent and treatment tolerance remain unresolved. \textbf{b.} Three E-Path and two F-Path candidates, a separate SoC reference and 42 source-linked CEUs organize local-treatment feasibility and toxicity boundaries. \textbf{c.} Critique and repair retain local radiotherapy with intensity conditional on verification of pathology and tolerance. \textbf{d.} Imaging, planning, pathological confirmation and respiratory assessment connect supporting, refuting or unresolved findings to advancement, de-escalation, bridging and reassessment. \textbf{e.} Internal stopping criteria trigger assembly of the Strategy Review Pack containing candidates, evidence, risks, verification tasks and fallback. ``Current preferred direction'' is the system's highest-ranked candidate in this run. ``Convergence'' denotes report-assembly stopping; ``COMPLETE, REPRODUCIBLE RUN'' denotes the traceable archived object and n0--n10 chain, not identical outputs from new calls. Investigations and treatments remain proposals for clinical review.\par}

\clearpage
\phantomsection
\label{fig:main-3}
\begin{center}
\centering
\includegraphics[width=\textwidth,height=0.68\textheight,keepaspectratio]{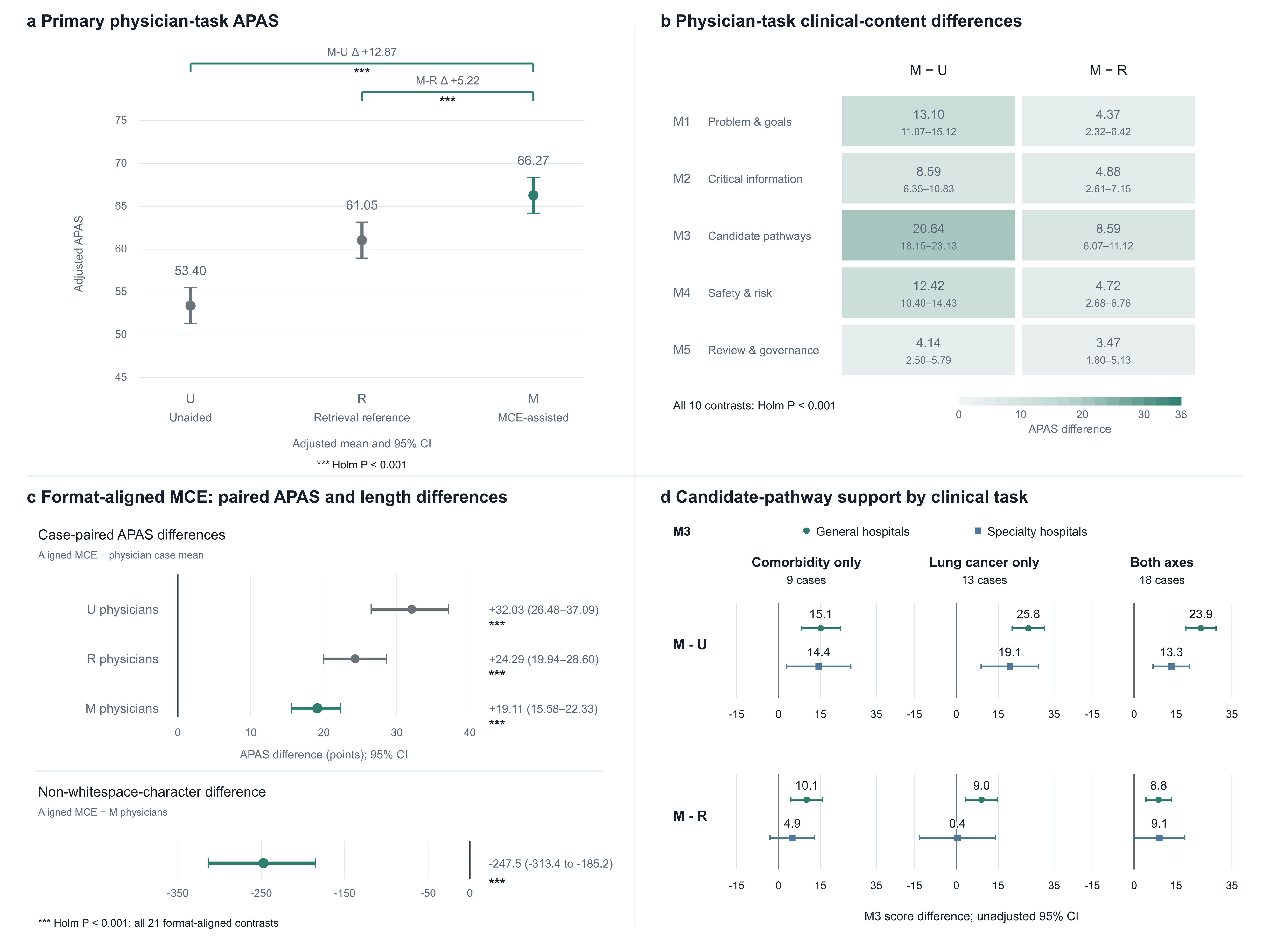}
\end{center}
{\small\noindent\textbf{Figure 3 | MCE support and expression of case-level clinical requirements in physicians' final strategies} \textbf{a.} Adjusted APAS means and 95\% CIs under U, R and M for 2,250 responses from 250 physicians. Brackets show M--U, 12.87 points (95\% CI, 11.18 to 14.55), and M--R, 5.22 (3.52 to 6.93); \texttt{***} denotes Holm P < 0.001 across these two model-based comparisons, with precise values in Supplementary Table 9a. Mixed-model estimates account for physician repeats and case structure and standardize to 40 cases and balanced task positions. \textbf{b.} M--U and M--R differences for M1 problem/goals, M2 decision-critical information, M3 candidate pathways, M4 safety/risk and M5 reassessment/governance. Cells display estimates and 95\% CIs; color uses a common 0--36 point scale. All ten contrasts have Holm P < 0.001; complete values are in Supplementary Table 9b. \textbf{c.} Case-paired APAS differences for format-aligned MCE minus physician means under U, R and M, with a separate character-length difference versus M. Forty cases are equally weighted; nine ratings are measurement repeats. CIs use 30,000 case-cluster bootstrap samples. Row-level \texttt{***} denotes Holm P < 0.001 for supplementary two-sided paired t tests across all 21 format-aligned contrasts. This system-output comparison differs from the physician-condition estimand in a. All format-aligned domain differences and precise P values are in Supplementary Table 9e. \textbf{d.} Exploratory M3 contrasts using 2,133 responses from 237 physicians at 91 institutional codes. Columns show comorbidity-complex only (9 cases), lung-cancer-complex only (13) and complex on both axes (18); rows show M--U and M--R. Green circles denote general hospitals and blue squares denote cancer or thoracic/pulmonary-specialty hospitals. Labels show point estimates; horizontal intervals show unadjusted 95\% CIs on a common scale. Estimates use category-specific case weighting, balanced positions, common professional-title standardization and institutional-code CR2 inference. Categories are unordered; a single-axis category means only that axis was labeled complex, not absence of the other disease. Within the complete 60-test family, four of six M--U and two of six M--R contrasts shown have Holm P < 0.05; individual P values are reported in Supplementary Table 9k. Complete five-domain patterns and direct hospital differences are in Extended Data Figure 9. Professional-title and CSPR results are in Extended Data Figure 1.\par}

\clearpage
\phantomsection
\label{fig:main-4}
\begin{center}
\centering
\includegraphics[width=\textwidth,height=0.68\textheight,keepaspectratio]{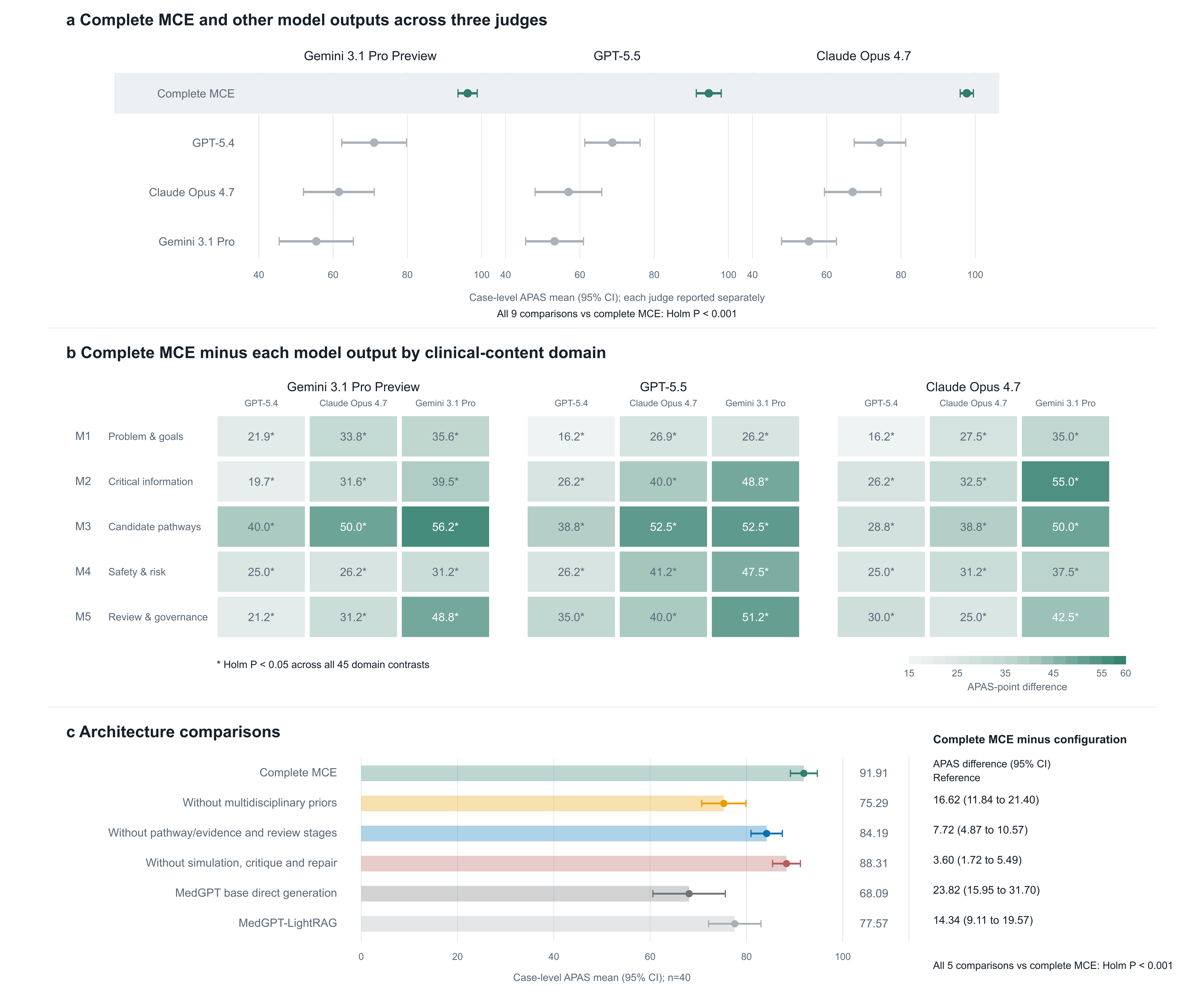}
\end{center}
{\fontsize{9.5}{11.5}\selectfont\noindent\textbf{Figure 4 | Conditional clinical-content increment of complete MCE outputs and support from the functional architecture} \textbf{a.} Case-level APAS means and 95\% CIs for complete MCE and strategy outputs generated by GPT-5.4, Claude Opus 4.7 and Gemini 3.1 Pro in the same 40 expert-reference cases. The three subplots correspond to the Gemini 3.1 Pro Preview, GPT-5.5 and Claude Opus 4.7 judges; results are reported separately and are not pooled across judges. All nine comparisons with complete MCE have Holm P < 0.001 for supplementary two-sided paired t tests across the complete nine-contrast family. Precise P values are in Supplementary Table 9d. \textbf{b.} Within-case APAS differences between complete MCE and the three other-model outputs for M1--M5, shown separately for each judge. Color and cell values encode point estimates; * denotes Holm P < 0.05 across the complete 45-domain-contrast family. Complete 95\% CIs and exact P values are in Supplementary Table 9d. For the Gemini 3.1 Pro Preview judge, all M2 absolute estimates and the three within-case contrasts use the 38 cases for which M2 was applicable; other domain estimates use 40 cases. M3 was the largest descriptive estimate in seven of nine model--judge combinations. \textbf{c.} Case-level APAS means and 95\% CIs for complete MCE, three grouped functional ablations and two reduced-orchestration reference configurations under GPT-5.5 scoring, with 40 cases per configuration. Horizontal bars use the complete 0--100 scale, and points and error bars show means and 95\% CIs. The right-hand column gives complete-MCE-minus-configuration paired differences and original 95\% CIs; all five comparisons with complete MCE have Holm P < 0.001 across the complete five-contrast family, with precise P values in Supplementary Table 7. Supplementary post hoc paired tests in a--c assess zero mean case differences; original normal-approximation intervals remain unadjusted. Panels a,b and c use independent output sets and rating batches; complete MCE means are not compared across branches. Within-case contrasts and judge-specific configuration rankings are in the text and Supplementary Table 7. Grouped ablations do not identify individual-node effects, functional-block interactions or the independent contribution of n3--n5. MedGPT base direct generation in this branch is a reduced-orchestration reference and is distinct from the GPT-5.4 output in panels a,b. Source-masked preference and dimension-specific re-review agreement are shown in Extended Data Figure 4.\par}

\clearpage
\phantomsection
\label{fig:main-5}
\begin{center}
\centering
\includegraphics[width=\textwidth,height=0.68\textheight,keepaspectratio]{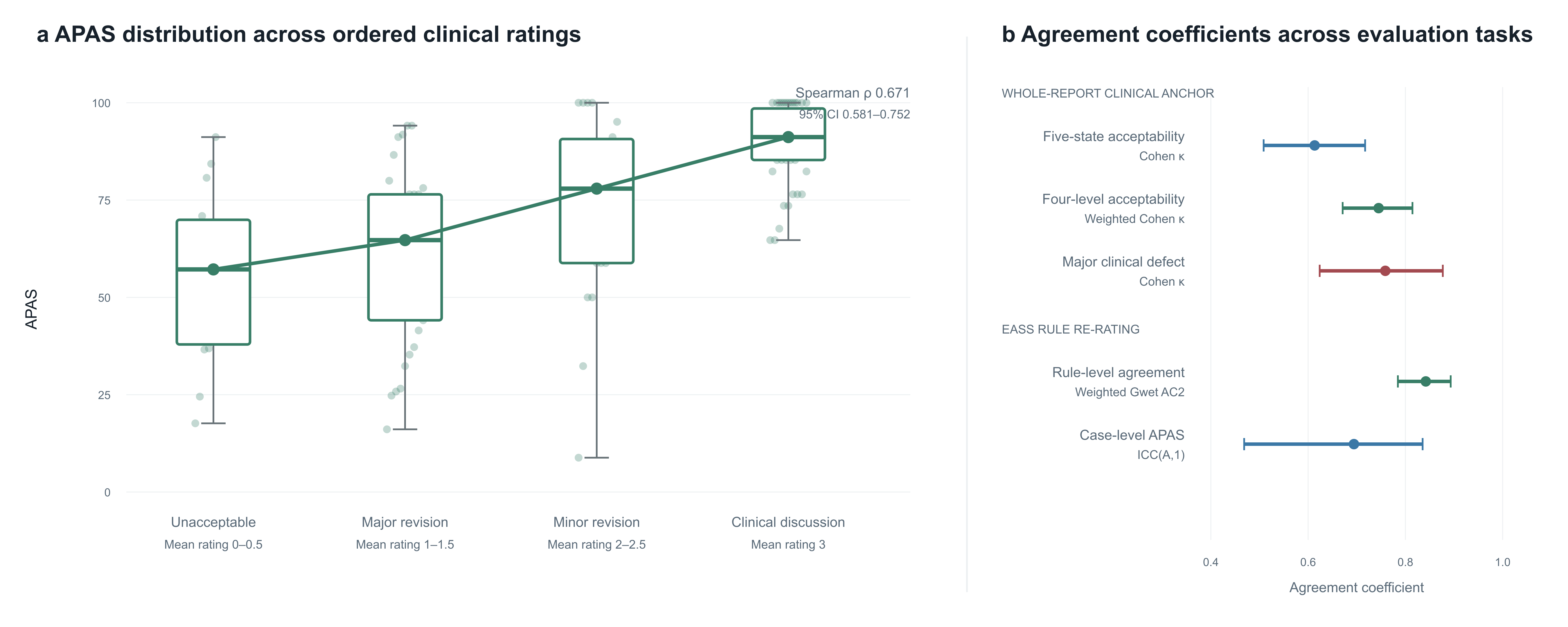}
\end{center}
{\small\noindent\textbf{Figure 5 | Clinical interpretation basis for the principal APAS results} \textbf{a.} APAS distributions for 120 outputs grouped by the mean four-category ordered overall clinical-acceptability judgment of two physicians. Original mean scores of 0, 0.5, 1, 1.5, 2, 2.5 and 3 are grouped as 0--0.5, 1--1.5, 2--2.5 and 3. Points are outputs; boxes show interquartile ranges, horizontal lines and connected points are medians, and whiskers show observed ranges. Spearman's rank correlation used the ungrouped mean score across all 120 outputs (\ensuremath{\rho} = 0.671; 95\% CI, 0.581 to 0.752), with interval estimation by resampling 40 case clusters. \textbf{b.} Reliability coefficients and 95\% CIs for whole-text clinical anchoring and EASS rule re-rating on a common 0.4--1.0 axis: five-category Cohen kappa for overall acceptability, linearly weighted Cohen kappa for the four-category ordered result, Cohen kappa for major clinical defects, linearly weighted Gwet AC2 at the rule level and ICC(A,1) at the case level. Exact agreement, Lin concordance correlation coefficient, case-level mean difference and limits of agreement are reported in the text and Supplementary Table 9i. Complete anchoring distributions and source-stratified descriptions are in Extended Data Figure 2. All controlled-text triplet changes are shown in Extended Data Figure 3 and Supplementary Table 9j.\par}

\clearpage
\appendix
\part*{Supplementary Information}
\addcontentsline{toc}{part}{Supplementary Information}
These methods describe evaluation of MCE support for physicians' strategy formulation. They cover construction of the case-specific clinical reference, MCE generation and the physician task, supportive system-output comparisons, clinical interpretation of APAS, pathway coherence in case variants, and statistical implementation. The principal abbreviations are Expert-Admissible Strategy Space (EASS), Admissible Pathway Attainment Score (APAS), Complete-and-Safe Pathway Rate (CSPR), and Condition-Change Response Rule (CCRR).

\section{Supplementary Methods 1. EASS expert process and operationalization of the case-specific clinical reference}

\subsection{Expert process}

The expert panel comprised 20 chief physicians from 15 institutions and six disciplines: thoracic surgery, respiratory medicine, medical oncology, radiology, pathology and radiation oncology. Nineteen participated in the first in-person round; the remaining member joined later online rounds. Fourteen treatment-focused panelists undertook the subsequent case review. In round 1, experts independently identified each case's clinical problem, treatment goals, decision points, information gaps, specialty boundaries and safety issues requiring verification. The resulting content was organized into five clinical content domains: M1, clinical problem and treatment goals; M2, decision-critical information; M3, candidate clinical pathways; M4, safety boundaries and risk governance; and M5, reassessment and governance.

In round 2, 22 key or high-risk cases were each reviewed by four experts and the other 18 cases by three experts, yielding 142 expert--case review units. A strict majority required at least two votes in three-person reviews and at least three votes in four-person reviews. Items proceeded to round 3 when the direction of judgment, relevant clinical conditions or scoring boundary remained unresolved, including absence of a strict majority, disagreement about safety, an incomplete pathway rationale, free text that could alter the clinical judgment, or an unresolved scoring denominator.

In round 3, three experts independently answered each item requiring clarification, and the direction was determined by a strict majority of valid responses. Across 116 unresolved items, round 3 yielded 348 valid responses. Clinical prerequisites recorded with conditionally accepted votes were incorporated into the final rule. A 1:1:1 split for decision-critical information or governance items was handled conservatively by requiring verification, permitting lower-risk parallel actions, and specifying reassessment or multidisciplinary-team review. Reasonable disagreement was recorded only when more than one strategy had a clinical basis and could remain admissible concurrently. The final EASS was converted into an executable case-specific reference with 0/1/2 scoring rules. The 40 final M4 safety rules were based primarily on the round-2 case cards, safety-boundary table and safety review; 30 safety items carried forward the round-2 expert judgment, with pathway-applicability conditions from round 3 added where relevant. These safety boundaries and the other domain rules entered both APAS and CSPR evaluation.

EASS was constructed independently from the fixed case materials and expert review. The MCE, comparator and physician outputs to be evaluated were not shown to the experts constructing EASS, and EASS rules were not used to generate those outputs. The expert process and item flow are summarized in Supplementary Table 1.

\subsection{Rule data structure}

The final EASS contained 240 case-specific rules, six per case. Each rule recorded a rule identifier, case identifier, clinical content domain, final state, clinical requirement, applicability conditions, acceptable actions, prohibited actions, fallback actions, weight, 0/1/2 anchors, links to CSPR gates, failure codes and a version key. The fixed rule dictionary, with historical technical version key \texttt{eapg-final-20260731}, was used throughout the reported scoring; the key is retained for linkage to immutable source records and derived analyses and is not a current concept label. M1--M5 define the clinical content to be addressed in a given case; they are distinct from the five structural relations used to audit whether candidates, prerequisites, unknowns, verification tasks and subsequent actions were connected. The former provide the content reference and the latter assess relational structure.

Final rules were classified as confirmed acceptable or conditionally acceptable. Prohibited circumstances, fallback pathways, reasonable disagreement and unresolved content were recorded as clinical boundaries. A conditionally accepted action was compliant only if its specified prerequisite was met; use of a restricted pathway without its prerequisite scored 0. Correct entry into a prespecified fallback pathway, reasonable disagreement within the admissible strategy space and currently non-applicable items were excluded from both the APAS numerator and denominator rather than scored as failures.

\subsection{APAS and CSPR rules}

Each applicable EASS rule was scored 0, 1 or 2. M1--M4 had weight 3 and M5 had weight 2. For response \(s\), APAS was

\[
\mathrm{APAS}(s)=100\times\frac{\sum_i w_i r_i(s)}{2\sum_i w_i},
\]

where the sum included only applicable rules entering the denominator. The clinical problem and treatment goal within M1 were scored separately; M2--M5 each contributed a case-level rule. APAS measures the extent to which a strategy expresses the applicable case-level clinical requirements.

Let \(G_s\) denote the set of applicable CSPR gates for response \(s\), with \(z_g(s)=1\) for a passing gate and \(z_g(s)=0\) for a failed gate. When a complete judge majority was available for every applicable gate,

\[
\mathrm{CSPR}(s)=\prod_{g\in G_s} z_g(s).
\]

Thus, failure of any applicable gate yielded CSPR failure. If any applicable gate lacked a complete majority, the response remained indeterminate rather than being forced to 0. APAS and CSPR were derived from the same case-specific reference but calculated separately. The word \emph{safe} in CSPR refers to satisfaction of prespecified textual scoring constraints, not measured clinical safety. The nine gates and their applicability, failure and denominator rules are specified in Supplementary Table 2; judge-level and cross-judge aggregation are described in Supplementary Methods 3.

\section{Supplementary Methods 2. Generation of MCE and comparison outputs}

\subsection{MCE input, processing and output}

The case-level input was a fixed Case Pack containing six fields: basic information, chief complaint, present illness, past and other medical history, investigations, and diagnosis. It excluded management, the treating physician's recommendations, subsequent plans and outcomes. Internal cases were managed as limited, de-identified restricted data. Neither EASS rules nor CCRRs entered generation.

MCE used MedGPT as its base model and organized processing around four scientific functions: representation of case state, goals and constraints; organization of candidate pathways and qualifying evidence; feasibility checks, clinical critique and repair; and management of uncertainty, stopping assessment and assembly of a Strategy Review Pack. The final report contained clinical goals, decision-critical information, candidate pathways, conditions under which pathways became admissible or failed, evidence and risk, monitoring, reassessment and governance arrangements.

Study runs used a version-controlled knowledge base and evidence corpus. For clinical disputes that could change candidate eligibility, sequence or intensity, MCE organized evidence by proposition type into source-linked clinical evidence units (CEUs) and recorded whether each source supported, limited or left candidate applicability unresolved. Evidence types were not collapsed into a single score, and insufficient evidence remained unresolved. The CEUs supported review at the proposition-and-source level. Complete PICO representation, formal applicability boundaries, evidence grading and calibrated confidence were not evaluated in this study.

The Strategy Review Pack was intended to help physicians compare reasonable treatment pathways, identify prerequisites for proceeding, connect patient constraints to eligibility, intensity and sequence, and review subsequent actions when information or risks changed. Its governance content identified questions for specialist input or collective judgment. Physicians or a formal multidisciplinary team retained responsibility for modification, rejection, adoption and escalation; these decisions remained outside the MCE system boundary.

\subsection{Representative run}

An illustrative complete MCE run, completed on 27 July 2026, was selected to show how a conditional strategy was formed and retained. It illustrated the trade-off between local control and irreversible pulmonary injury described in the main text. The version-bound archived run linked the case input and study condition with system and prompt versions, input and output hashes, execution stages, 42 source-linked clinical evidence units (CEUs), intermediate candidates, structured review, candidate repair, verification tasks, the stopping decision and the final report. These linked objects support the functional sequence shown in Figure 2.

The unit of analysis for this example was one system run, not a patient outcome or clinical-workflow event. The example documents the method and object traceability; it does not estimate run completeness across the cohort or imply that a new call to an external model would reproduce identical wording.

\subsection{Audit of five structural relations in complete MCE outputs}

Forty complete MCE Strategy Review Packs were assessed for five relations: candidate-pathway differentiation; the action-premise relation connecting an action to conditions for initiation, restriction or withdrawal; a decision-changing unknown capable of altering pathway eligibility, sequence or intensity; the verification-candidate relation connecting a verification result to a specific candidate and subsequent action; and the reassessment/fallback relation retained when the preferred pathway became infeasible, risk increased or information remained insufficient. Each dimension was coded as complete, partly complete, missing or incorrect, not applicable, or not assessable.

Gemini 3.1 Pro Preview, GPT-5.5 and Claude Opus 4.7 independently evaluated the same 40 outputs. Each judge--output combination was assessed once; results were reported by judge without voting or averaging and without the nine-run aggregation used for the U/R/M task. Overall relationship closure required every applicable dimension to be complete, no critical omission or relationship conflict, and every applicable verification task to specify subsequent action after a supportive, refuting or still-unresolved result, or an explicit reason why a state was not applicable. Defects were separately coded as omission, misconnection, internal conflict or not assessable. Verification tasks were observations nested within outputs; neither judge count nor task count increased the number of outputs or clinical samples.

Targeted human checking covered 18 identified objects: ten still-unresolved branches and eight objects requiring additional judgment rationale. Review of the corresponding outputs located the relevant text and confirmed subsequent arrangements in nine of the ten unresolved branches and one omission. This review resolved the identified trace questions without replacing the original dimension ratings from the three AI judges. Full definitions, judge-specific summaries and review counts are reported in Supplementary Table 11 and Source Data.

\subsection{U/R/M conditions and other comparators}

The unaided condition (U) displayed only the standardized Case Pack. The retrieval-reference condition (R) additionally displayed case-specific reference information generated by GPT-5.4 with LightRAG using the same knowledge base as MCE. Sharing a knowledge base did not imply use of the same retriever, index, reranker, prompt, input budget, orchestration or output structure. The MCE-assisted condition (M) displayed the Strategy Review Pack for review before physicians composed their own final strategy. Physicians could question, revise or supplement the support content; item-level viewing, adoption, revision and rejection were not recorded.

Comparators in the source-masked content evaluation were Direct, IR-RAG, MDT-Debate and Clinician Plan. MedGPT base direct generation and MedGPT-LightRAG in the separate six-configuration comparison were different system configurations and were not interchangeable with these source-masked comparators. Inputs, knowledge relationships, outputs and evaluation purposes are summarized in Supplementary Table 3.

\subsection{Format-aligned MCE output}

To examine whether report length and presentation structure accounted for differences in rule coverage, the MCE system generated one additional format-aligned output for each of the same 40 cases. This output retained the clinical content formed within MCE but presented it in the same five-part structure as the physician task, using paragraphs and a length constraint based on the physician-response distribution. Internal pathway labels were rendered as concrete clinical actions. The format-aligned output was not a post hoc summary or rewrite of the complete report by a general-purpose AI model.

The format-aligned MCE outputs and the physician responses under U, R and M were each scored three times by Gemini 3.1 Pro Preview, GPT-5.5 and Claude Opus 4.7. APAS and M1--M5 scores were averaged equally across the nine measurements at the response level. Physician responses were then averaged separately within each case and each condition so that all 40 cases contributed equally to paired comparisons. Non-whitespace characters were counted. Figure 3c displays signed APAS differences between the format-aligned output and the U, R and M physician case means, together with the character-count difference relative to M. Supplementary Table 9e retains the domain estimates and the complete paired APAS, M1--M5 and character-count contrasts against each physician condition. CIs were obtained from 30,000 case-cluster percentile bootstrap samples with seed 20260907, retaining the system output, physician responses and all ratings from the resampled case. This supporting estimand compared a system output with the physician mean for the same case and support condition. It differed from the physician-condition effects obtained from the mixed model incorporating physician random intercepts, case-by-condition terms and task position, and did not measure whether physicians viewed, adopted, changed or rejected particular content.

\subsection{Output processing and reviewable objects}

For source-masked content evaluation, candidate strategies were placed in the same display structure and explicit source labels were removed, but body text was not normalized for length or style. For the six-configuration comparison, identifying evidence numbers were removed and labels were rotated anonymously, again without rewriting the output length or style. Source masking therefore could not prevent reviewers from inferring source from length, format or content.

Generated, comparator and judge outputs were recorded separately by model or functional role, case input, evaluation input and evaluation purpose. Fixed case inputs, preserved outputs, scoring rules, scoring requests, panel aggregation and statistical analyses constituted the reviewable analytic objects.

\section{Supplementary Methods 3. Physician-task allocation, presentation and strategy scoring}

\subsection{Assignment-list generation and allocation}

Assignment lists were generated offline before the physician roster was imported. The generator tracked cumulative assignment counts across 120 case--condition slots formed by 40 cases under U, R and M. For each list, nine cases were sampled without replacement, with higher weight assigned to cases with lower cumulative coverage. Three cases were then assigned to each condition, prioritizing slots with lower coverage, and the nine tasks were independently ordered for presentation.

The initial generation specification used random seed 42. Later list expansion continued from the existing slot counts, but the complete extended random state was unavailable; allocation was therefore not treated as a fully reproducible randomized design. After roster import, participants received pre-numbered lists in sequence. Allocation did not use department, professional title, hospital or subsequent response data.

The main analysis included 250 physicians who completed all nine tasks and 2,250 final strategy responses. Each physician answered nine different cases, three per condition. All 40 cases and all 120 case--condition slots were represented; each slot contained 8--29 responses (mean, 18.75). Unassigned case--condition combinations were structural gaps in the design rather than missing outcomes. Constraints and coverage are reported in Supplementary Table 4.

\subsection{AI judging and panel aggregation}

Gemini 3.1 Pro Preview, GPT-5.5 and Claude Opus 4.7 independently scored each response in three runs. Each run applied the case-level atomic EASS rules and returned 0, 1, 2 or not applicable for each rule, from which APAS, M1--M5 and the nine CSPR gate states were derived.

For the primary APAS panel, rule scores were first averaged across the three runs from each judge and then averaged equally across the three judges. For CSPR, a majority was formed across the three runs within each judge and then across judges; responses without a complete majority were indeterminate. The nine ratings were repeated measurements of the same response and did not increase the number of physicians, cases or responses. CSPR and professional-title-group results are shown in Extended Data Figure 1; full estimates are reported in Supplementary Tables 9a and 9c.

\subsection{Post hoc hospital-background and case-category analysis}

The reliably classified hospital subset contained 237 physicians, 91 registered institutional codes and 2,133 responses. General hospitals contributed 167 physicians and 68 codes; cancer or thoracic/pulmonary-specialty hospitals contributed 70 physicians and 23 codes. These labels described organizational context. Thirteen physicians outside this subset were excluded only from hospital-background analyses, not from the primary task.

Physician-reviewed binary labels described complexity attributable to accompanying disease or constraints and complexity attributable to lung cancer or its treatment. The resulting categories were comorbidity-complex only (9 cases), lung-cancer-complex only (13), and complex on both axes (18). A zero on one axis meant that complexity was not marked on that axis, not that the corresponding disease was absent. There were no neither-complex cases. Labels were post hoc, unordered and separate from the five gray-zone corpus attributes; they did not stratify the original task allocation.

All 18 category-by-hospital-by-condition cells contained responses. One specialty-hospital R cell at the individual-case level was empty. The main analysis retained all 40 cases without imputation; its standardized estimate for that cell relied on the specified interaction structure. A common-support sensitivity excluded that case from both hospital groups and all conditions, giving 39 cases and 2,088 responses. Categories and hospital groups are described in Supplementary Tables 4 and 10.

Three non-exclusive AI-coded markers were assigned to the frozen EASS requirements before examining the new hospital/category effects: explicit lung-cancer decisions, specific accompanying constraints and explicit links from constraints to treatment pathways, sequence or redirection. A marker required explicit textual evidence. Unmarked content did not establish absence of clinical need. General governance lists alone were not coded as case-specific links. The markers were interpretive coding of existing requirements, not a clinical scale or a new evaluation of physician responses.

For each response, original rule ratings were averaged first within judge across repeats, then equally across the three judges. Marker-group scores used original rule weights among scorable marked rules, then averaged equally across observed cases. Marker groups had different eligible rule and case sets. A composite-rule score could not isolate performance on an individual link. Supplementary Table 13 reports complete marker coverage and descriptive scores; the original five-domain panel remained the primary measurement.

\subsection{Post hoc exploratory analysis of subsequent clinical events}

This analysis was conducted after the primary offline evaluation. Follow-up fields were not part of the fixed Case Pack, EASS construction, MCE input, U/R/M task or text-scoring materials. Grouping used de-identified linkage and follow-up text without copying names, medical-record numbers or free-text case summaries into analytic outputs.

Deterministic post hoc text rules separately identified explicit records of death; disease progression or suspected new metastasis; and treatment interruption or non-initiation or deterioration in health status. The categories could overlap. They identified five cases with a death record, two with progression or suspected new metastasis, and 11 with treatment interruption, non-initiation or health-status deterioration, yielding 14 cases with at least one component event record. These rules were not independently clinically adjudicated, and the composite grouping was an exploratory record-based marker rather than overall survival, progression-free survival or a validated clinical endpoint.

Cases were grouped as having at least one component event record, having follow-up without a recorded component event, or having unavailable follow-up; the six cases with unavailable follow-up remained separate. Absence of a recorded event was not treated as absence of progression, survival or favorable prognosis. Date and record-linkage checks identified three prespecified data-integrity concerns: one chronology inconsistency, one unavailable parsable index date and one uncertain record linkage. The main text reports a data-quality sensitivity target excluding these three cases, comprising 12 cases with a component event record and 19 with follow-up but no recorded component event. Analyses including all 34 cases with follow-up, the recorded-death subgroup and content-domain estimates remain supplementary (Supplementary Tables 9f--h).

\section{Supplementary Methods 4. Supportive system-output evaluations}

Three analyses are reported in the order used in the main text: cross-model strategy-output comparison, source-masked content evaluation and the six-configuration functional-module ablation analysis. They used different output sets, judge aggregation and estimands and were not pooled.

\subsection{Cross-model strategy-output comparison}

The comparison used the 40 expert-reference cases. For each case, one complete MCE output and one model-generated strategy from each of GPT-5.4, Claude Opus 4.7 and Gemini 3.1 Pro yielded 160 outputs. The analysis workbook preserves model identities, output texts and scoring records but not the complete system or user instructions used to generate the three model outputs; these outputs are not defined as zero-prompt generations and do not support a claim that prompt conditions were identical. Complete MCE outputs came from the existing study runs and used the same system configuration and fixed knowledge-base version, but they were not generated in the same batch as the three other-model outputs. The four outputs per case were randomly mapped to masked labels A--D. Only numeric citation markers were removed; length, section structure and writing style were not normalized. Judges were instructed not to treat style, structure or length as evidence of quality, although these features could remain source cues.

Gemini 3.1 Pro Preview, GPT-5.5 and Claude Opus 4.7 independently evaluated all four outputs for every case. Each judge--output combination was rated once, yielding 480 judge--output units and 2,880 rule ratings. The scoring records for this batch were generated on 31 August 2026; model access dates were not stored separately in the source records. Each judge--strategy combination used 40 paired cases. APAS was summarized as the arithmetic mean across cases, with a 95\% CI calculated as the mean \ensuremath{\pm} 1.96 \ensuremath{\times} the case-level sample standard deviation / \ensuremath{\sqrt{40}}; CSPR was reported as the number and proportion passing. Judge results were retained separately without voting, averaging or the nine-run U/R/M aggregation.

Domain-level APAS was reconstructed from the same rule ratings for M1--M5. Each case--strategy--judge--domain estimate included only applicable rules scored 0, 1 or 2 and retained the original rule weights. Complete MCE was compared with each other-model output within judge and case. For Gemini 3.1 Pro Preview, the M2 rule was not applicable across all four strategies in two cases, giving 38 paired cases for those M2 estimates; all other domain estimates used 40 cases. Supplementary post hoc two-sided paired t tests assessed zero mean complete-MCE-minus-other-output differences, with separate complete Holm families for nine overall APAS and 45 domain contrasts. Original normal-approximation means and intervals were retained. Domain rankings remained descriptive, with no between-domain test. This analysis compared fixed outputs under the fixed EASS reference; joint differences in model, run, length and organization precluded attribution to a single MCE component (Fig. 4a,b and Supplementary Table 9d).

\subsection{Source-masked content evaluation}

\subsubsection{First-round allocation and presentation}

Each of 60 cases contributed two within-case A/B modules: a primary comparison of MCE with MDT-Debate and a secondary comparison of MCE with one of IR-RAG, Direct or Clinician Plan. Six clinical experts completed 120 first-round modules. The same expert evaluated both modules for a case; each expert assessed ten cases and 20 modules. Each secondary comparator contributed 20 modules. Cases with an empty or explicitly undetermined treating-physician recommendation were ineligible for the Clinician Plan comparison.

Explicit source labels were removed and candidates were labeled A or B. MCE appeared in position A in 58 modules and B in 62. Source guesses were not collected, so source-identification accuracy was not estimable. Each of five dimensions was rated as A better, B better or no difference: D1, key clinical conflict and treatment goals; D2, patient state and treatment fit; D3, safety constraints and unacceptable risks; D4, verification steps capable of changing the decision; and D5, subsequent pathways and alternatives. Experts also made an overall judgment, selected the principal reason and recorded safety-screen flags.

\subsubsection{Independent re-review}

Sixty modules were selected from the initial 120, stratified by case and comparator, with exactly one module per case. The re-review set comprised 30 MDT-Debate comparisons and ten each for IR-RAG, Direct and Clinician Plan. Sampling used only the case and comparator design fields and did not use first-round preferences, D1--D5 ratings, safety flags, response time or text length.

A separate group of six clinicians from the EASS panel who had not participated in the first round completed the re-review, ten modules each. A/B positions were reassigned. Reviewers were masked to source, first-round ratings, safety-screen fields and statistical results. Both rounds used the same case materials, candidate strategies and rating anchors. Allocation and masking are detailed in Supplementary Table 6.

\subsubsection{Analysis}

The unit of preference analysis was the case--comparator A/B module. After decoding with the prespecified source map, preferences were classified as MCE preferred, no difference or comparator preferred. Counts and proportions were reported separately for each round. For the 60 modules rated in both rounds, analyses included the 3 \ensuremath{\times} 3 table, exact agreement, Gwet AC1 and supplementary Cohen \ensuremath{\kappa} for the overall judgment; D1--D5 analyses included exact agreement, Wilson 95\% CIs and unweighted \ensuremath{\kappa}. These dimension-specific estimates describe reproducibility of judgments on the same fixed modules, not the proportion of modules in which MCE was preferred for a given dimension.

Safety-screen flags were described with a 2 \ensuremath{\times} 2 table, positive agreement, \(2a/(2a+b+c)\), and negative agreement, \(2d/(2d+b+c)\). Content flagged by either group entered human review, but a screening flag was not a confirmed safety event. Intervals did not adjust for correlation induced by an expert rating multiple modules, so agreement applies to this fixed re-review set. Preference distributions are shown in Extended Data Figure 4.

\subsection{Functional-module ablation across six configurations}

The six configurations shared the MedGPT base model. The archived n0-n10 workflow was grouped by the dependent clinical and computational objects passed between stages. Multidisciplinary priors entered at n0; n1-n2 represented the case state, goals and constraints; n3-n5 constructed foundational and derived pathways and linked qualifying evidence; n6-n8 performed scenario simulation, critique and candidate repair; n9 performed candidate-strategy reranking and formed candidate-linked verification objects; and n10 controlled stopping assessment and report assembly.

Complete MCE retained the full workflow. Three configurations were targeted grouped functional ablations: removal of the n0 multidisciplinary-prior function; joint removal of n3-n5 pathway and qualifying-evidence construction and n6-n8 simulation, critique and repair; and removal of n6-n8 simulation, critique and repair alone. Both stage-removal configurations retained candidate-strategy reranking at n9, although the joint n3-n8 ablation used configuration-limited retrieval; this does not imply that all candidate-linked verification behavior at n9 was retained. MedGPT base direct generation and MedGPT-LightRAG provided reduced-orchestration reference configurations within the six-configuration design. MedGPT-LightRAG used the same knowledge base as MCE but omitted the remaining MCE orchestration. The functional-block mapping and configuration matrix are provided in Supplementary Table 7.

Each configuration generated one report for each of 40 cases, with one report per case--configuration combination and 240 reports overall. The six reports per case were assigned balanced, rotated labels A--F, and identifying evidence numbers were removed. GPT-5.5, Gemini 3.1 Pro Preview and Claude Opus 4.7 scored the reports against the corresponding EASS rules. This branch used independent judge results rather than the nine-run U/R/M aggregation. GPT-5.5 provided the primary displayed APAS, CSPR and within-case differences. Gemini and Claude Opus 4.7 were not pooled with GPT-5.5 at the case level; their rankings of the six configuration-level mean APAS values were used to describe ranking stability.

Complete MCE was compared with each other configuration using the case as the paired unit, reporting the mean within-case difference and normal-approximation 95\% CI. Supplementary post hoc two-sided paired t tests assessed zero mean differences with Holm adjustment across the five contrasts; the original unadjusted intervals were retained. CSPR for each configuration was described as the proportion passing among 40 cases without a separate confirmatory comparison. Pairwise Spearman's rank correlations between judges used the six configurations as the units. The targeted ablation contrasts estimated differences after removal of defined functional bundles in these fixed cases and realized single runs. The joint n3-n8 ablation did not isolate the contribution of n3-n5, and the design did not estimate individual-node effects, interactions among blocks or run-to-run stability. Output length and organization could also change with the functional intervention. Figure 4c shows GPT-5.5 case-level means, intervals and the paired differences; complete summaries, paired tests and ranking stability are reported in Supplementary Table 7 and Source Data.

\section{Supplementary Methods 5. APAS interpretation and inter-rater reproducibility}

\subsection{Overall clinical anchoring}

Forty case clusters were sampled from existing outputs in the expert-reference set. Each cluster included one physician strategy, one complete MCE output and one comparator output, for 120 outputs. Two attending thoracic surgeons independently rated overall clinical acceptability and the presence of a major clinical defect. Explicit source labels were removed, but original content and structure were retained. Reviewers were not shown source, configuration, U/R/M condition, EASS rules, APAS or CSPR, sampling stratum, or the other reviewer's rating.

Overall clinical acceptability concerned whether a complete strategy could enter clinical discussion and the extent of revision or additional verification required. It was the primary anchor item, while major clinical defect was a supportive diagnostic item; they were not combined into a new score. The original five-category acceptability item comprised four ordered categories and a special not-assessable state and was used for the main distribution and nominal agreement analysis. A prespecified four-category ordinal transformation was used for supplementary weighted agreement and correlation with APAS. Raw independent ratings were used for inter-rater analyses and case-cluster intervals; subsequent targeted reviews, if any, did not overwrite them. The sample was balanced and enriched by source, existing score tier and risk or disagreement, so source-specific proportions were descriptive. Results are shown in Extended Data Figure 2.

\subsection{Re-rating EASS rules for complete MCE}

Two physicians independently re-rated the 240 EASS rules associated with 40 complete MCE reports. Materials concealed configuration, model name, AI ratings and the other physician's judgment. Physicians used the same states of 0, 1, 2, not applicable and not assessable. Analyses used pre-adjudication ratings. Not assessable was not recoded as a numeric score or as not applicable, and paired denominators included rules assessable by both raters. Case-level APAS was calculated only when a physician supplied a 0/1/2 rating for all six rules in the case. Rule-level outcomes were exact agreement and linearly weighted Gwet AC2. Case-level outcomes were ICC(A,1), Lin concordance correlation coefficient, the mean rater difference and Bland--Altman 95\% limits of agreement. Denominators and complete estimates are reported in Supplementary Tables 5, 9a and 9i.

\subsection{Controlled-text perturbation}

The perturbation set comprised 12 cases, 36 original--variant triplets and 108 texts. Each triplet contained the original, a prespecified section-order variant designed to retain content, and a rule-linked target-content-deletion variant. Before AI judging, physicians reviewed 72 candidate variants for attainment of the intended transformation and recorded off-target changes; the texts were then finalized. Targeted domains for content deletion were M2--M5, with nine triplets per domain.

Each text was scored three times by each of Gemini 3.1 Pro Preview, GPT-5.5 and Claude Opus 4.7. A request contained only the anonymized case, strategy text and six selected rules; it did not identify variant type, target domain, source condition or physician. The equal-weight mean of the nine ratings formed the text-level APAS panel value. The triplet was the paired unit and case was the clustering unit. Differences in overall APAS and target-domain APAS between each variant and its original were summarized with 30,000 case-cluster percentile bootstrap samples. Partial gate states and rule transitions derived from the six rule links supplied in each request were supplementary. Not applicable, missing, parse failure and retry states were retained in source denominators; the partial gates were not treated as complete nine-gate CSPR, and repeated ratings were not independent samples. Results are shown in Extended Data Figure 3.

\section{Supplementary Methods 6. Purposive case variants, relationship closure and evaluator dependence}

This analysis describes whether clinical content remained connected to pathway premises, action sequence and fallback in fixed, purposively constructed case variants. It was designed to locate relationships requiring clinical review, not to estimate an overall MCE error rate, a real-world safety-event rate or a causal effect of changing case conditions.

\subsection{Scenario and CCRR construction}

Sixteen source-case templates were purposively selected from the expert-reference set, with four reserves. Each core template produced one baseline scenario and two modified scenarios: one altered a safety constraint or decision-critical information, and the other altered management sequence, branching or reassessment escalation. This yielded 48 scenarios. Six CCRRs covering clinical problems and goals, decision-critical information, candidate pathways, safety constraints and reassessment requirements were specified for each scenario before MCE output was inspected, yielding 288 requirements.

Each member of the fixed three-AI-judge panel independently assigned each requirement one of five states: pass, clear gap, clinical review, not applicable or not assessable. This branch did not use the nine-run U/R/M aggregation. A strict majority determined consensus. A three-way split was assigned clinical review; when one result was missing and the other two agreed, the common state was retained and the missing call recorded separately. AI consensus was used to localize potential relationship gaps structurally and was not treated as a clinical reference standard. States and denominator handling are defined in Supplementary Table 8.

Raw requirement states were reported for all 288 requirements. A conservative definition retained 45 clinical-review and zero not-assessable requirements in the denominator without counting them as passing and excluded one not-applicable requirement, giving 228 of 287. A binary-evaluable definition included only pass and clear gap, giving 228 of 242. These proportions describe denominator handling and are not presented as overall performance measures in the main figure. Scenario status was jointly derived from its six requirements; a scenario was completely closed only when all six passed and none was clinical review, clear gap or not assessable. Requirement-level proportions and scenario-level closure used different analysis units and decision rules. Scenario composition, the template matrix and requirement states are shown in Extended Data Figure 7.

\subsection{Physician calibration and review cohorts}

The prespecified calibration sample selected 24 of 48 scenarios by a fixed procedure: one modified scenario from each source case, four in each of the four modification classes, plus eight baseline scenarios selected across safety strata. Sampling did not use AI judgments or scenario outcomes. Two physicians independently assessed the 144 CCRRs without seeing the existing AI consensus or governance map.

Requirements entered targeted review if they had a clear gap, required clinical review, had judge disagreement or missingness, or were prespecified as high-priority or ungraded safety requirements. Two physicians independently reviewed this cohort. All safety requirements and all requirements on which the physicians disagreed were escalated to a third physician who could see the preceding opinions. The third physician addressed high-risk, discordant or uncertain requirements; this was not blinded independent scoring and did not create a three-rater reference standard. Coverage is summarized in Supplementary Table 5.

Separately from the 24-scenario calibration sample and the 161-requirement targeted review cohort, two physicians independently re-rated all 48 scenarios and 288 CCRRs before adjudication. Main-text requirement-level and scenario-card-level exact agreement and Gwet AC1 came from this complete re-review. Comparisons of each physician with AI consensus on the same objects were evaluator-dependence diagnostics and did not define either as the reference standard. The three sets, 24 scenarios/144 CCRRs, 161 targeted requirements, and 48 scenarios/288 CCRRs, retained separate denominators. Physician--physician and physician--AI agreement and individual-judge states are shown in Extended Data Figure 8; safety-stratified requirement states are reported in Supplementary Table 8.

\section{Supplementary Methods 7. Statistical implementation, diagnostics and sensitivity analyses}

\subsection{U/R/M models and standardization}

Let physician \(i\), case \(j\), task position \(k\), and condition \(a\in\{U,R,M\}\) index a response. The primary APAS model was

\[
E(Y_{ijka}\mid b_i)=\beta_0+\beta_a+\gamma_j+(\beta\gamma)_{aj}+\tau_k+b_i,
\]

where \(b_i\) was a physician random intercept and case and task position were categorical variables. APAS was estimated by restricted maximum likelihood. M--U and M--R were standardized to a reference grid comprising the 40 fixed study cases and nine equally weighted task positions:

\[
\widehat{\mu}^{\mathrm{APAS}}_a=\frac{1}{40\times9}\sum_{j=1}^{40}\sum_{k=1}^{9}\widehat{m}_{ajk}(b=0),
\qquad
\widehat{\Delta}_{M-U}=\widehat{\mu}_M-\widehat{\mu}_U.
\]

M--R was calculated analogously. CSPR used the common-condition-effect model \texttt{CSPR \textasciitilde{} condition + case + task position + (1 | physician)}. With \texttt{expit(x)=1/(1+e\textasciicircum{}\{-x\})}, the standardized probability under condition \(a\) was

\[
\widehat{p}^{\mathrm{CSPR}}_a=\frac{1}{40\times9}\sum_{j=1}^{40}\sum_{k=1}^{9}\mathrm{expit}(\widehat{\eta}_{ajk}(b=0)),
\qquad
\widehat{\Delta}^{\mathrm{CSPR}}_{M-U}=\widehat{p}_M-\widehat{p}_U.
\]

M--R was calculated analogously. Standard errors used the model-parameter covariance matrix and the numerical gradient of the standardization function with the delta method. Predictions set the physician random effect to zero rather than integrating over its distribution; the probability differences are therefore conditional standardized contrasts, not physician-population-average risk differences.

\subsection{Model estimates for subsequent clinical events}

Event-stratified estimates were derived from the primary APAS model fitted to all 40 cases, \texttt{APAS \textasciitilde{} condition \ensuremath{\times} case + task position + (1 | physician)}, rather than from a simplified model refitted within the follow-up subset. Case-specific marginal estimates under U, R and M were obtained and then equally standardized within the specified event target set. The main text reports the 31-case data-quality sensitivity target excluding three records with date or record-linkage concerns; results for all 34 cases with follow-up are supplementary. Within-group U, R and M estimates and M--U and M--R contrasts assessed whether the text differences remained in cases with component event records. Between-group differences in contrasts were secondary exploratory effect-modification estimates. Model covariance was propagated through the linear contrasts to produce asymptotic Wald 95\% CIs.

Event-stratified M1--M5 estimates used the same case-heterogeneous mixed-model structure among responses for which the domain was applicable and entered the denominator, using all 34 cases with follow-up. Recorded-death strata, goal structure, major clinical-decision scenario and the subgroup with non-death records of treatment interruption, non-initiation or deterioration were post hoc exploratory analyses without multiplicity adjustment. All estimates targeted the fixed purposive cases; they did not include uncertainty from case sampling or post hoc target-set selection and did not estimate a causal effect of MCE on subsequent events.

\subsection{Model diagnostics and fitted specifications}

For each mixed model, the formula, number of analytic records, convergence message, fitting warnings and singularity were recorded. Singularity was assessed with \texttt{lme4::isSingular} at tolerance \(10^{-4}\); the optimizer was \texttt{nloptwrap} with a maximum of 100,000 function evaluations. Binary models were also checked for separation or weak-identification signals. APAS judge-run, equal-weight panel, leave-one-judge-out, common-applicability and length-diagnostic models used case-heterogeneous structures. CSPR used the prespecified common-condition-effect model. A case-by-condition binomial model was unstable in the initial GPT-5.5 analysis; the available diagnostics did not identify a more specific separation or weak-identification mechanism. Models, denominators and diagnostics are summarized in Supplementary Table 9.

\subsection{Inter-rater agreement}

For ordinal EASS re-ratings, the paired-complete rules formed the denominator for exact agreement, agreement within one category, linearly weighted Gwet AC2 and linearly weighted Cohen \ensuremath{\kappa}. Case-level APAS for complete MCE outputs was summarized using absolute-agreement, single-rater ICC(A,1), Lin concordance correlation coefficient, the mean difference between raters A and B, and Bland--Altman 95\% limits of agreement, each among complete paired cases.

The case-condition calibration and complete independent re-review remained separate. For the complete re-review, five-state exact agreement, nominal Gwet AC1 and unweighted Cohen \ensuremath{\kappa} were calculated for 288 CCRRs. The six CCRRs in each scenario were combined by the main closure rule into a binary complete versus incomplete outcome, for which exact agreement, Gwet AC1 and category-specific agreement were calculated. The same analyses compared each physician with AI consensus on identical objects as evaluator-dependence diagnostics. Unless otherwise specified, 95\% CIs used 10,000 cluster-percentile bootstrap samples: complete MCE analyses resampled cases, and case-condition analyses resampled the 16 source templates, retaining all within-cluster rules, responses or scenarios. Raw special states from each rater were retained.

Overall clinical anchoring reported exact agreement and nominal Cohen \ensuremath{\kappa} for the original five-category acceptability item, and adjacent agreement and linearly weighted Cohen \ensuremath{\kappa} for the four-category ordinal outcome. Major clinical defects used exact agreement and nominal Cohen \ensuremath{\kappa}. Spearman's rank correlation between APAS and the mean four-category ordinal acceptability rating from the two physicians was calculated across 120 outputs; its 95\% CI used percentile bootstrap resampling of 40 case clusters. Source-level proportions used outputs as denominators; within-case comparisons were descriptive or exploratory.

For controlled-text perturbations, differences between each variant and its original were calculated within triplet and summarized separately for the section-order and target-content-deletion variants. Target-domain differences were stratified by M2--M5, with nine triplets per domain. Overall and domain-specific 95\% CIs used 30,000 case-cluster percentile bootstrap samples. Repeated ratings were first aggregated into text-level panel values and were not independent observations. Partial gate-state transitions used the same triplets and applicability denominator for the six rule links provided in each request and were not combined with APAS differences as complete CSPR.

Format-aligned comparisons also used the case as paired and clustering unit. The nine ratings of each system output were averaged equally; ratings of physician responses were averaged at the response level and then separately within each case and condition. Signed differences in APAS, M1--M5 and non-whitespace character count were equally averaged over 40 cases. CIs used 30,000 case-cluster percentile bootstrap samples with seed 20260907. Supplementary post hoc two-sided paired t tests assessed zero mean differences across the complete 21-contrast family: three overall APAS, 15 domain and three character-count contrasts, with Holm adjustment. The original bootstrap intervals were retained unchanged. This supporting estimand was the within-case system-output-minus-physician-case-mean difference, not a second estimate of the U/R/M condition effect from the physician-task mixed model. Repeated judge runs and multiple physician responses within a case did not increase the case sample size.

Supplementary paired tests used R 4.6.1 \texttt{stats::t.test} and \texttt{stats::p.adjust(method = "holm")}. For each contrast, the statistic was the mean case difference divided by its sample standard deviation over the square root of the number of pairs, with degrees of freedom equal to that number minus one. Gemini M2 had 37 df; the remaining tests had 39 df. Pair keys, applicability, finite values and non-zero variance were checked, and summaries of differences, ties and skewness were retained as diagnostics. These checks did not introduce exclusions or result-dependent test selection. The four new complete Holm families were separate; they did not provide one global error guarantee across all analyses. Existing physician-task and hospital/category model tests and correction families were unchanged. The original bootstrap or normal-approximation CIs were unadjusted and were not inverted to calculate the supplementary t-test P values.

\subsection{Secondary analyses, multiplicity and sensitivity analyses}

M1--M5 used case-heterogeneous linear mixed models, with no imputation of non-applicable records. Holm adjustment was applied to the ten M--U and M--R comparisons across five domains. Figure 3b reports estimates and 95\% CIs; applicable denominators and adjusted P values are in Supplementary Table 9b, and paired domain differences between format-aligned MCE output and M-condition physician responses are in Supplementary Table 9e. Domain-by-professional-title-group interactions were post hoc. Professional titles were grouped as senior (chief physicians), intermediate (associate chief physicians) and junior (attending or resident physicians). The model included condition-by-professional-title-group interactions, assessed with a joint four-degree-of-freedom Wald test. The senior, intermediate and junior groups contained 68, 91 and 91 physicians. Extended Data Figure 1c shows the complete M--U and M--R estimates and 95\% CIs; full values are in Supplementary Table 9c. Professional title was treated only as a coarse indicator of clinical experience.

Overall APAS, M1--M5, recorded death, goal structure and major clinical-decision scenario analyses involving subsequent events were post hoc and outside the Holm family for the primary analysis. The main text reports only the data-quality sensitivity target excluding three records with date or record-linkage concerns; full-record, domain, recorded-death and other case-feature analyses are in Supplementary Tables 9g and 9h. CIs quantify model-estimation uncertainty, and P values, when presented, are unadjusted exploratory statistics. Inclusion or exclusion of zero was not used to declare effect modification.

Sensitivity analyses included crossed physician and case random intercepts; common applicability restricted to rules judged applicable in all nine runs; an input-adaptation analysis excluding 94 responses containing system-specific pathway identifiers; a length-conditioned diagnostic using centered log response length and its interaction with condition; and leave-one-judge-out panels that equally averaged the six ratings from the remaining two judges and refitted the model. Length occurred after condition assignment and was not a baseline covariate in the primary model. Within APAS and CSPR, P values for M--U and M--R were Holm adjusted separately; CIs were unadjusted.

U/R/M mixed models were implemented in R 4.6.1 using \texttt{lme4} 2.0-6 and \texttt{emmeans} 2.0.4. Agreement analyses used Python 3.12.13, pandas 2.2.3 and NumPy 2.3.5. Fixed software versions and random seeds were used for the offline statistical analyses.

\subsection{Models, multiplicity and sensitivity for hospital-by-category analyses}

Five separate REML linear mixed models retained the physician random intercept and condition-by-case, condition-by-hospital-background, condition-by-professional-title-group and task-position fixed effects. Let H indicate specialty-hospital background; L and B indicate lung-cancer-only and both-complex categories, with comorbidity-only as reference; and R and M indicate assistance condition. Additional numeric columns H:L, H:B, H:L:R, H:B:R, H:L:M and H:B:M represented hospital-by-category baseline differences and condition-specific changes. Category main effects and condition-by-category effects were already absorbed by case and condition-by-case fixed effects and were not redundantly entered.

Standardized means weighted cases equally within each category, averaged equally across the nine task positions and used the common 237-physician professional-title distribution: 64 senior, 88 intermediate and 85 junior. Senior denotes chief physicians, intermediate associate chief physicians, and junior attending or resident physicians. Physician random effects were set to zero. These category-specific empirical targets were distinct from prevalence-weighted population estimates.

CR2 covariance estimation clustered on registered institutional code. Single contrasts used Satterthwaite degrees of freedom; two independent constraints used HTZ joint tests. Hospital differences were consistently specialty minus general. The principal heterogeneity family comprised five domains times two support contrasts (M-U and M-R), or 10 tests. Separate complete Holm families covered category-pair differences in hospital increments (30), category-pair differences in U hospital gaps (15), within-category hospital-specific support contrasts (60), direct hospital increment differences within category (30), and six-rule joint heterogeneity tests (12). CIs were unadjusted. These separate families did not provide a global error guarantee across every exploratory analysis.

Prespecified sensitivity analyses used strictly sourced hospital labels, merged same-hospital aliases for clustering, hospital-equal weights with a fixed-effect model, bounded response-level quasi-binomial logit models, the 39-case common-support set, hospital-by-case variable baseline differences, three single-judge panels and three leave-one-judge-out panels. Bounded models averaged pointwise inverse-logit predictions on the original scale and used delta-method contrasts; HTZ inference tested the local nonlinear contrast approximation. Actual applicable denominators were retained for each judge branch. Two hundred domain-by-case leave-one-out fits summarized point-estimate influence without additional significance tests.

All five main models had full fixed-effect rank (143/143) and no singular fits. Linear predictions outside 0-100 occurred for 17, 285, 65, 52 and 13 responses in M1-M5, respectively; bounded sensitivities therefore remained part of the analysis. Estimates with fewer than four effective degrees of freedom were flagged. Complete main comparisons, model diagnostics, sensitivity summaries and machine-readable aggregate estimates are linked through Supplementary Table 9k.

\subsection{Supplementary summaries of relationship closure in case variants}

Requirement states were summarized using two descriptive denominators. The conservative definition counted clinical-review and not-assessable states as not passing and excluded not-applicable requirements. The binary-evaluable definition used pass and clear gap as its denominator and reported the other states separately. Proportions were first calculated within each source template and then averaged equally across 16 templates. They were not interpreted as overall performance, error or clinical-safety rates.

A scenario was completely closed only if all six requirements returned a state, at least one entered the pass/clear-gap denominator, every requirement entering that denominator passed, and none was a clear gap, clinical review or not assessable. Each modified scenario was paired with the baseline scenario from its template. The overall paired difference first averaged the two modified scenarios within template and then averaged equally across templates.

Rule-level proportions used Wilson 95\% CIs. Template-weighted estimates, scenario-closure proportions and paired differences used 10,000 nonparametric source-template cluster bootstrap samples with seed 20260812. Safety requirements were stratified by prespecified priority. When no clear gap was observed, a one-sided Clopper--Pearson 95\% upper confidence bound was calculated. Judge quality was described by return completeness, vote composition, any-disagreement proportion, Fleiss \ensuremath{\kappa} and scenario closure by individual judge. No P values were calculated, and the 288 requirements were not treated as independent samples.

Complete independent physician re-review retained the original five-state requirement ratings and derived binary scenario closure using the same joint rule. Exact agreement and Gwet AC1 were calculated between physicians and between each physician and AI consensus at both levels. CIs used 10,000 bootstrap samples clustered by the 16 source templates. Rule-level and scenario-card-level results were reported separately; agreement at the card level was not used to imply interchangeability of specific rule judgments.

\section{Supplementary-material correspondence}

Supplementary Methods 1--7 cover EASS and APAS/CSPR, MCE and comparison objects, physician tasks and post hoc case-context analyses, supporting system-output evaluations, APAS interpretation, case-condition changes, and statistical implementation. Supplementary Tables 1--13 and Extended Data Figures 1--9 retain their existing identifiers. Supplementary Table 9k and Table 13 report the hospital/category analysis and rule-content interpretation; Extended Data Figure 9 presents the complete five-domain patterns. Source Data provide aggregate machine-readable interfaces for displayed and sensitivity results. A separate case-level reproducibility layer will provide de-identified and disclosure-reviewed case-level rules, physician responses, model outputs and retained prompts. Original unpublished clinical text and free-text follow-up remain governed by the applicable privacy, ethics and data-governance requirements.

\clearpage
These tables report the case-specific clinical reference and scoring rules, study design, analytic coverage, denominators, statistical models and object-level summaries. Abbreviations: Expert-Admissible Strategy Space (EASS), Admissible Pathway Attainment Score (APAS), Complete-and-Safe Pathway Rate (CSPR), and Condition-Change Response Rule (CCRR).

\section{Supplementary Table 1. Expert process and rule data structure for the EASS case-specific clinical reference}

\subsection{a. Expert process}

\begingroup
\scriptsize
\setlength{\tabcolsep}{3pt}
\renewcommand{\arraystretch}{1.12}
\begin{longtable}{@{}>{\raggedright\arraybackslash}p{0.087\linewidth}>{\raggedright\arraybackslash}p{0.187\linewidth}>{\raggedright\arraybackslash}p{0.252\linewidth}>{\raggedright\arraybackslash}p{0.254\linewidth}>{\raggedright\arraybackslash}p{0.160\linewidth}@{}}
\toprule
\textbf{Stage} & \textbf{Expert or review scale} & \textbf{Unit} & \textbf{Prespecified handling rule} & \textbf{Resulting study object} \\
\midrule
\endfirsthead
\toprule
\textbf{Stage} & \textbf{Expert or review scale} & \textbf{Unit} & \textbf{Prespecified handling rule} & \textbf{Resulting study object} \\
\midrule
\endhead
\midrule
\multicolumn{5}{r}{\footnotesize Continued on next page} \\
\endfoot
\bottomrule
\endlastfoot
Panel composition & 20 chief physicians from 15 institutions and six disciplines & 19 attended the first in-person round; one joined later online rounds & Subsequent case review was undertaken by 14 treatment-focused experts & EASS expert panel \\
Round 1 & 19 experts & Clinical problems, goals, information gaps, pathways, safety and governance in 40 cases & Independent review organized into M1--M5 & Structured case-review material \\
Round 2 & 14 treatment-focused experts & 22 cases reviewed by 4 experts and 18 by 3; 142 expert--case units & Strict majority: at least 2 votes in 3-person review and 3 in 4-person review & Baseline item judgments and items requiring clarification \\
Round 3 & 3 experts per item & 116 independent items; 348 valid responses & Strict majority of valid responses; conservative handling of 1:1:1 splits in information or governance items & Conditions, disagreement and fallback rules \\
M4 safety-rule formation & 40 final rules; 30 safety items carried forward round-2 expert opinions & Round-2 case cards, safety-boundary table and safety review; round-3 pathway-applicability conditions added & Incorporated into APAS rules and linked CSPR gates & Final M4 safety rules \\
Final reference & 240 case-level rules & 40 cases \ensuremath{\times} 6 rules & 139 confirmed-acceptable and 101 conditionally acceptable rules & EASS case-specific clinical reference \\
\end{longtable}
\endgroup

M1--M5 are content domains describing what a case-specific strategy should address, including reasonable pathways, treatment prerequisites, patient constraints and subsequent arrangements. The five structural relations in the main text assess candidate-pathway differentiation, action-premise relations, decision-changing unknowns, verification-candidate relations and reassessment/fallback relations. The content reference measures attainment of applicable requirements; the structural audit checks whether pathways, prerequisites and subsequent actions are coherently connected.

EASS was independently constructed from fixed case materials and expert review. MCE, comparator and physician outputs were not shown to the EASS experts, and EASS rules did not enter generation of those outputs.

\subsection{b. Rule fields}

\begingroup
\small
\setlength{\tabcolsep}{3pt}
\renewcommand{\arraystretch}{1.12}
\begin{longtable}{@{}>{\raggedright\arraybackslash}p{0.224\linewidth}>{\raggedright\arraybackslash}p{0.391\linewidth}>{\raggedright\arraybackslash}p{0.325\linewidth}@{}}
\toprule
\textbf{Field} & \textbf{Meaning} & \textbf{Analytic use} \\
\midrule
\endfirsthead
\toprule
\textbf{Field} & \textbf{Meaning} & \textbf{Analytic use} \\
\midrule
\endhead
\midrule
\multicolumn{3}{r}{\footnotesize Continued on next page} \\
\endfoot
\bottomrule
\endlastfoot
\texttt{rule\_id} & Unique within-case rule identifier & Scoring, audit and result linkage \\
\texttt{case\_id} & Case identifier & Case-level clustering and mapping \\
\texttt{module} & M1--M5 clinical content domain & Domain summaries \\
\texttt{final\_state} & Confirmed- or conditionally acceptable classification & Rule interpretation boundary \\
\texttt{final\_requirement} & Clinical requirement to be met & Atomic rule scoring \\
\texttt{applicability} & Conditions under which the rule applies & Denominator determination \\
\texttt{accepted\_actions} & Acceptable management actions & Score 2 or conditional compliance \\
\texttt{prohibited\_actions} & Prohibited management actions & Score 0 and failure coding \\
\texttt{fallback\_actions} & Prespecified fallback actions & Correct fallback and denominator handling \\
\texttt{weight} & 3 for M1--M4; 2 for M5 & APAS weighting \\
\texttt{score\_rule} & Anchors for scores 0, 1 and 2 & Judge scoring \\
\texttt{cspr\_links} & Linked CSPR gates & Connection between APAS and CSPR rules \\
\texttt{failure\_codes} & Failure-type codes & Failure-mode summaries \\
\texttt{version} & Rule-version key & Version linkage \\
\end{longtable}
\endgroup

\section{Supplementary Table 2. APAS rules and the nine CSPR gates}

\subsection{a. APAS operational rules}

\begingroup
\small
\setlength{\tabcolsep}{3pt}
\renewcommand{\arraystretch}{1.12}
\begin{longtable}{@{}>{\raggedright\arraybackslash}p{0.254\linewidth}>{\raggedright\arraybackslash}p{0.686\linewidth}@{}}
\toprule
\textbf{Item} & \textbf{Definition} \\
\midrule
\endfirsthead
\toprule
\textbf{Item} & \textbf{Definition} \\
\midrule
\endhead
\midrule
\multicolumn{2}{r}{\footnotesize Continued on next page} \\
\endfoot
\bottomrule
\endlastfoot
Score 2 & Fully satisfies the case-level clinical requirement \\
Score 1 & Partly satisfies the requirement or follows a compliant action under its specified condition \\
Score 0 & Omits or conflicts with the requirement, or uses a restricted pathway when its condition is unmet \\
Not applicable & Excluded from APAS numerator and denominator \\
Correct fallback & Excluded from APAS numerator and denominator \\
Reasonable disagreement & Excluded from numerator and denominator when within the prespecified admissible strategy space \\
Weight & 3 for M1--M4 and 2 for M5 \\
Formula & \texttt{100 \ensuremath{\times} \ensuremath{\Sigma}(w\_i r\_i) /\allowbreak{} (2 \ensuremath{\times} \ensuremath{\Sigma}w\_i)}, summing only applicable rules entering the denominator \\
\end{longtable}
\endgroup

APAS measures how fully an evaluated strategy expresses applicable case-level clinical requirements. APAS and CSPR are derived from the same EASS reference but calculated separately; a higher APAS does not imply CSPR passage. The word \emph{safe} in CSPR refers to satisfaction of prespecified textual scoring constraints, not measured clinical safety.

\subsection{b. CSPR gates}

\begingroup
\small
\setlength{\tabcolsep}{3pt}
\renewcommand{\arraystretch}{1.12}
\begin{longtable}{@{}>{\raggedright\arraybackslash}p{0.228\linewidth}>{\raggedright\arraybackslash}p{0.712\linewidth}@{}}
\toprule
\textbf{Gate} & \textbf{Passing requirement} \\
\midrule
\endfirsthead
\toprule
\textbf{Gate} & \textbf{Passing requirement} \\
\midrule
\endhead
\midrule
\multicolumn{2}{r}{\footnotesize Continued on next page} \\
\endfoot
\bottomrule
\endlastfoot
CSPR-01 & Strategy addresses the principal clinical problem \\
CSPR-02 & Includes an acceptable or conditionally acceptable main pathway, or correctly chooses verification before commitment \\
CSPR-03 & Does not violate a prespecified high-priority safety constraint in EASS \\
CSPR-04 & Addresses all critical judgment items \\
CSPR-05 & Avoids an unjustified definitive commitment when blocking information is missing \\
CSPR-06 & Respects applicable constraints from comorbidity, organ function, performance status, resources and patient preferences \\
CSPR-07 & Includes monitoring, reassessment or fallback for high-risk pathways \\
CSPR-08 & Includes appropriate governance arrangements \\
CSPR-09 & Contains no sequence, mutual-exclusivity or premise conflict in the main pathway \\
\end{longtable}
\endgroup

Failure of any applicable gate yielded CSPR failure; a non-applicable gate was not a failure. CSPR was a judge-defined secondary or exploratory joint gate and was not interpreted as a clinical safety rate or patient-risk estimate.

\begin{landscape}
\section{Supplementary Table 3. Generation-transparency matrix for systems, assistance conditions and comparators}

\begingroup
\scriptsize
\setlength{\tabcolsep}{3pt}
\renewcommand{\arraystretch}{1.12}
\begin{longtable}{@{}>{\raggedright\arraybackslash}p{0.103\linewidth}>{\raggedright\arraybackslash}p{0.134\linewidth}>{\raggedright\arraybackslash}p{0.236\linewidth}>{\raggedright\arraybackslash}p{0.169\linewidth}>{\raggedright\arraybackslash}p{0.092\linewidth}>{\raggedright\arraybackslash}p{0.206\linewidth}@{}}
\toprule
\textbf{Object} & \textbf{Case input} & \textbf{Verified model or method} & \textbf{Knowledge or retrieval relation} & \textbf{Output} & \textbf{Evaluation use} \\
\midrule
\endfirsthead
\toprule
\textbf{Object} & \textbf{Case input} & \textbf{Verified model or method} & \textbf{Knowledge or retrieval relation} & \textbf{Output} & \textbf{Evaluation use} \\
\midrule
\endhead
\midrule
\multicolumn{6}{r}{\footnotesize Continued on next page} \\
\endfoot
\bottomrule
\endlastfoot
MCE & Fixed case input & MedGPT base; state/\allowbreak{}goal representation, pathway and evidence organization, critique and repair, uncertainty handling and report assembly & Prespecified evidence scope; EASS and CCRR excluded from generation & Strategy Review Pack & M condition, complete MCE in six-configuration comparison, source-masked candidate, and case-condition evaluation \\
U condition & Fixed case input & No additional generation & No additional knowledge material & Physician-submitted strategy & Primary U/\allowbreak{}R/\allowbreak{}M evaluation \\
R condition & Fixed case input & GPT-5.4 with LightRAG & Same knowledge base as MCE; does not imply the same retriever, prompt or orchestration & Case-specific reference material & Primary U/\allowbreak{}R/\allowbreak{}M evaluation \\
M condition & Fixed case input plus Strategy Review Pack & MCE output available for physician review & Physician could question, revise and supplement the support content & Physician-submitted final strategy & Primary U/\allowbreak{}R/\allowbreak{}M evaluation \\
Direct & Fixed case material & Direct generation comparator & No MCE orchestration & Candidate strategy & Secondary source-masked comparison \\
Cross-model other-model strategy output set & 40 expert-reference cases & One strategy from each of GPT-5.4, Claude Opus 4.7 and Gemini 3.1 Pro & No MCE orchestration; not generated in the same batch as complete MCE & Three other-model strategies per case & Supportive system-output comparison in Figure 4a,b \\
IR-RAG & Fixed case material & Independent retrieval-augmented comparator & Retrieval process independent of MCE & Candidate strategy & Secondary source-masked comparison \\
MDT-Debate & Fixed case material & Multi-role discussion comparator & Not a formal human multidisciplinary team & Candidate strategy & Primary source-masked comparison \\
Clinician Plan & Treating physician's original recommendation in the fixed case & Original human recommendation & No additional generation & Original clinical recommendation & Secondary source-masked comparison \\
AI judge & Case material, evaluated strategy and corresponding rules & GPT-5.5, Gemini 3.1 Pro Preview and Claude Opus 4.7 & Did not generate the evaluated physician response & Atomic rule and gate judgments & APAS, CSPR and CCRR evaluation \\
\end{longtable}
\endgroup

The term \emph{six-configuration functional-module ablation analysis} refers to the six systems sharing the MedGPT base model. The matrix separates model or functional role, case input, knowledge relation, output object and evaluation purpose.

\end{landscape}
\section{Supplementary Table 4. U/R/M physician-task design and analytic-set coverage}

\begingroup
\small
\setlength{\tabcolsep}{3pt}
\renewcommand{\arraystretch}{1.12}
\begin{longtable}{@{}>{\raggedright\arraybackslash}p{0.200\linewidth}>{\raggedright\arraybackslash}p{0.358\linewidth}>{\raggedright\arraybackslash}p{0.382\linewidth}@{}}
\toprule
\textbf{Item} & \textbf{Design or observation} & \textbf{Interpretation} \\
\midrule
\endfirsthead
\toprule
\textbf{Item} & \textbf{Design or observation} & \textbf{Interpretation} \\
\midrule
\endhead
\midrule
\multicolumn{3}{r}{\footnotesize Continued on next page} \\
\endfoot
\bottomrule
\endlastfoot
Physicians analyzed & 250 & All completed 9 tasks \\
Complete responses & 2,250 & 250 \ensuremath{\times} 9 \\
Cases per physician & 9 & No repeated case within an assignment list \\
Conditions per physician & 3 each under U, R and M & Balanced number of conditions within physician \\
Cases & 40/\allowbreak{}40 covered & Fixed empirical target set \\
Case--condition slots & 120/\allowbreak{}120 covered & 40 \ensuremath{\times} 3 \\
Responses per slot & 8--29 & Incompletely balanced \\
Mean responses per slot & 18.75 & 2,250 /\allowbreak{} 120 \\
Initial generation seed & 42 & Initial assignment-list specification \\
Allocation basis & Sequence of pregenerated assignment lists & Department, title, hospital and outcome fields were not used \\
Structural gaps & Case--condition combinations not assigned to a physician & Not missing outcomes and not imputed \\
\end{longtable}
\endgroup

\subsection{Hospital-background analytic subset}

\begingroup
\footnotesize
\setlength{\tabcolsep}{3pt}
\renewcommand{\arraystretch}{1.12}
\begin{longtable}{@{}>{\raggedright\arraybackslash}p{0.228\linewidth}>{\raggedright\arraybackslash}p{0.197\linewidth}>{\raggedright\arraybackslash}p{0.319\linewidth}>{\raggedright\arraybackslash}p{0.197\linewidth}@{}}
\toprule
\textbf{Item} & \textbf{General hospitals} & \textbf{Cancer or thoracic/\allowbreak{}pulmonary-specialty hospitals} & \textbf{Total} \\
\midrule
\endfirsthead
\toprule
\textbf{Item} & \textbf{General hospitals} & \textbf{Cancer or thoracic/\allowbreak{}pulmonary-specialty hospitals} & \textbf{Total} \\
\midrule
\endhead
\midrule
\multicolumn{4}{r}{\footnotesize Continued on next page} \\
\endfoot
\bottomrule
\endlastfoot
Physicians & 167 & 70 & 237 \\
Registered institutional codes & 68 & 23 & 91 \\
Responses & 1,503 & 630 & 2,133 \\
\end{longtable}
\endgroup

Thirteen physicians and 117 responses from the full task were outside the reliably classifiable hospital subset. Institutional codes are registration units, not verified independent hospital campuses. All 40 cases and all 18 category-by-hospital-by-condition cells were represented. One case-level specialty-hospital R cell had no observed response; the 40-case standardized estimate therefore retains model dependence for that cell. The prespecified common-support sensitivity used 39 cases and 2,088 responses. Hospital categories do not measure clinical resources, MDT access or treatment capability.

\section{Supplementary Table 5. Coverage matrix for AI judging and physician review}

\begingroup
\scriptsize
\setlength{\tabcolsep}{3pt}
\renewcommand{\arraystretch}{1.12}
\begin{longtable}{@{}>{\raggedright\arraybackslash}p{0.149\linewidth}>{\raggedright\arraybackslash}p{0.096\linewidth}>{\raggedright\arraybackslash}p{0.253\linewidth}>{\raggedright\arraybackslash}p{0.178\linewidth}>{\raggedright\arraybackslash}p{0.263\linewidth}@{}}
\toprule
\textbf{Evaluation object} & \textbf{Rater} & \textbf{Repeats or sample} & \textbf{Main output} & \textbf{Inferential role} \\
\midrule
\endfirsthead
\toprule
\textbf{Evaluation object} & \textbf{Rater} & \textbf{Repeats or sample} & \textbf{Main output} & \textbf{Inferential role} \\
\midrule
\endhead
\midrule
\multicolumn{5}{r}{\footnotesize Continued on next page} \\
\endfoot
\bottomrule
\endlastfoot
U/\allowbreak{}R/\allowbreak{}M physician responses & 3 named AI judges & 3 runs per judge; 9 ratings per response & APAS, M1--M5 and nine CSPR gates & Fixed-panel primary scoring; repeats did not increase the clinical sample \\
Format-aligned MCE output & 3 named AI judges & 3 runs per judge; 9 ratings per output & APAS and M1--M5 & Same repeat and aggregation scheme as physician responses; case-level paired comparison \\
Cross-model other-model strategy outputs & 3 named AI judges & One rating per judge--output; 480 judge--output units & Judge-specific APAS and CSPR & Judges reported separately; no voting, averaging or nine-run aggregation \\
Structural-relation audit of complete MCE outputs & 3 named AI judges & One assessment per judge--output; 120 judge--output units & Five relations, overall relationship closure and defect labels & Judges reported separately; verification tasks nested within outputs \\
Six-configuration functional-module ablation analysis & 3 named AI judges & 40 cases \ensuremath{\times} 6 configurations; one report per case--configuration & GPT-5.5 APAS, CSPR and within-case contrasts; rankings from 3 judges & No nine-run aggregation; other judges not pooled with GPT-5.5 at case level \\
AI evaluation of case-condition changes & 3 named AI judges & 48 scenarios and 288 CCRRs; one judgment per judge--requirement & Five-state consensus; requirement and scenario outcomes & Strict-majority consensus; no nine-run aggregation \\
Overall clinical anchoring & 2 attending thoracic surgeons & 40 case clusters, 120 outputs and 240 ratings & Overall acceptability and major clinical defect & Independent physician ratings, separate from rule-level re-rating; case-cluster analysis; no human reference standard \\
Controlled-text perturbation & 3 named AI judges & 12 cases, 36 triplets and 108 texts; 9 ratings per text & Text-level APAS, target-domain APAS and supplementary rule states & Paired triplets; case-cluster bootstrap; repeats did not increase sample size \\
EASS rules for complete MCE & 2 physicians & 40 cases and 240 rules; 240 paired rules for physician--physician, physician 1--AI and physician 2--AI comparisons & 0/\allowbreak{}1/\allowbreak{}2, not applicable or not assessable & Rule-level agreement; complete cases used for case-level APAS \\
Case-condition calibration & 2 physicians & 24 scenarios and 144 CCRRs & Five-state judgments & CCRR-state agreement \\
Targeted review cohort & 2 physicians & 161 requirements & Independent review state and rationale & Review of high-risk, discordant and uncertain requirements \\
Escalation cohort & Third physician & 61 requirements & Targeted review with earlier opinions visible & Escalation review, not three-rater blinded agreement \\
Complete independent case-condition re-review & 2 physicians & 48 scenarios and 288 CCRRs & Five-state ratings and scenario closure & Main rule-level and scenario-card agreement; separate from 24/\allowbreak{}144 calibration and 161-requirement targeted review \\
\end{longtable}
\endgroup

\section{Supplementary Table 6. Allocation, masking and re-review in the source-masked content evaluation}

\subsection{a. Allocation and masking}

\begingroup
\small
\setlength{\tabcolsep}{3pt}
\renewcommand{\arraystretch}{1.12}
\begin{longtable}{@{}>{\raggedright\arraybackslash}p{0.230\linewidth}>{\raggedright\arraybackslash}p{0.236\linewidth}>{\raggedright\arraybackslash}p{0.474\linewidth}@{}}
\toprule
\textbf{Item} & \textbf{First round} & \textbf{Independent re-review} \\
\midrule
\endfirsthead
\toprule
\textbf{Item} & \textbf{First round} & \textbf{Independent re-review} \\
\midrule
\endhead
\midrule
\multicolumn{3}{r}{\footnotesize Continued on next page} \\
\endfoot
\bottomrule
\endlastfoot
Cases & 60 & 60 \\
Modules & 120 & 60 \\
Experts & 6 & A separate group of 6 EASS panelists who did not participate in the first round \\
Per expert & 10 cases, 20 modules & 10 modules \\
MCE versus MDT-Debate & 60 & 30 \\
MCE versus IR-RAG & 20 & 10 \\
MCE versus Direct & 20 & 10 \\
MCE versus Clinician Plan & 20 & 10 \\
Modules per case & Two & Exactly one \\
MCE A/\allowbreak{}B position & A in 58; B in 62 & Reassigned \\
Fields available for sampling & Not applicable & Case and comparator design fields \\
Fields excluded from sampling & Not applicable & First-round preference, D1--D5, safety fields, response time and text length \\
Re-review masking & Explicit source labels removed & First-round ratings, safety screen and statistical results also concealed \\
\end{longtable}
\endgroup

\subsection{b. Overall preference and agreement at independent re-review}

\begingroup
\scriptsize
\setlength{\tabcolsep}{3pt}
\renewcommand{\arraystretch}{1.12}
\begin{longtable}{@{}>{\raggedright\arraybackslash}p{0.253\linewidth}>{\raggedright\arraybackslash}p{0.172\linewidth}>{\raggedright\arraybackslash}p{0.172\linewidth}>{\raggedright\arraybackslash}p{0.172\linewidth}>{\raggedright\arraybackslash}p{0.172\linewidth}@{}}
\toprule
\textbf{Evaluation set} & \textbf{Modules} & \textbf{MCE preferred} & \textbf{No difference} & \textbf{Comparator preferred} \\
\midrule
\endfirsthead
\toprule
\textbf{Evaluation set} & \textbf{Modules} & \textbf{MCE preferred} & \textbf{No difference} & \textbf{Comparator preferred} \\
\midrule
\endhead
\midrule
\multicolumn{5}{r}{\footnotesize Continued on next page} \\
\endfoot
\bottomrule
\endlastfoot
All first-round modules & 120 & 100 (83.3\%) & 6 (5.0\%) & 14 (11.7\%) \\
First-round modules selected for re-review & 60 & 54 (90.0\%) & 2 (3.3\%) & 4 (6.7\%) \\
Independent re-review & 60 & 57 (95.0\%) & 1 (1.7\%) & 2 (3.3\%) \\
\end{longtable}
\endgroup

Among the 60 fixed modules rated in both rounds, three-category judgments agreed exactly in 55/60 (91.7\%; 95\% CI, 81.9\%--96.4\%), with Gwet AC1 = 0.910 (0.826--0.982). Supplementary unweighted Cohen \ensuremath{\kappa} was 0.414 (\ensuremath{-}0.034 to 0.793). Sparse no-difference and comparator-preferred categories affected \ensuremath{\kappa} through the marginal distribution, so exact agreement and Gwet AC1 were emphasized. Machine-readable preference distributions, comparator strata, the cross-tabulation and agreement estimates are provided in Source Data.

\subsection{c. Five clinical-content dimensions in the independent re-review}

\begingroup
\footnotesize
\setlength{\tabcolsep}{3pt}
\renewcommand{\arraystretch}{1.12}
\begin{longtable}{@{}>{\raggedright\arraybackslash}p{0.311\linewidth}>{\raggedright\arraybackslash}p{0.227\linewidth}>{\raggedright\arraybackslash}p{0.201\linewidth}>{\raggedright\arraybackslash}p{0.201\linewidth}@{}}
\toprule
\textbf{Dimension} & \textbf{Exact agreement across rounds} & \textbf{Exact agreement rate} & \textbf{Wilson 95\% CI} \\
\midrule
\endfirsthead
\toprule
\textbf{Dimension} & \textbf{Exact agreement across rounds} & \textbf{Exact agreement rate} & \textbf{Wilson 95\% CI} \\
\midrule
\endhead
\midrule
\multicolumn{4}{r}{\footnotesize Continued on next page} \\
\endfoot
\bottomrule
\endlastfoot
D1 key clinical conflicts and treatment goals & 40/\allowbreak{}60 & 66.7\% & 54.1\%--77.3\% \\
D2 patient state and treatment fit & 44/\allowbreak{}60 & 73.3\% & 61.0\%--82.9\% \\
D3 safety constraints and unacceptable risk & 48/\allowbreak{}60 & 80.0\% & 68.2\%--88.2\% \\
D4 decision-changing verification steps & 56/\allowbreak{}60 & 93.3\% & 84.1\%--97.4\% \\
D5 subsequent pathways and alternatives & 48/\allowbreak{}60 & 80.0\% & 68.2\%--88.2\% \\
\end{longtable}
\endgroup

Dimension-specific results describe reproducibility of judgments across the two rounds for the same 60 fixed text modules; they do not represent the proportion judged to favor MCE in the corresponding dimension. Safety-screen and other diagnostic results remain in governed materials.

\section{Supplementary Table 7. Functional-module ablation across six configurations sharing the MedGPT base model}

\subsection{a. Functional-block and Figure 2 node mapping}

\begingroup
\footnotesize
\setlength{\tabcolsep}{3pt}
\renewcommand{\arraystretch}{1.12}
\begin{longtable}{@{}>{\raggedright\arraybackslash}p{0.173\linewidth}>{\raggedright\arraybackslash}p{0.099\linewidth}>{\raggedright\arraybackslash}p{0.311\linewidth}>{\raggedright\arraybackslash}p{0.357\linewidth}@{}}
\toprule
\textbf{Functional block} & \textbf{Figure 2 nodes} & \textbf{Scientific role} & \textbf{Treatment in the six configurations} \\
\midrule
\endfirsthead
\toprule
\textbf{Functional block} & \textbf{Figure 2 nodes} & \textbf{Scientific role} & \textbf{Treatment in the six configurations} \\
\midrule
\endhead
\midrule
\multicolumn{4}{r}{\footnotesize Continued on next page} \\
\endfoot
\bottomrule
\endlastfoot
Multidisciplinary priors & n0 & Supplies multidisciplinary priors before case-specific construction & Removed only in the corresponding grouped ablation \\
Case-state, goal and constraint representation & n1-n2 & Compiles the fixed case and represents goals, constraints and decision-changing unknowns & Retained in the MCE-derived ablation configurations \\
Pathway and qualifying-evidence construction & n3-n5 & Constructs foundational and derived pathways and links evidence that supports, limits or leaves applicability unresolved & Removed jointly with n6-n8 in the n3-n8 ablation \\
Simulation, critique and repair & n6-n8 & Tests candidate benefits, risks, feasibility and dependencies and repairs or removes candidates & Removed alone in one grouped ablation and jointly with n3-n5 in another \\
Candidate reranking and candidate-linked verification & n9 & Reranks candidate strategies and forms verification objects linked to subsequent actions & Candidate-strategy reranking was retained in both stage-removal configurations; complete retention of every verification behavior was not assumed \\
Stopping assessment and report assembly & n10 & Determines whether the object returns to an earlier stage or is assembled as a Strategy Review Pack & Retained in the MCE-derived ablation configurations \\
\end{longtable}
\endgroup

Nodes were grouped because stages within a block operated on dependent intermediate objects. The experimental interventions therefore concerned functional bundles rather than independently acting nodes.

\begin{landscape}
\subsection{b. Configuration matrix}

\begingroup
\scriptsize
\setlength{\tabcolsep}{3pt}
\renewcommand{\arraystretch}{1.12}
\begin{longtable}{@{}>{\raggedright\arraybackslash}p{0.166\linewidth}>{\raggedright\arraybackslash}p{0.112\linewidth}>{\raggedright\arraybackslash}p{0.136\linewidth}>{\raggedright\arraybackslash}p{0.133\linewidth}>{\raggedright\arraybackslash}p{0.124\linewidth}>{\raggedright\arraybackslash}p{0.148\linewidth}>{\raggedright\arraybackslash}p{0.121\linewidth}@{}}
\toprule
\textbf{Configuration} & \textbf{Multidisciplinary priors} & \textbf{Pathways and qualifying evidence} & \textbf{Simulation, critique and repair} & \textbf{Candidate-strategy reranking} & \textbf{Retrieval augmentation} & \textbf{Remaining MCE orchestration} \\
\midrule
\endfirsthead
\toprule
\textbf{Configuration} & \textbf{Multidisciplinary priors} & \textbf{Pathways and qualifying evidence} & \textbf{Simulation, critique and repair} & \textbf{Candidate-strategy reranking} & \textbf{Retrieval augmentation} & \textbf{Remaining MCE orchestration} \\
\midrule
\endhead
\midrule
\multicolumn{7}{r}{\footnotesize Continued on next page} \\
\endfoot
\bottomrule
\endlastfoot
Complete MCE & Retained & Retained & Retained & Retained & Retained & Retained \\
Without multidisciplinary priors & Removed & Retained & Retained & Retained & Retained & Retained \\
Without pathway/\allowbreak{}evidence and review stages & Retained & Removed & Removed & Retained & Configuration-limited & Other stages retained \\
Without simulation, critique and repair & Retained & Retained & Removed & Retained & Retained & Other stages retained \\
MedGPT base direct generation & Not applicable & Not applicable & Not applicable & Not applicable & Removed & Removed \\
MedGPT-LightRAG & Not applicable & Not applicable & Not applicable & Not applicable & Retained; same knowledge base as MCE & Removed \\
\end{longtable}
\endgroup

All six configurations shared the MedGPT base model. Complete MCE was the full configuration; the next three rows were targeted grouped functional ablations; and MedGPT base direct generation and MedGPT-LightRAG were reduced-orchestration reference configurations. The n3-n8 configuration jointly removed two adjacent functional blocks and did not isolate the n3-n5 pathway and evidence block. Each case--configuration report was generated once; the comparisons do not identify individual-node effects, interactions among functional blocks or run-to-run stability.

\end{landscape}
\subsection{c. Case-level summaries scored by GPT-5.5}

\begingroup
\footnotesize
\setlength{\tabcolsep}{3pt}
\renewcommand{\arraystretch}{1.12}
\begin{longtable}{@{}>{\raggedright\arraybackslash}p{0.310\linewidth}>{\raggedright\arraybackslash}p{0.210\linewidth}>{\raggedright\arraybackslash}p{0.210\linewidth}>{\raggedright\arraybackslash}p{0.210\linewidth}@{}}
\toprule
\textbf{Configuration} & \textbf{Cases} & \textbf{Mean APAS (95\% CI)} & \textbf{CSPR passing/\allowbreak{}40 (\%)} \\
\midrule
\endfirsthead
\toprule
\textbf{Configuration} & \textbf{Cases} & \textbf{Mean APAS (95\% CI)} & \textbf{CSPR passing/\allowbreak{}40 (\%)} \\
\midrule
\endhead
\midrule
\multicolumn{4}{r}{\footnotesize Continued on next page} \\
\endfoot
\bottomrule
\endlastfoot
Complete MCE & 40 & 91.91 (89.12--94.71) & 37/\allowbreak{}40 (92.5\%) \\
Without multidisciplinary priors & 40 & 75.29 (70.69--79.89) & 32/\allowbreak{}40 (80.0\%) \\
Without pathway/\allowbreak{}evidence and review stages & 40 & 84.19 (80.94--87.44) & 37/\allowbreak{}40 (92.5\%) \\
Without simulation, critique and repair & 40 & 88.31 (85.42--91.20) & 37/\allowbreak{}40 (92.5\%) \\
MedGPT base direct generation & 40 & 68.09 (60.57--75.61) & 25/\allowbreak{}40 (62.5\%) \\
MedGPT-LightRAG & 40 & 77.57 (72.13--83.01) & 33/\allowbreak{}40 (82.5\%) \\
\end{longtable}
\endgroup

\subsection{d. Within-case paired comparisons of complete MCE with other configurations}

\begingroup
\scriptsize
\setlength{\tabcolsep}{3pt}
\renewcommand{\arraystretch}{1.12}
\begin{longtable}{@{}>{\raggedright\arraybackslash}p{0.302\linewidth}>{\raggedright\arraybackslash}p{0.175\linewidth}>{\raggedright\arraybackslash}p{0.155\linewidth}>{\raggedright\arraybackslash}p{0.155\linewidth}>{\raggedright\arraybackslash}p{0.155\linewidth}@{}}
\toprule
\textbf{Comparison} & \textbf{Mean APAS difference (95\% CI)} & \textbf{t (df)} & \textbf{Raw P} & \textbf{Holm P} \\
\midrule
\endfirsthead
\toprule
\textbf{Comparison} & \textbf{Mean APAS difference (95\% CI)} & \textbf{t (df)} & \textbf{Raw P} & \textbf{Holm P} \\
\midrule
\endhead
\midrule
\multicolumn{5}{r}{\footnotesize Continued on next page} \\
\endfoot
\bottomrule
\endlastfoot
Complete MCE versus without multidisciplinary priors & 16.62 (11.84--21.40) & 6.819 (39) & 3.80e-08 & 1.90e-07 \\
Complete MCE versus without pathway/\allowbreak{}evidence and review stages & 7.72 (4.87--10.57) & 5.309 (39) & 4.70e-06 & 1.15e-05 \\
Complete MCE versus without simulation, critique and repair & 3.60 (1.72--5.49) & 3.745 (39) & 5.83e-04 & 5.83e-04 \\
Complete MCE versus MedGPT base direct generation & 23.82 (15.95--31.70) & 5.929 (39) & 6.50e-07 & 2.60e-06 \\
Complete MCE versus MedGPT-LightRAG & 14.34 (9.11--19.57) & 5.373 (39) & 3.83e-06 & 1.15e-05 \\
\end{longtable}
\endgroup

Differences are complete MCE minus the comparison configuration. CIs retain the normal approximation for the mean of the 40 within-case differences. Supplementary post hoc two-sided paired t tests assessed a zero mean difference (39 df), with Holm adjustment across all five contrasts. Intervals are unadjusted and were not used to calculate P values.

\subsection{e. Stability of configuration-level rankings}

\begingroup
\small
\setlength{\tabcolsep}{3pt}
\renewcommand{\arraystretch}{1.12}
\begin{longtable}{@{}>{\raggedright\arraybackslash}p{0.410\linewidth}>{\raggedright\arraybackslash}p{0.265\linewidth}>{\raggedright\arraybackslash}p{0.265\linewidth}@{}}
\toprule
\textbf{Judge pair} & \textbf{Spearman's \ensuremath{\rho}} & \textbf{Ranking units} \\
\midrule
\endfirsthead
\toprule
\textbf{Judge pair} & \textbf{Spearman's \ensuremath{\rho}} & \textbf{Ranking units} \\
\midrule
\endhead
\midrule
\multicolumn{3}{r}{\footnotesize Continued on next page} \\
\endfoot
\bottomrule
\endlastfoot
GPT-5.5 versus Gemini 3.1 Pro Preview & 1.000 & 6 configurations \\
GPT-5.5 versus Claude Opus 4.7 & 0.943 & 6 configurations \\
Gemini 3.1 Pro Preview versus Claude Opus 4.7 & 0.943 & 6 configurations \\
\end{longtable}
\endgroup

These correlations describe stability of the aggregate ranking across six configurations. They are not case-level inter-judge agreement and do not form a pooled score.

\section{Supplementary Table 8. Relationship states, evaluation process and illustrative gap in purposive case variants}

\subsection{a. State dictionary}

\begingroup
\small
\setlength{\tabcolsep}{3pt}
\renewcommand{\arraystretch}{1.12}
\begin{longtable}{@{}>{\raggedright\arraybackslash}p{0.157\linewidth}>{\raggedright\arraybackslash}p{0.321\linewidth}>{\raggedright\arraybackslash}p{0.461\linewidth}@{}}
\toprule
\textbf{State} & \textbf{Definition} & \textbf{Main denominator handling} \\
\midrule
\endfirsthead
\toprule
\textbf{State} & \textbf{Definition} & \textbf{Main denominator handling} \\
\midrule
\endhead
\midrule
\multicolumn{3}{r}{\footnotesize Continued on next page} \\
\endfoot
\bottomrule
\endlastfoot
Pass & Output meets the prespecified requirement & Included in pass/\allowbreak{}clear-gap denominator \\
Clear gap & Output clearly omits or violates the requirement & Included in pass/\allowbreak{}clear-gap denominator \\
Clinical review & Structured judging alone does not support a reliable determination & Counted as not passing in the conservative descriptive denominator; separate in the binary-evaluable analysis \\
Not applicable & Requirement does not apply to the scenario & Reported separately; excluded from both proportion denominators \\
Not assessable & Material or returned output is insufficient for assessment & Counted as not passing in the conservative descriptive denominator; separate in the binary-evaluable analysis \\
\end{longtable}
\endgroup

\subsection{b. Escalation process}

\begingroup
\footnotesize
\setlength{\tabcolsep}{3pt}
\renewcommand{\arraystretch}{1.12}
\begin{longtable}{@{}>{\raggedright\arraybackslash}p{0.111\linewidth}>{\raggedright\arraybackslash}p{0.419\linewidth}>{\raggedright\arraybackslash}p{0.170\linewidth}>{\raggedright\arraybackslash}p{0.240\linewidth}@{}}
\toprule
\textbf{Stage} & \textbf{Entry rule} & \textbf{Rater or reviewer} & \textbf{Interpretation} \\
\midrule
\endfirsthead
\toprule
\textbf{Stage} & \textbf{Entry rule} & \textbf{Rater or reviewer} & \textbf{Interpretation} \\
\midrule
\endhead
\midrule
\multicolumn{4}{r}{\footnotesize Continued on next page} \\
\endfoot
\bottomrule
\endlastfoot
AI consensus & Strict majority of three judges; a three-state split assigned clinical review & 3 named AI judges & Structured localization of potential gaps, not a clinical reference standard \\
Review cohort & Clear gap, clinical review, judge disagreement, missing judge result, or prespecified high-priority or ungraded safety requirement & 2 physicians independently & Did not define a human reference standard \\
Third-physician escalation & All safety requirements and all physician disagreements & Third physician with access to earlier opinions & Targeted escalation, not blinded independent scoring \\
Complete scenario closure & All 6 requirements returned; at least one entered the main denominator; all in-denominator requirements passed; no clear gap, clinical review or not-assessable state & Derived separately for each rating source & Not a clinical safety or efficacy endpoint \\
\end{longtable}
\endgroup

\subsection{c. AI-consensus requirement states and descriptive denominators}

\begingroup
\footnotesize
\setlength{\tabcolsep}{3pt}
\renewcommand{\arraystretch}{1.12}
\begin{longtable}{@{}>{\raggedright\arraybackslash}p{0.151\linewidth}>{\raggedright\arraybackslash}p{0.215\linewidth}>{\raggedright\arraybackslash}p{0.151\linewidth}>{\raggedright\arraybackslash}p{0.422\linewidth}@{}}
\toprule
\textbf{Unit} & \textbf{State or analysis} & \textbf{Count} & \textbf{Denominator and interpretation} \\
\midrule
\endfirsthead
\toprule
\textbf{Unit} & \textbf{State or analysis} & \textbf{Count} & \textbf{Denominator and interpretation} \\
\midrule
\endhead
\midrule
\multicolumn{4}{r}{\footnotesize Continued on next page} \\
\endfoot
\bottomrule
\endlastfoot
Requirement & Pass & 228 & Consensus among 288 prespecified requirements \\
Requirement & Clinical review & 45 & Counted as not passing in the conservative descriptive analysis; separate in binary-evaluable analysis \\
Requirement & Clear gap & 14 & Together with pass, formed 242 binary-evaluable requirements \\
Requirement & Not applicable & 1 & Excluded from denominators 242 and 287 \\
Requirement & Not assessable & 0 & Counted as not passing in conservative analysis; none observed \\
Requirement & Conservative descriptive pass proportion & 228/\allowbreak{}287 (79.4\%) & Clinical review and not assessable counted as not passing; not applicable excluded \\
Requirement & Binary-evaluable pass proportion & 228/\allowbreak{}242 (94.2\%) & Denominator limited to pass and clear gap \\
\end{longtable}
\endgroup

The two requirement-level proportions document denominator handling and are not overall MCE performance, error or clinical-safety rates.

\subsection{d. Scenario states by condition category}

\begingroup
\scriptsize
\setlength{\tabcolsep}{3pt}
\renewcommand{\arraystretch}{1.12}
\begin{longtable}{@{}>{\raggedright\arraybackslash}p{0.226\linewidth}>{\raggedright\arraybackslash}p{0.178\linewidth}>{\raggedright\arraybackslash}p{0.178\linewidth}>{\raggedright\arraybackslash}p{0.178\linewidth}>{\raggedright\arraybackslash}p{0.178\linewidth}@{}}
\toprule
\textbf{Condition category} & \textbf{Scenarios} & \textbf{Complete closure} & \textbf{Clinical review} & \textbf{Clear gap} \\
\midrule
\endfirsthead
\toprule
\textbf{Condition category} & \textbf{Scenarios} & \textbf{Complete closure} & \textbf{Clinical review} & \textbf{Clear gap} \\
\midrule
\endhead
\midrule
\multicolumn{5}{r}{\footnotesize Continued on next page} \\
\endfoot
\bottomrule
\endlastfoot
All scenarios & 48 & 13 & 23 & 12 \\
BASE baseline anchor & 16 & 5 & 8 & 3 \\
SAFE safety constraint & 11 & 4 & 5 & 2 \\
INFO decision-critical information & 5 & 1 & 2 & 2 \\
SEQ pathway sequence & 8 & 3 & 4 & 1 \\
GOV reassessment and governance & 8 & 0 & 4 & 4 \\
\end{longtable}
\endgroup

Scenario states were jointly derived from the six requirements in each scenario and use a different unit and decision rule from requirement-level proportions.

\subsection{e. Non-exclusive mechanisms among clear relationship gaps}

\begingroup
\footnotesize
\setlength{\tabcolsep}{3pt}
\renewcommand{\arraystretch}{1.12}
\begin{longtable}{@{}>{\raggedright\arraybackslash}p{0.303\linewidth}>{\raggedright\arraybackslash}p{0.212\linewidth}>{\raggedright\arraybackslash}p{0.212\linewidth}>{\raggedright\arraybackslash}p{0.212\linewidth}@{}}
\toprule
\textbf{Mechanism} & \textbf{Mechanism labels} & \textbf{Scenario cards} & \textbf{Source templates} \\
\midrule
\endfirsthead
\toprule
\textbf{Mechanism} & \textbf{Mechanism labels} & \textbf{Scenario cards} & \textbf{Source templates} \\
\midrule
\endhead
\midrule
\multicolumn{4}{r}{\footnotesize Continued on next page} \\
\endfoot
\bottomrule
\endlastfoot
Pathway prerequisite unmet & 8 & 8 & 6 \\
Sequence or fallback violation & 6 & 6 & 5 \\
Core clinical problem not covered & 2 & 2 & 1 \\
Decision-critical information unresolved & 2 & 2 & 1 \\
Safety-constraint relationship unmet & 2 & 2 & 1 \\
\end{longtable}
\endgroup

A clear gap could carry multiple mechanism labels. Counts of labels therefore do not equal numbers of gap rules, scenarios or independent clinical events.

\subsection{f. Agreement in complete independent re-review and comparison with AI consensus}

\begingroup
\scriptsize
\setlength{\tabcolsep}{3pt}
\renewcommand{\arraystretch}{1.12}
\begin{longtable}{@{}>{\raggedright\arraybackslash}p{0.205\linewidth}>{\raggedright\arraybackslash}p{0.173\linewidth}>{\raggedright\arraybackslash}p{0.215\linewidth}>{\raggedright\arraybackslash}p{0.173\linewidth}>{\raggedright\arraybackslash}p{0.173\linewidth}@{}}
\toprule
\textbf{Comparison} & \textbf{Level} & \textbf{Objects} & \textbf{Exact agreement} & \textbf{Gwet AC1 (95\% CI)} \\
\midrule
\endfirsthead
\toprule
\textbf{Comparison} & \textbf{Level} & \textbf{Objects} & \textbf{Exact agreement} & \textbf{Gwet AC1 (95\% CI)} \\
\midrule
\endhead
\midrule
\multicolumn{5}{r}{\footnotesize Continued on next page} \\
\endfoot
\bottomrule
\endlastfoot
Physician 1 versus physician 2 & Rule & 288 CCRRs; 16 source templates & 67.01\% & 0.607 (0.530--0.684) \\
Physician 1 versus physician 2 & Scenario card & 48 scenarios; 16 source templates & 93.75\% & 0.892 (0.757--1.000) \\
Physician 1 versus AI consensus & Rule & 288 CCRRs; 16 source templates & 39.93\% & 0.301 (0.204--0.402) \\
Physician 2 versus AI consensus & Rule & 288 CCRRs; 16 source templates & 39.93\% & 0.303 (0.199--0.399) \\
Physician 1 versus AI consensus & Scenario card & 48 scenarios; 16 source templates & 58.33\% & 0.311 (\ensuremath{-}0.050 to 0.618) \\
Physician 2 versus AI consensus & Scenario card & 48 scenarios; 16 source templates & 56.25\% & 0.244 (\ensuremath{-}0.149 to 0.600) \\
\end{longtable}
\endgroup

Results use independent, pre-adjudication physician ratings and same-object comparisons with AI consensus. Rule-level and scenario-card-level analyses use different units. Physician--AI comparisons diagnose evaluator dependence and do not assign either side as the reference standard.

\subsection{g. Descriptive correspondence between BASE and modified-scenario states}

\begingroup
\scriptsize
\setlength{\tabcolsep}{3pt}
\renewcommand{\arraystretch}{1.12}
\begin{longtable}{@{}>{\raggedright\arraybackslash}p{0.226\linewidth}>{\raggedright\arraybackslash}p{0.178\linewidth}>{\raggedright\arraybackslash}p{0.178\linewidth}>{\raggedright\arraybackslash}p{0.178\linewidth}>{\raggedright\arraybackslash}p{0.178\linewidth}@{}}
\toprule
\textbf{Modification category} & \textbf{Pairs} & \textbf{Toward complete closure} & \textbf{Unchanged} & \textbf{Toward clear gap} \\
\midrule
\endfirsthead
\toprule
\textbf{Modification category} & \textbf{Pairs} & \textbf{Toward complete closure} & \textbf{Unchanged} & \textbf{Toward clear gap} \\
\midrule
\endhead
\midrule
\multicolumn{5}{r}{\footnotesize Continued on next page} \\
\endfoot
\bottomrule
\endlastfoot
SAFE safety constraint & 11 & 3 & 6 & 2 \\
INFO decision-critical information & 5 & 0 & 3 & 2 \\
SEQ pathway sequence & 8 & 2 & 3 & 3 \\
GOV reassessment and governance & 8 & 1 & 4 & 3 \\
\end{longtable}
\endgroup

Directions describe correspondence between BASE and modified scenarios from the same source template and are not attributed causally to the changed case condition. Model snapshots, prompts, adapters, replicate identifiers and random conditions were not completely locked across the paired scenarios.

\subsection{h. Requirement states by prespecified safety label}

\begingroup
\scriptsize
\setlength{\tabcolsep}{3pt}
\renewcommand{\arraystretch}{1.12}
\begin{longtable}{@{}>{\raggedright\arraybackslash}p{0.222\linewidth}>{\raggedright\arraybackslash}p{0.179\linewidth}>{\raggedright\arraybackslash}p{0.179\linewidth}>{\raggedright\arraybackslash}p{0.179\linewidth}>{\raggedright\arraybackslash}p{0.179\linewidth}@{}}
\toprule
\textbf{Safety label} & \textbf{Requirements} & \textbf{Pass} & \textbf{Clinical review} & \textbf{Clear gap} \\
\midrule
\endfirsthead
\toprule
\textbf{Safety label} & \textbf{Requirements} & \textbf{Pass} & \textbf{Clinical review} & \textbf{Clear gap} \\
\midrule
\endhead
\midrule
\multicolumn{5}{r}{\footnotesize Continued on next page} \\
\endfoot
\bottomrule
\endlastfoot
P0 & 9 & 7 & 0 & 2 \\
P1 & 24 & 23 & 1 & 0 \\
Ungraded safety condition & 12 & 10 & 2 & 0 \\
No case-specific P0/\allowbreak{}P1 identified & 3 & 3 & 0 & 0 \\
\end{longtable}
\endgroup

Safety labels define requirement strata; their state distributions are not clinical safety rates or patient-risk estimates.

\section{Supplementary Table 9. Statistical models, diagnostics, fallback procedures and sensitivity analyses}

\subsection{Supplementary Table 9a. Models and sensitivity analyses}

\begingroup
\scriptsize
\setlength{\tabcolsep}{3pt}
\renewcommand{\arraystretch}{1.12}
\begin{longtable}{@{}>{\raggedright\arraybackslash}p{0.097\linewidth}>{\raggedright\arraybackslash}p{0.221\linewidth}>{\raggedright\arraybackslash}p{0.221\linewidth}>{\raggedright\arraybackslash}p{0.207\linewidth}>{\raggedright\arraybackslash}p{0.196\linewidth}@{}}
\toprule
\textbf{Analysis} & \textbf{Model or method} & \textbf{Unit or clustering} & \textbf{Diagnostics and fallback} & \textbf{Multiplicity or interval} \\
\midrule
\endfirsthead
\toprule
\textbf{Analysis} & \textbf{Model or method} & \textbf{Unit or clustering} & \textbf{Diagnostics and fallback} & \textbf{Multiplicity or interval} \\
\midrule
\endhead
\midrule
\multicolumn{5}{r}{\footnotesize Continued on next page} \\
\endfoot
\bottomrule
\endlastfoot
Primary APAS & \texttt{APAS \textasciitilde{} condition \ensuremath{\times} case + task position + (1 | physician)} & Physician--case--condition response & REML; singularity tolerance 1e-4; convergence and warnings recorded & Holm adjustment for M--U and M--R P values; Wald 95\% CI \\
APAS standardization & Equally weighted 40-case \ensuremath{\times} 9-position reference grid & Fixed-case empirical target set & Physician random effect set to 0 & Delta-method standard error \\
CSPR & \texttt{CSPR \textasciitilde{} condition + case + task position + (1 | physician)}; fixed effects transformed with expit and averaged over the 40 \ensuremath{\times} 9 grid & 2,237 evaluable responses & Common-condition-effect model used after instability of the initial case-by-condition binomial model; available diagnostics did not identify a more specific trigger & Probability differences and delta-method standard errors; Holm adjustment for M--U and M--R P values \\
Judge stability & Nine judge-run models, equal-weight panel and leave-one-judge-out panels & Same 2,250 responses & Repeated ratings were not independent samples & Direction and interval reported separately \\
Professional-title-group analysis & Condition \ensuremath{\times} case + condition \ensuremath{\times} professional-title group + task position + physician random effect & Physician--case response & Joint four-degree-of-freedom Wald test & Absence of detected interaction was not interpreted as equivalence \\
M1--M5 domains & Case-heterogeneous models with the APAS structure & Responses with an applicable domain score & Non-applicable records not imputed & Holm adjustment across ten contrasts; interactions post hoc \\
Post hoc subsequent-event analysis & Case-specific condition estimates derived from the APAS model fitted to all 40 cases and equally standardized within each event target set; within-group contrasts contextualized the main results and event-group-by-condition interactions were secondary exploratory estimates & Main data-quality sensitivity target: 12 cases with a component event record and 19 with follow-up but no recorded component event after excluding 3 records with chronology, index-date or record-linkage concerns; 6 cases with unavailable follow-up retained separately & Follow-up groups were derived from non-structured text by post hoc rules without independent clinical adjudication; not OS, PFS or a validated composite endpoint & Post hoc, no multiplicity adjustment; asymptotic Wald 95\% CIs; no confirmatory effect-modification claim based on inclusion or exclusion of zero \\
Format-aligned MCE comparison & Equal-weight average of nine ratings at response level; physician responses then averaged within case and condition; paired system-versus-case-mean comparison & 40 equally weighted cases, retaining system output, physician responses and repeated ratings within case & Output generated within MCE using the physician task's five-part structure and length constraint & Signed differences; 30,000 case-cluster percentile bootstrap samples, seed 20260907; supplementary post hoc paired t tests, Holm across all 21 contrasts \\
Crossed-random-intercept sensitivity & \texttt{outcome \textasciitilde{} condition + task position + (1 | physician) + (1 | case)} & Physician--case response & Alternative case modeling & Compared with primary direction \\
Common-applicability sensitivity & Rules applicable in all nine ratings only & Physician response & Assessed applicability selection & Compared with primary direction \\
Input-adaptation sensitivity & Excluded 94 responses containing system-specific pathway identifiers & 2,156 responses & Refitted each judge-run model & Compared with primary direction \\
Length diagnostic & Condition \ensuremath{\times} centered log length & Physician response & Length was post-assignment and not a primary-model covariate & Diagnostic \\
Source-masked agreement & 3 \ensuremath{\times} 3 table, exact agreement, Gwet AC1 and \ensuremath{\kappa}; positive and negative agreement for safety flags & Fixed 60-module re-review set & Ties and sparse categories retained; Wilson intervals for proportions & Agreement coefficients interpreted for the fixed set \\
Overall clinical anchoring & Exact agreement and Cohen \ensuremath{\kappa} for the five-category acceptability item; agreement and linearly weighted \ensuremath{\kappa} for the four-category ordinal sensitivity outcome; \ensuremath{\kappa} for major clinical defects; Spearman's \ensuremath{\rho} between APAS and mean four-category acceptability & 120 outputs, 40 case clusters, 2 physicians & Original five-category item was primary; four-category outcome was supplementary; cluster-percentile bootstrap & 95\% CIs; source comparisons descriptive \\
Controlled-text perturbation APAS & Mean paired triplet difference, variant minus original; stratified by target domain M2--M5 & 36 triplets clustered within 12 cases & Text panel value averaged 3 judges \ensuremath{\times} 3 repeats; non-applicable, missing and parse-failure states retained & 30,000 case-cluster percentile bootstrap samples; 108 texts not treated as independent \\
Controlled-text perturbation partial gates or rule states & Within-triplet state transitions and target-rule score changes derived from the six rule links supplied in each request & Same 36 triplets clustered within 12 cases & Supplementary; not equivalent to complete nine-gate CSPR & Original denominators reported; no additional confirmatory test \\
EASS rule re-rating of complete MCE outputs & Ordinal-rule AC2/\allowbreak{}weighted \ensuremath{\kappa}; case-level ICC(A,1), Lin CCC, mean difference and Bland--Altman limits & 40 cases; paired-complete rules or cases & Not-assessable states not recoded; case-cluster resampling & 10,000 case-cluster percentile bootstrap samples \\
CCRR physician calibration & Five-state AC1/\allowbreak{}\ensuremath{\kappa} and binary complete-scenario AC1 & 144 CCRRs, 24 scenarios, 16 source templates & All five states retained; scenario result derived by the main rule & 10,000 source-template cluster bootstrap samples \\
Complete CCRR physician re-review & Five-state exact agreement and Gwet AC1; binary scenario-closure agreement and AC1; physician--AI same-object diagnostics & 288 CCRRs, 48 scenarios, 16 source templates & Independent pre-adjudication judgments; rule and card levels separate & 10,000 source-template cluster bootstrap samples \\
Six-configuration functional-module ablation analysis & Within-case paired mean difference and 95\% CI; supplementary paired t test; Spearman's rank correlation & 40 cases; ranking n=6 configurations & Single generation per case--configuration; grouped-ablation or reduced-orchestration reference interpretation & Post hoc two-sided tests; Holm across five contrasts; no confirmatory hypothesis test \\
CCRR pass proportion & Calculated within template and equally averaged across 16 templates & Source template & Conservative main analysis counted clinical review/\allowbreak{}not assessable as not passing; binary-evaluable proportion supplementary & Wilson or template-cluster bootstrap 95\% CI \\
CCRR paired difference & Modified versus baseline scenario, equally weighted within and across templates & 16 source templates & 10,000 cluster bootstrap samples, seed 20260812 & No P value \\
\end{longtable}
\endgroup

For partial gate states derived from the six selected rule links supplied in each request, the paired difference was +0.03 (95\% CI, 0.01--0.05) for section-order variants and \ensuremath{-}0.02 (\ensuremath{-}0.05 to 0.01) for target-content-deletion variants. These transitions used the same triplets and applicability denominators and were supplementary descriptions rather than validation of the complete nine-gate CSPR.

U/R/M mixed models used R 4.6.1, \texttt{lme4} 2.0-6 and \texttt{emmeans} 2.0.4. Agreement analyses used Python 3.12.13, pandas 2.2.3 and NumPy 2.3.5.

\textbf{Primary APAS support contrasts (Fig. 3a)}

\begingroup
\small
\setlength{\tabcolsep}{3pt}
\renewcommand{\arraystretch}{1.12}
\begin{longtable}{@{}>{\raggedright\arraybackslash}p{0.290\linewidth}>{\raggedright\arraybackslash}p{0.360\linewidth}>{\raggedright\arraybackslash}p{0.290\linewidth}@{}}
\toprule
\textbf{Contrast} & \textbf{Adjusted APAS difference (95\% CI)} & \textbf{Holm-adjusted P value} \\
\midrule
\endfirsthead
\toprule
\textbf{Contrast} & \textbf{Adjusted APAS difference (95\% CI)} & \textbf{Holm-adjusted P value} \\
\midrule
\endhead
\midrule
\multicolumn{3}{r}{\footnotesize Continued on next page} \\
\endfoot
\bottomrule
\endlastfoot
M--U & 12.87 (11.18 to 14.55) & \ensuremath{8.49\times 10^{-51}} \\
M--R & 5.22 (3.52 to 6.93) & \ensuremath{1.98\times 10^{-9}} \\
\end{longtable}
\endgroup

These are the existing physician-task mixed-model comparisons, with Holm adjustment across the two APAS contrasts. They are not supplementary paired t tests; the original unadjusted 95\% CIs are retained.

\begin{landscape}
\subsection{Supplementary Table 9b. Exploratory APAS differences in M1--M5 clinical content domains}

\begingroup
\scriptsize
\setlength{\tabcolsep}{3pt}
\renewcommand{\arraystretch}{1.12}
\begin{longtable}{@{}>{\raggedright\arraybackslash}p{0.202\linewidth}>{\raggedright\arraybackslash}p{0.142\linewidth}>{\raggedright\arraybackslash}p{0.142\linewidth}>{\raggedright\arraybackslash}p{0.142\linewidth}>{\raggedright\arraybackslash}p{0.142\linewidth}>{\raggedright\arraybackslash}p{0.172\linewidth}@{}}
\toprule
\textbf{Domain} & \textbf{Contrast} & \textbf{APAS difference} & \textbf{95\% CI} & \textbf{Holm-adjusted P value} & \textbf{Evaluable responses by condition} \\
\midrule
\endfirsthead
\toprule
\textbf{Domain} & \textbf{Contrast} & \textbf{APAS difference} & \textbf{95\% CI} & \textbf{Holm-adjusted P value} & \textbf{Evaluable responses by condition} \\
\midrule
\endhead
\midrule
\multicolumn{6}{r}{\footnotesize Continued on next page} \\
\endfoot
\bottomrule
\endlastfoot
M1 clinical problem and treatment goals & M--U & 13.10 & 11.07--15.12 & \ensuremath{6.79\times 10^{-36}} & U 750; R 750; M 750 \\
M1 clinical problem and treatment goals & M--R & 4.37 & 2.32--6.42 & \ensuremath{7.57\times 10^{-5}} & U 750; R 750; M 750 \\
M2 decision-critical information & M--U & 8.59 & 6.35--10.83 & \ensuremath{4.01\times 10^{-13}} & U 429; R 465; M 505 \\
M2 decision-critical information & M--R & 4.88 & 2.61--7.15 & \ensuremath{7.57\times 10^{-5}} & U 429; R 465; M 505 \\
M3 candidate clinical pathways & M--U & 20.64 & 18.15--23.13 & \ensuremath{3.60\times 10^{-58}} & U 742; R 731; M 735 \\
M3 candidate clinical pathways & M--R & 8.59 & 6.07--11.12 & \ensuremath{1.61\times 10^{-10}} & U 742; R 731; M 735 \\
M4 safety boundaries and risk governance & M--U & 12.42 & 10.40--14.43 & \ensuremath{1.22\times 10^{-32}} & U 591; R 620; M 628 \\
M4 safety boundaries and risk governance & M--R & 4.72 & 2.68--6.76 & \ensuremath{2.41\times 10^{-5}} & U 591; R 620; M 628 \\
M5 reassessment and governance & M--U & 4.14 & 2.50--5.79 & \ensuremath{3.86\times 10^{-6}} & U 750; R 750; M 750 \\
M5 reassessment and governance & M--R & 3.47 & 1.80--5.13 & \ensuremath{7.57\times 10^{-5}} & U 750; R 750; M 750 \\
\end{longtable}
\endgroup

Domain scores were calculated only for responses in which the domain was applicable and entered the denominator. Non-applicable records were not imputed. Holm adjustment covered all ten comparisons; these analyses describe the clinical-content distribution of the overall APAS difference.

\end{landscape}
\subsection{Supplementary Table 9c. Exploratory APAS differences by professional title}

\begingroup
\scriptsize
\setlength{\tabcolsep}{3pt}
\renewcommand{\arraystretch}{1.12}
\begin{longtable}{@{}>{\raggedright\arraybackslash}p{0.195\linewidth}>{\raggedright\arraybackslash}p{0.186\linewidth}>{\raggedright\arraybackslash}p{0.186\linewidth}>{\raggedright\arraybackslash}p{0.186\linewidth}>{\raggedright\arraybackslash}p{0.186\linewidth}@{}}
\toprule
\textbf{Professional-title stratum} & \textbf{Physicians} & \textbf{Contrast} & \textbf{APAS difference} & \textbf{95\% CI} \\
\midrule
\endfirsthead
\toprule
\textbf{Professional-title stratum} & \textbf{Physicians} & \textbf{Contrast} & \textbf{APAS difference} & \textbf{95\% CI} \\
\midrule
\endhead
\midrule
\multicolumn{5}{r}{\footnotesize Continued on next page} \\
\endfoot
\bottomrule
\endlastfoot
Senior & 68 & M--U & 14.61 & 11.36--17.86 \\
Senior & 68 & M--R & 5.16 & 1.89--8.43 \\
Intermediate & 91 & M--U & 11.68 & 8.89--14.48 \\
Intermediate & 91 & M--R & 4.24 & 1.43--7.04 \\
Junior & 91 & M--U & 12.76 & 9.97--15.54 \\
Junior & 91 & M--R & 6.22 & 3.42--9.03 \\
\end{longtable}
\endgroup

The joint four-degree-of-freedom Wald test for condition by professional title gave P = 0.548. Professional title was a coarse proxy for experience; estimates and the interaction test were exploratory and did not establish equivalence between strata.

\begin{landscape}
\subsection{Supplementary Table 9d. Judge-specific APAS and CSPR for complete MCE and other-model strategy outputs}

\textbf{a. Overall APAS and CSPR}

\begingroup
\scriptsize
\setlength{\tabcolsep}{3pt}
\renewcommand{\arraystretch}{1.12}
\begin{longtable}{@{}>{\raggedright\arraybackslash}p{0.157\linewidth}>{\raggedright\arraybackslash}p{0.157\linewidth}>{\raggedright\arraybackslash}p{0.157\linewidth}>{\raggedright\arraybackslash}p{0.157\linewidth}>{\raggedright\arraybackslash}p{0.157\linewidth}>{\raggedright\arraybackslash}p{0.157\linewidth}@{}}
\toprule
\textbf{Output} & \textbf{AI judge} & \textbf{Mean APAS} & \textbf{95\% CI} & \textbf{CSPR passing/\allowbreak{}40} & \textbf{CSPR pass rate} \\
\midrule
\endfirsthead
\toprule
\textbf{Output} & \textbf{AI judge} & \textbf{Mean APAS} & \textbf{95\% CI} & \textbf{CSPR passing/\allowbreak{}40} & \textbf{CSPR pass rate} \\
\midrule
\endhead
\midrule
\multicolumn{6}{r}{\footnotesize Continued on next page} \\
\endfoot
\bottomrule
\endlastfoot
Complete MCE & Gemini 3.1 Pro Preview & 96.23 & 93.64--98.82 & 36/\allowbreak{}40 & 90.0\% \\
Complete MCE & GPT-5.5 & 94.71 & 91.35--98.07 & 36/\allowbreak{}40 & 90.0\% \\
Complete MCE & Claude Opus 4.7 & 97.72 & 95.94--99.50 & 39/\allowbreak{}40 & 97.5\% \\
GPT-5.4 & Gemini 3.1 Pro Preview & 71.06 & 62.33--79.78 & 29/\allowbreak{}40 & 72.5\% \\
GPT-5.4 & GPT-5.5 & 68.75 & 61.30--76.20 & 27/\allowbreak{}40 & 67.5\% \\
GPT-5.4 & Claude Opus 4.7 & 74.34 & 67.40--81.28 & 35/\allowbreak{}40 & 87.5\% \\
Claude Opus 4.7 & Gemini 3.1 Pro Preview & 61.54 & 52.02--71.07 & 20/\allowbreak{}40 & 50.0\% \\
Claude Opus 4.7 & GPT-5.5 & 56.91 & 47.93--65.89 & 19/\allowbreak{}40 & 47.5\% \\
Claude Opus 4.7 & Claude Opus 4.7 & 66.99 & 59.37--74.60 & 30/\allowbreak{}40 & 75.0\% \\
Gemini 3.1 Pro & Gemini 3.1 Pro Preview & 55.47 & 45.47--65.46 & 18/\allowbreak{}40 & 45.0\% \\
Gemini 3.1 Pro & GPT-5.5 & 53.16 & 45.38--60.95 & 19/\allowbreak{}40 & 47.5\% \\
Gemini 3.1 Pro & Claude Opus 4.7 & 55.22 & 47.81--62.63 & 25/\allowbreak{}40 & 62.5\% \\
\end{longtable}
\endgroup

\textbf{b. M1--M5 domain-specific mean APAS}

\begingroup
\scriptsize
\setlength{\tabcolsep}{3pt}
\renewcommand{\arraystretch}{1.12}
\begin{longtable}{@{}>{\raggedright\arraybackslash}p{0.134\linewidth}>{\raggedright\arraybackslash}p{0.134\linewidth}>{\raggedright\arraybackslash}p{0.134\linewidth}>{\raggedright\arraybackslash}p{0.134\linewidth}>{\raggedright\arraybackslash}p{0.134\linewidth}>{\raggedright\arraybackslash}p{0.134\linewidth}>{\raggedright\arraybackslash}p{0.134\linewidth}@{}}
\toprule
\textbf{Output} & \textbf{AI judge} & \textbf{M1} & \textbf{M2} & \textbf{M3} & \textbf{M4} & \textbf{M5} \\
\midrule
\endfirsthead
\toprule
\textbf{Output} & \textbf{AI judge} & \textbf{M1} & \textbf{M2} & \textbf{M3} & \textbf{M4} & \textbf{M5} \\
\midrule
\endhead
\midrule
\multicolumn{7}{r}{\footnotesize Continued on next page} \\
\endfoot
\bottomrule
\endlastfoot
Complete MCE & Gemini 3.1 Pro Preview & 96.25 (92.51--99.99) & 96.05 (90.35--101.76)\textsuperscript{\dag} & 93.75 (87.49--100.01) & 97.50 (92.60--102.40) & 98.75 (96.30--101.20) \\
Complete MCE & GPT-5.5 & 94.38 (89.91--98.84) & 96.25 (90.83--101.67) & 91.25 (83.49--99.01) & 96.25 (90.83--101.67) & 96.25 (92.12--100.38) \\
Complete MCE & Claude Opus 4.7 & 99.38 (98.15--100.60) & 97.50 (94.08--100.92) & 95.00 (89.13--100.87) & 97.50 (94.08--100.92) & 97.50 (94.08--100.92) \\
GPT-5.4 & Gemini 3.1 Pro Preview & 74.38 (64.53--84.22) & 76.32 (64.78--87.85)\textsuperscript{\dag} & 53.75 (42.44--65.06) & 72.50 (60.89--84.11) & 77.50 (68.25--86.75) \\
GPT-5.4 & GPT-5.5 & 78.12 (68.65--87.60) & 70.00 (59.59--80.41) & 52.50 (40.89--64.11) & 70.00 (59.59--80.41) & 61.25 (53.82--68.68) \\
GPT-5.4 & Claude Opus 4.7 & 83.12 (75.20--91.05) & 71.25 (62.74--79.76) & 66.25 (55.50--77.00) & 72.50 (63.94--81.06) & 67.50 (59.23--75.77) \\
Claude Opus 4.7 & Gemini 3.1 Pro Preview & 62.50 (51.13--73.87) & 64.47 (50.69--78.26)\textsuperscript{\dag} & 43.75 (32.55--54.95) & 71.25 (59.15--83.35) & 67.50 (56.10--78.90) \\
Claude Opus 4.7 & GPT-5.5 & 67.50 (57.54--77.46) & 56.25 (44.00--68.50) & 38.75 (26.35--51.15) & 55.00 (42.45--67.55) & 56.25 (45.05--67.45) \\
Claude Opus 4.7 & Claude Opus 4.7 & 71.88 (63.43--80.32) & 65.00 (55.58--74.42) & 56.25 (44.51--67.99) & 66.25 (57.38--75.12) & 72.50 (63.94--81.06) \\
Gemini 3.1 Pro & Gemini 3.1 Pro Preview & 60.62 (50.11--71.14) & 56.58 (42.18--70.98)\textsuperscript{\dag} & 37.50 (26.00--49.00) & 66.25 (53.41--79.09) & 50.00 (36.41--63.59) \\
Gemini 3.1 Pro & GPT-5.5 & 68.12 (59.18--77.07) & 47.50 (35.89--59.11) & 38.75 (26.86--50.64) & 48.75 (37.39--60.11) & 45.00 (35.85--54.15) \\
Gemini 3.1 Pro & Claude Opus 4.7 & 64.38 (55.11--73.64) & 42.50 (33.52--51.48) & 45.00 (33.47--56.53) & 60.00 (50.58--69.42) & 55.00 (48.16--61.84) \\
\end{longtable}
\endgroup

\textbf{c. Within-case M1--M5 differences between complete MCE and other-model strategy outputs}

\begingroup
\scriptsize
\setlength{\tabcolsep}{3pt}
\renewcommand{\arraystretch}{1.12}
\begin{longtable}{@{}>{\raggedright\arraybackslash}p{0.164\linewidth}>{\raggedright\arraybackslash}p{0.129\linewidth}>{\raggedright\arraybackslash}p{0.129\linewidth}>{\raggedright\arraybackslash}p{0.129\linewidth}>{\raggedright\arraybackslash}p{0.129\linewidth}>{\raggedright\arraybackslash}p{0.129\linewidth}>{\raggedright\arraybackslash}p{0.129\linewidth}@{}}
\toprule
\textbf{Comparison} & \textbf{AI judge} & \textbf{M1} & \textbf{M2} & \textbf{M3} & \textbf{M4} & \textbf{M5} \\
\midrule
\endfirsthead
\toprule
\textbf{Comparison} & \textbf{AI judge} & \textbf{M1} & \textbf{M2} & \textbf{M3} & \textbf{M4} & \textbf{M5} \\
\midrule
\endhead
\midrule
\multicolumn{7}{r}{\footnotesize Continued on next page} \\
\endfoot
\bottomrule
\endlastfoot
Complete MCE minus GPT-5.4 & Gemini 3.1 Pro Preview & 21.88 (11.77--31.98) & 19.74 (6.65--32.83)\textsuperscript{\dag} & 40.00 (25.45--54.55) & 25.00 (11.87--38.13) & 21.25 (12.04--30.46) \\
Complete MCE minus GPT-5.4 & GPT-5.5 & 16.25 (5.64--26.86) & 26.25 (15.73--36.77) & 38.75 (24.94--52.56) & 26.25 (15.73--36.77) & 35.00 (27.00--43.00) \\
Complete MCE minus GPT-5.4 & Claude Opus 4.7 & 16.25 (8.30--24.20) & 26.25 (17.66--34.84) & 28.75 (17.72--39.78) & 25.00 (16.40--33.60) & 30.00 (21.55--38.45) \\
Complete MCE minus Claude Opus 4.7 & Gemini 3.1 Pro Preview & 33.75 (22.58--44.92) & 31.58 (16.15--47.01)\textsuperscript{\dag} & 50.00 (35.96--64.04) & 26.25 (12.67--39.83) & 31.25 (19.78--42.72) \\
Complete MCE minus Claude Opus 4.7 & GPT-5.5 & 26.88 (15.45--38.30) & 40.00 (27.25--52.75) & 52.50 (37.22--67.78) & 41.25 (28.17--54.33) & 40.00 (27.25--52.75) \\
Complete MCE minus Claude Opus 4.7 & Claude Opus 4.7 & 27.50 (18.94--36.06) & 32.50 (22.86--42.14) & 38.75 (26.35--51.15) & 31.25 (22.18--40.32) & 25.00 (15.08--34.92) \\
Complete MCE minus Gemini 3.1 Pro & Gemini 3.1 Pro Preview & 35.62 (25.26--45.99) & 39.47 (23.30--55.65)\textsuperscript{\dag} & 56.25 (40.88--71.62) & 31.25 (16.92--45.58) & 48.75 (34.94--62.56) \\
Complete MCE minus Gemini 3.1 Pro & GPT-5.5 & 26.25 (15.73--36.77) & 48.75 (36.35--61.15) & 52.50 (36.83--68.17) & 47.50 (35.37--59.63) & 51.25 (41.65--60.85) \\
Complete MCE minus Gemini 3.1 Pro & Claude Opus 4.7 & 35.00 (25.58--44.42) & 55.00 (45.20--64.80) & 50.00 (37.35--62.65) & 37.50 (27.73--47.27) & 42.50 (34.23--50.77) \\
\end{longtable}
\endgroup

Values in panels b and c are means (95\% CIs). Each cell in panel a used 40 paired cases; APAS was the arithmetic case-level mean, with 95\% CI calculated as mean \ensuremath{\pm} 1.96 \ensuremath{\times} sample standard deviation / \ensuremath{\sqrt{40}}. Domain APAS retained the original rule weights and included applicable rules only. \textsuperscript{\dag}For Gemini 3.1 Pro Preview, the M2 rule was not applicable across all four strategies in two cases, giving n=38 for those estimates; all other domain estimates used n=40. Panel c reports complete MCE minus other-model strategy output differences within judge and case. Normal-approximation intervals were not truncated at the 0--100 scale boundaries. Each judge--output combination was rated once. The three judges were reported separately, without cross-judge pooling or the nine-run U/R/M aggregation. No between-domain test was performed. Supplementary post hoc two-sided paired t tests used separate complete Holm families for nine overall APAS contrasts and 45 domain contrasts. The original normal-approximation intervals remain unadjusted. These supportive comparisons do not isolate the effect of a single MCE component or estimate clinical safety.

\textbf{d. Supplementary overall APAS paired comparisons}

\begingroup
\scriptsize
\setlength{\tabcolsep}{3pt}
\renewcommand{\arraystretch}{1.12}
\begin{longtable}{@{}>{\raggedright\arraybackslash}p{0.175\linewidth}>{\raggedright\arraybackslash}p{0.128\linewidth}>{\raggedright\arraybackslash}p{0.128\linewidth}>{\raggedright\arraybackslash}p{0.128\linewidth}>{\raggedright\arraybackslash}p{0.128\linewidth}>{\raggedright\arraybackslash}p{0.128\linewidth}>{\raggedright\arraybackslash}p{0.128\linewidth}@{}}
\toprule
\textbf{Comparison (complete MCE minus output)} & \textbf{AI judge} & \textbf{Paired cases} & \textbf{APAS difference (95\% CI)} & \textbf{t (df)} & \textbf{Raw P} & \textbf{Holm P} \\
\midrule
\endfirsthead
\toprule
\textbf{Comparison (complete MCE minus output)} & \textbf{AI judge} & \textbf{Paired cases} & \textbf{APAS difference (95\% CI)} & \textbf{t (df)} & \textbf{Raw P} & \textbf{Holm P} \\
\midrule
\endhead
\midrule
\multicolumn{7}{r}{\footnotesize Continued on next page} \\
\endfoot
\bottomrule
\endlastfoot
GPT-5.4 & Gemini 3.1 Pro Preview & 40 & 25.17 (15.90 to 34.44) & 5.322 (39) & 4.51e-06 & 4.51e-06 \\
Claude Opus 4.7 & Gemini 3.1 Pro Preview & 40 & 34.69 (24.89 to 44.48) & 6.937 (39) & 2.61e-08 & 1.04e-07 \\
Gemini 3.1 Pro & Gemini 3.1 Pro Preview & 40 & 40.76 (30.17 to 51.35) & 7.546 (39) & 3.86e-09 & 2.32e-08 \\
GPT-5.4 & GPT-5.5 & 40 & 25.96 (17.55 to 34.36) & 6.055 (39) & 4.34e-07 & 8.68e-07 \\
Claude Opus 4.7 & GPT-5.5 & 40 & 37.79 (27.87 to 47.72) & 7.463 (39) & 5.00e-09 & 2.50e-08 \\
Gemini 3.1 Pro & GPT-5.5 & 40 & 41.55 (32.43 to 50.66) & 8.938 (39) & 5.57e-11 & 4.45e-10 \\
GPT-5.4 & Claude Opus 4.7 & 40 & 23.38 (16.26 to 30.51) & 6.434 (39) & 1.29e-07 & 3.88e-07 \\
Claude Opus 4.7 & Claude Opus 4.7 & 40 & 30.74 (23.04 to 38.43) & 7.828 (39) & 1.61e-09 & 1.13e-08 \\
Gemini 3.1 Pro & Claude Opus 4.7 & 40 & 42.50 (34.62 to 50.38) & 10.573 (39) & 5.16e-13 & 4.64e-12 \\
\end{longtable}
\endgroup

All nine overall contrasts form one complete Holm family. Differences are paired within judge and case; no judge pooling was used. The paired-difference CIs use the same normal approximation as the original summaries.

\textbf{e. Supplementary domain paired tests}

\begingroup
\scriptsize
\setlength{\tabcolsep}{3pt}
\renewcommand{\arraystretch}{1.12}
\begin{longtable}{@{}>{\raggedright\arraybackslash}p{0.175\linewidth}>{\raggedright\arraybackslash}p{0.128\linewidth}>{\raggedright\arraybackslash}p{0.128\linewidth}>{\raggedright\arraybackslash}p{0.128\linewidth}>{\raggedright\arraybackslash}p{0.128\linewidth}>{\raggedright\arraybackslash}p{0.128\linewidth}>{\raggedright\arraybackslash}p{0.128\linewidth}@{}}
\toprule
\textbf{Comparison (complete MCE minus output)} & \textbf{AI judge} & \textbf{Domain} & \textbf{Paired cases} & \textbf{t (df)} & \textbf{Raw P} & \textbf{Holm P} \\
\midrule
\endfirsthead
\toprule
\textbf{Comparison (complete MCE minus output)} & \textbf{AI judge} & \textbf{Domain} & \textbf{Paired cases} & \textbf{t (df)} & \textbf{Raw P} & \textbf{Holm P} \\
\midrule
\endhead
\midrule
\multicolumn{7}{r}{\footnotesize Continued on next page} \\
\endfoot
\bottomrule
\endlastfoot
GPT-5.4 & Gemini 3.1 Pro Preview & M1 & 40 & 4.242 (39) & 1.32e-04 & 9.60e-04 \\
GPT-5.4 & Gemini 3.1 Pro Preview & M2 & 38 & 2.956 (37) & 0.0054 & 0.0093 \\
GPT-5.4 & Gemini 3.1 Pro Preview & M3 & 40 & 5.387 (39) & 3.67e-06 & 6.60e-05 \\
GPT-5.4 & Gemini 3.1 Pro Preview & M4 & 40 & 3.732 (39) & 6.05e-04 & 0.0021 \\
GPT-5.4 & Gemini 3.1 Pro Preview & M5 & 40 & 4.523 (39) & 5.57e-05 & 5.01e-04 \\
GPT-5.4 & GPT-5.5 & M1 & 40 & 3.003 (39) & 0.0047 & 0.0093 \\
GPT-5.4 & GPT-5.5 & M2 & 40 & 4.891 (39) & 1.77e-05 & 2.47e-04 \\
GPT-5.4 & GPT-5.5 & M3 & 40 & 5.500 (39) & 2.56e-06 & 4.87e-05 \\
GPT-5.4 & GPT-5.5 & M4 & 40 & 4.891 (39) & 1.77e-05 & 2.47e-04 \\
GPT-5.4 & GPT-5.5 & M5 & 40 & 8.573 (39) & 1.66e-10 & 6.95e-09 \\
GPT-5.4 & Claude Opus 4.7 & M1 & 40 & 4.005 (39) & 2.70e-04 & 0.0016 \\
GPT-5.4 & Claude Opus 4.7 & M2 & 40 & 5.992 (39) & 5.31e-07 & 1.17e-05 \\
GPT-5.4 & Claude Opus 4.7 & M3 & 40 & 5.107 (39) & 8.93e-06 & 1.43e-04 \\
GPT-5.4 & Claude Opus 4.7 & M4 & 40 & 5.701 (39) & 1.35e-06 & 2.70e-05 \\
GPT-5.4 & Claude Opus 4.7 & M5 & 40 & 6.958 (39) & 2.45e-08 & 8.32e-07 \\
Claude Opus 4.7 & Gemini 3.1 Pro Preview & M1 & 40 & 5.921 (39) & 6.67e-07 & 1.40e-05 \\
Claude Opus 4.7 & Gemini 3.1 Pro Preview & M2 & 38 & 4.012 (37) & 2.81e-04 & 0.0016 \\
Claude Opus 4.7 & Gemini 3.1 Pro Preview & M3 & 40 & 6.982 (39) & 2.27e-08 & 7.94e-07 \\
Claude Opus 4.7 & Gemini 3.1 Pro Preview & M4 & 40 & 3.787 (39) & 5.14e-04 & 0.0021 \\
Claude Opus 4.7 & Gemini 3.1 Pro Preview & M5 & 40 & 5.339 (39) & 4.27e-06 & 7.26e-05 \\
Claude Opus 4.7 & GPT-5.5 & M1 & 40 & 4.611 (39) & 4.23e-05 & 4.23e-04 \\
Claude Opus 4.7 & GPT-5.5 & M2 & 40 & 6.150 (39) & 3.21e-07 & 8.02e-06 \\
Claude Opus 4.7 & GPT-5.5 & M3 & 40 & 6.736 (39) & 4.94e-08 & 1.52e-06 \\
Claude Opus 4.7 & GPT-5.5 & M4 & 40 & 6.183 (39) & 2.89e-07 & 7.50e-06 \\
Claude Opus 4.7 & GPT-5.5 & M5 & 40 & 6.150 (39) & 3.21e-07 & 8.02e-06 \\
Claude Opus 4.7 & Claude Opus 4.7 & M1 & 40 & 6.297 (39) & 2.00e-07 & 5.41e-06 \\
Claude Opus 4.7 & Claude Opus 4.7 & M2 & 40 & 6.607 (39) & 7.46e-08 & 2.16e-06 \\
Claude Opus 4.7 & Claude Opus 4.7 & M3 & 40 & 6.125 (39) & 3.47e-07 & 8.02e-06 \\
Claude Opus 4.7 & Claude Opus 4.7 & M4 & 40 & 6.750 (39) & 4.73e-08 & 1.51e-06 \\
Claude Opus 4.7 & Claude Opus 4.7 & M5 & 40 & 4.937 (39) & 1.53e-05 & 2.29e-04 \\
Gemini 3.1 Pro & Gemini 3.1 Pro Preview & M1 & 40 & 6.738 (39) & 4.92e-08 & 1.52e-06 \\
Gemini 3.1 Pro & Gemini 3.1 Pro Preview & M2 & 38 & 4.782 (37) & 2.76e-05 & 3.03e-04 \\
Gemini 3.1 Pro & Gemini 3.1 Pro Preview & M3 & 40 & 7.173 (39) & 1.24e-08 & 4.47e-07 \\
Gemini 3.1 Pro & Gemini 3.1 Pro Preview & M4 & 40 & 4.273 (39) & 1.20e-04 & 9.60e-04 \\
Gemini 3.1 Pro & Gemini 3.1 Pro Preview & M5 & 40 & 6.919 (39) & 2.77e-08 & 9.13e-07 \\
Gemini 3.1 Pro & GPT-5.5 & M1 & 40 & 4.891 (39) & 1.77e-05 & 2.47e-04 \\
Gemini 3.1 Pro & GPT-5.5 & M2 & 40 & 7.706 (39) & 2.35e-09 & 9.39e-08 \\
Gemini 3.1 Pro & GPT-5.5 & M3 & 40 & 6.565 (39) & 8.51e-08 & 2.38e-06 \\
Gemini 3.1 Pro & GPT-5.5 & M4 & 40 & 7.675 (39) & 2.58e-09 & 1.01e-07 \\
Gemini 3.1 Pro & GPT-5.5 & M5 & 40 & 10.462 (39) & 7.02e-13 & 3.09e-11 \\
Gemini 3.1 Pro & Claude Opus 4.7 & M1 & 40 & 7.286 (39) & 8.71e-09 & 3.22e-07 \\
Gemini 3.1 Pro & Claude Opus 4.7 & M2 & 40 & 11.000 (39) & 1.61e-13 & 7.24e-12 \\
Gemini 3.1 Pro & Claude Opus 4.7 & M3 & 40 & 7.746 (39) & 2.07e-09 & 8.50e-08 \\
Gemini 3.1 Pro & Claude Opus 4.7 & M4 & 40 & 7.524 (39) & 4.13e-09 & 1.57e-07 \\
Gemini 3.1 Pro & Claude Opus 4.7 & M5 & 40 & 10.077 (39) & 2.06e-12 & 8.86e-11 \\
\end{longtable}
\endgroup

All 45 domain contrasts form one complete Holm family. The complete differences and original unadjusted 95\% CIs are in panel c above. Gemini M2 uses 38 pairs (37 df); all other tests use 40 pairs (39 df). These supplementary tests assess zero mean differences, not differences between domains.

\subsection{Supplementary Table 9e. Paired 40-case comparison of format-aligned MCE output with physician responses}

\begingroup
\scriptsize
\setlength{\tabcolsep}{3pt}
\renewcommand{\arraystretch}{1.12}
\begin{longtable}{@{}>{\raggedright\arraybackslash}p{0.192\linewidth}>{\raggedright\arraybackslash}p{0.116\linewidth}>{\raggedright\arraybackslash}p{0.166\linewidth}>{\raggedright\arraybackslash}p{0.233\linewidth}>{\raggedright\arraybackslash}p{0.116\linewidth}>{\raggedright\arraybackslash}p{0.116\linewidth}@{}}
\toprule
\textbf{Outcome or domain} & \textbf{Format-aligned MCE mean} & \textbf{Physician mean within case and condition} & \textbf{Paired difference (format-aligned MCE minus physician case mean)} & \textbf{Case-cluster 95\% CI} & \textbf{Paired cases} \\
\midrule
\endfirsthead
\toprule
\textbf{Outcome or domain} & \textbf{Format-aligned MCE mean} & \textbf{Physician mean within case and condition} & \textbf{Paired difference (format-aligned MCE minus physician case mean)} & \textbf{Case-cluster 95\% CI} & \textbf{Paired cases} \\
\midrule
\endhead
\midrule
\multicolumn{6}{r}{\footnotesize Continued on next page} \\
\endfoot
\bottomrule
\endlastfoot
APAS versus U & 85.29 & 53.26 & 32.03 & 26.48--37.09 & 40 \\
APAS versus R & 85.29 & 61.00 & 24.29 & 19.94--28.60 & 40 \\
APAS versus M & 85.29 & 66.18 & 19.11 & 15.58--22.33 & 40 \\
Non-whitespace characters versus M & 363.6 & 611.1 & \ensuremath{-}247.5 & \ensuremath{-}313.4 to \ensuremath{-}185.2 & 40 \\
M1 clinical problem and treatment goals versus M & 84.58 & 63.95 & 20.63 & 16.48--24.51 & 40 \\
M2 decision-critical information versus M & 91.39 & 81.66 & 9.73 & 5.75--13.59 & 40 \\
M3 candidate clinical pathways versus M & 84.44 & 54.79 & 29.65 & 23.54--35.30 & 40 \\
M4 safety boundaries and risk governance versus M & 87.12 & 66.62 & 20.50 & 16.10--24.53 & 40 \\
M5 reassessment and governance versus M & 77.08 & 68.95 & 8.13 & 4.99--11.41 & 40 \\
\end{longtable}
\endgroup

The format-aligned MCE output was generated within the MCE system using the same five-part structure and a length constraint based on the physician task; it was not a post hoc summary of the complete report by a general-purpose AI model. APAS and M1--M5 were averaged equally across three ratings from each of three judges at the response level; physician responses were then averaged separately within each case and each of the U, R and M conditions. Differences were equally weighted across 40 cases. CIs used 30,000 case-cluster percentile bootstrap samples with seed 20260907. The paired differences compare each system output with the physician mean for the same case and support condition. These physician case means are not the mixed-model adjusted condition means in Figure 3a and Supplementary Table 9a. Figure 3c displays the overall paired APAS and length differences; this table provides the complete five-domain estimates and paired contrasts.

\textbf{Complete supplementary paired tests}

\begingroup
\scriptsize
\setlength{\tabcolsep}{3pt}
\renewcommand{\arraystretch}{1.12}
\begin{longtable}{@{}>{\raggedright\arraybackslash}p{0.134\linewidth}>{\raggedright\arraybackslash}p{0.134\linewidth}>{\raggedright\arraybackslash}p{0.136\linewidth}>{\raggedright\arraybackslash}p{0.134\linewidth}>{\raggedright\arraybackslash}p{0.134\linewidth}>{\raggedright\arraybackslash}p{0.134\linewidth}>{\raggedright\arraybackslash}p{0.134\linewidth}@{}}
\toprule
\textbf{Outcome} & \textbf{Physician condition} & \textbf{Difference (95\% CI)} & \textbf{Paired cases} & \textbf{t (df)} & \textbf{Raw P} & \textbf{Holm P} \\
\midrule
\endfirsthead
\toprule
\textbf{Outcome} & \textbf{Physician condition} & \textbf{Difference (95\% CI)} & \textbf{Paired cases} & \textbf{t (df)} & \textbf{Raw P} & \textbf{Holm P} \\
\midrule
\endhead
\midrule
\multicolumn{7}{r}{\footnotesize Continued on next page} \\
\endfoot
\bottomrule
\endlastfoot
APAS & U & 32.03 (26.48 to 37.09) & 40 & 11.799 (39) & 1.93e-14 & 4.06e-13 \\
APAS & R & 24.29 (19.94 to 28.60) & 40 & 10.938 (39) & 1.90e-13 & 3.23e-12 \\
APAS & M & 19.11 (15.58 to 22.33) & 40 & 10.987 (39) & 1.67e-13 & 3.00e-12 \\
Length & U & -197.5 (-258.2 to -136.3) & 40 & -6.282 (39) & 2.10e-07 & 1.47e-06 \\
Length & R & -239.9 (-313.1 to -169.0) & 40 & -6.422 (39) & 1.34e-07 & 1.07e-06 \\
Length & M & -247.5 (-313.4 to -185.2) & 40 & -7.433 (39) & 5.48e-09 & 4.94e-08 \\
M1 & U & 33.83 (27.60 to 39.52) & 40 & 11.050 (39) & 1.41e-13 & 2.67e-12 \\
M1 & R & 24.89 (19.82 to 29.86) & 40 & 9.583 (39) & 8.43e-12 & 1.01e-10 \\
M1 & M & 20.63 (16.48 to 24.51) & 40 & 9.971 (39) & 2.78e-12 & 4.17e-11 \\
M2 & U & 18.19 (11.94 to 24.32) & 40 & 5.748 (39) & 1.16e-06 & 4.63e-06 \\
M2 & R & 14.57 (9.06 to 20.52) & 40 & 4.935 (39) & 1.54e-05 & 4.62e-05 \\
M2 & M & 9.73 (5.75 to 13.59) & 40 & 4.866 (39) & 1.91e-05 & 4.62e-05 \\
M3 & U & 50.41 (40.25 to 59.56) & 40 & 10.143 (39) & 1.71e-12 & 2.73e-11 \\
M3 & R & 38.29 (30.66 to 45.62) & 40 & 9.912 (39) & 3.28e-12 & 4.60e-11 \\
M3 & M & 29.65 (23.54 to 35.30) & 40 & 9.744 (39) & 5.31e-12 & 6.90e-11 \\
M4 & U & 33.01 (27.33 to 38.59) & 40 & 11.427 (39) & 5.14e-14 & 1.03e-12 \\
M4 & R & 25.18 (20.03 to 30.29) & 40 & 9.534 (39) & 9.71e-12 & 1.07e-10 \\
M4 & M & 20.50 (16.10 to 24.53) & 40 & 9.442 (39) & 1.27e-11 & 1.27e-10 \\
M5 & U & 12.26 (8.39 to 16.23) & 40 & 6.052 (39) & 4.38e-07 & 2.63e-06 \\
M5 & R & 11.45 (7.90 to 15.18) & 40 & 6.018 (39) & 4.89e-07 & 2.63e-06 \\
M5 & M & 8.13 (4.99 to 11.41) & 40 & 4.900 (39) & 1.72e-05 & 4.62e-05 \\
\end{longtable}
\endgroup

All 21 contrasts form one complete Holm family: three overall APAS, 15 domain and three non-whitespace-character contrasts. The supplementary post hoc two-sided paired t tests use the 40 original case differences (39 df). The original 30,000-draw case-cluster percentile bootstrap intervals are retained unchanged and are unadjusted; they were not inverted to calculate P values.

\end{landscape}
\subsection{Supplementary Table 9f. Data-quality sensitivity analysis of subsequent clinical events after exclusion of three records with prespecified data-integrity concerns}

\begingroup
\footnotesize
\setlength{\tabcolsep}{3pt}
\renewcommand{\arraystretch}{1.12}
\begin{longtable}{@{}>{\raggedright\arraybackslash}p{0.152\linewidth}>{\raggedright\arraybackslash}p{0.209\linewidth}>{\raggedright\arraybackslash}p{0.258\linewidth}>{\raggedright\arraybackslash}p{0.321\linewidth}@{}}
\toprule
\textbf{Estimand} & \textbf{Component clinical-event record (n=12)} & \textbf{Follow-up without a recorded component event (n=19)} & \textbf{Between-group difference, event-record minus no-record group (95\% CI)} \\
\midrule
\endfirsthead
\toprule
\textbf{Estimand} & \textbf{Component clinical-event record (n=12)} & \textbf{Follow-up without a recorded component event (n=19)} & \textbf{Between-group difference, event-record minus no-record group (95\% CI)} \\
\midrule
\endhead
\midrule
\multicolumn{4}{r}{\footnotesize Continued on next page} \\
\endfoot
\bottomrule
\endlastfoot
APAS under U & 55.27 (52.51--58.03) & 49.57 (47.09--52.05) & 5.70 (2.84--8.57) \\
APAS under R & 64.44 (61.52--67.35) & 56.26 (53.80--58.73) & 8.17 (5.18--11.17) \\
APAS under M & 68.47 (65.62--71.31) & 64.37 (61.93--66.81) & 4.10 (1.18--7.01) \\
M--U & 13.20 (10.03--16.37) & 14.80 (12.29--17.31) & \ensuremath{-}1.61 (\ensuremath{-}5.74 to 2.53) \\
M--R & 4.03 (0.74--7.33) & 8.11 (5.62--10.60) & \ensuremath{-}4.08 (\ensuremath{-}8.28 to 0.12) \\
\end{longtable}
\endgroup

Three records were removed from the standardization target because of a chronology inconsistency, an unparseable index date and uncertain record linkage. Estimates remained derived from \texttt{APAS \textasciitilde{} condition \ensuremath{\times} case + task position + (1 | physician)} fitted to all 40 cases and were equally standardized over the 31 cases shown; the model was not refitted after exclusion. Six cases with unavailable follow-up remained separate. Groups were derived from post hoc text rules without independent clinical adjudication. Absence of a recorded component event did not imply absence of progression or favorable prognosis. This analysis was post hoc and unadjusted; inclusion or exclusion of zero was not used for confirmatory effect-modification claims.

\subsection{Supplementary Table 9g. M1--M5 APAS differences by subsequent-event record among all cases with available follow-up}

\begingroup
\scriptsize
\setlength{\tabcolsep}{3pt}
\renewcommand{\arraystretch}{1.12}
\begin{longtable}{@{}>{\raggedright\arraybackslash}p{0.201\linewidth}>{\raggedright\arraybackslash}p{0.190\linewidth}>{\raggedright\arraybackslash}p{0.179\linewidth}>{\raggedright\arraybackslash}p{0.190\linewidth}>{\raggedright\arraybackslash}p{0.179\linewidth}@{}}
\toprule
\textbf{Domain} & \textbf{Event-record group M--U (n=14; 95\% CI)} & \textbf{No-record group M--U (n=20; 95\% CI)} & \textbf{Event-record group M--R (n=14; 95\% CI)} & \textbf{No-record group M--R (n=20; 95\% CI)} \\
\midrule
\endfirsthead
\toprule
\textbf{Domain} & \textbf{Event-record group M--U (n=14; 95\% CI)} & \textbf{No-record group M--U (n=20; 95\% CI)} & \textbf{Event-record group M--R (n=14; 95\% CI)} & \textbf{No-record group M--R (n=20; 95\% CI)} \\
\midrule
\endhead
\midrule
\multicolumn{5}{r}{\footnotesize Continued on next page} \\
\endfoot
\bottomrule
\endlastfoot
M1 clinical problem and treatment goals & 8.89 (5.36--12.43) & 15.86 (12.90--18.82) & 2.08 (\ensuremath{-}1.56 to 5.71) & 6.93 (4.01--9.85) \\
M2 decision-critical information & 8.65 (4.76--12.54) & 9.65 (6.39--12.91) & 2.32 (\ensuremath{-}1.68 to 6.32) & 8.03 (4.80--11.25) \\
M3 candidate clinical pathways & 18.91 (14.56--23.26) & 20.09 (16.45--23.74) & 5.66 (1.18--10.13) & 11.47 (7.87--15.06) \\
M4 safety boundaries and risk governance & 10.69 (7.16--14.22) & 13.11 (10.16--16.07) & 4.62 (1.00--8.25) & 5.38 (2.46--8.29) \\
M5 reassessment and governance & 3.25 (0.38--6.12) & 5.44 (3.04--7.85) & 3.59 (0.64--6.55) & 4.63 (2.26--7.00) \\
\end{longtable}
\endgroup

Domain scores were calculated only when the domain was applicable and entered the denominator. This table includes all 34 cases with available follow-up, including the three records with chronology, index-date or record-linkage concerns, and is supplementary and exploratory. Twenty within-group contrasts and ten between-group interaction contrasts were not adjusted for multiplicity. Inclusion or exclusion of zero was not used to confirm domain-specific effect modification.

\begin{landscape}
\subsection{Supplementary Table 9h. Data-quality sensitivity and hypothesis-generating strata for subsequent clinical events}

\subsubsection{a. Data-quality and recorded-death strata}

\begingroup
\scriptsize
\setlength{\tabcolsep}{3pt}
\renewcommand{\arraystretch}{1.12}
\begin{longtable}{@{}>{\raggedright\arraybackslash}p{0.223\linewidth}>{\raggedright\arraybackslash}p{0.225\linewidth}>{\raggedright\arraybackslash}p{0.101\linewidth}>{\raggedright\arraybackslash}p{0.101\linewidth}>{\raggedright\arraybackslash}p{0.101\linewidth}>{\raggedright\arraybackslash}p{0.189\linewidth}@{}}
\toprule
\textbf{Analysis} & \textbf{Groups and case counts} & \textbf{APAS under M} & \textbf{M--U} & \textbf{M--R} & \textbf{Between-group difference in contrast (95\% CI)} \\
\midrule
\endfirsthead
\toprule
\textbf{Analysis} & \textbf{Groups and case counts} & \textbf{APAS under M} & \textbf{M--U} & \textbf{M--R} & \textbf{Between-group difference in contrast (95\% CI)} \\
\midrule
\endhead
\midrule
\multicolumn{6}{r}{\footnotesize Continued on next page} \\
\endfoot
\bottomrule
\endlastfoot
All records with available follow-up & Component event record, 14; follow-up without recorded component event, 20 & 66.71 vs 64.46 & 10.69 vs 14.20 & 3.60 vs 7.45 & M--U: \ensuremath{-}3.50 (\ensuremath{-}7.44 to 0.43); M--R: \ensuremath{-}3.85 (\ensuremath{-}7.81 to 0.10) \\
Recorded-death stratum & Death recorded, 5; follow-up without recorded death, 29 & 69.01 vs 64.76 & 5.94 vs 13.93 & 0.89 vs 6.72 & M--U: \ensuremath{-}7.99 (\ensuremath{-}13.57 to \ensuremath{-}2.41); M--R: \ensuremath{-}5.83 (\ensuremath{-}11.42 to \ensuremath{-}0.24) \\
Recorded-death stratum excluding the record with a reversed date sequence & Death recorded, 4; follow-up without recorded death, 29 & 67.90 vs 64.76 & 7.42 vs 13.93 & 1.10 vs 6.72 & M--U: \ensuremath{-}6.51 (\ensuremath{-}12.65 to \ensuremath{-}0.36); M--R: \ensuremath{-}5.62 (\ensuremath{-}11.82 to 0.58) \\
\end{longtable}
\endgroup

All rows were derived from the same model fitted to all 40 cases; only the equally weighted standardization target changed. Follow-up lacked a uniform index date, observation window, surveillance intensity and administrative censoring, so between-group differences may reflect opportunity for event ascertainment. The recorded-death strata contained only four to five cases. Wald intervals are conditional on the fixed case set and fitted model and do not include uncertainty from post hoc selection of the target set. These analyses are not survival analyses. The three prespecified data-integrity concerns were a chronology inconsistency, an unparseable index date and uncertain record linkage.

\end{landscape}
\subsubsection{b. Goal structure and primary clinical-decision scenario}

\begingroup
\footnotesize
\setlength{\tabcolsep}{3pt}
\renewcommand{\arraystretch}{1.12}
\begin{longtable}{@{}>{\raggedright\arraybackslash}p{0.373\linewidth}>{\raggedright\arraybackslash}p{0.189\linewidth}>{\raggedright\arraybackslash}p{0.189\linewidth}>{\raggedright\arraybackslash}p{0.189\linewidth}@{}}
\toprule
\textbf{Post hoc stratum} & \textbf{Cases} & \textbf{M--U (95\% CI)} & \textbf{M--R (95\% CI)} \\
\midrule
\endfirsthead
\toprule
\textbf{Post hoc stratum} & \textbf{Cases} & \textbf{M--U (95\% CI)} & \textbf{M--R (95\% CI)} \\
\midrule
\endhead
\midrule
\multicolumn{4}{r}{\footnotesize Continued on next page} \\
\endfoot
\bottomrule
\endlastfoot
Dual goals & 12 & 7.45 (4.34--10.57) & 1.90 (\ensuremath{-}1.26 to 5.07) \\
Primary goal with fallback goal & 27 & 15.05 (12.93--17.16) & 6.60 (4.46--8.73) \\
Diagnosis, staging, tumor origin or biological boundary & 11 & 14.61 (11.37--17.85) & 1.59 (\ensuremath{-}1.71 to 4.88) \\
Local-treatment resectability and local control & 7 & 10.57 (6.33--14.80) & 5.68 (1.47--9.89) \\
Systemic-treatment selection, resistance and sequence & 5 & 8.53 (3.71--13.34) & 7.31 (2.10--12.52) \\
Treatment toxicity, comorbidity and tolerance constraints & 16 & 13.68 (10.92--16.44) & 6.80 (4.00--9.60) \\
Prioritization of coexisting disease or multiple primary tumors & 1 & 18.46 (4.99--31.93) & 6.23 (\ensuremath{-}6.19 to 18.66) \\
\end{longtable}
\endgroup

One case had a single goal and was not included in the dual-goal versus primary-plus-fallback comparison. Primary decision scenarios use the mutually exclusive exploratory classification in Supplementary Table 10c. These strata identify candidates for prespecified future analyses. No multiplicity adjustment was applied, and small strata were not ranked or interpreted deterministically.

\subsubsection{c. Domain point estimates in cases with non-death records of treatment interruption, non-initiation or health-status deterioration}

\begingroup
\scriptsize
\setlength{\tabcolsep}{3pt}
\renewcommand{\arraystretch}{1.12}
\begin{longtable}{@{}>{\raggedright\arraybackslash}p{0.247\linewidth}>{\raggedright\arraybackslash}p{0.173\linewidth}>{\raggedright\arraybackslash}p{0.173\linewidth}>{\raggedright\arraybackslash}p{0.173\linewidth}>{\raggedright\arraybackslash}p{0.173\linewidth}@{}}
\toprule
\textbf{Subgroup} & \textbf{Cases} & \textbf{M2 M--U} & \textbf{M3 M--U} & \textbf{M4 M--U} \\
\midrule
\endfirsthead
\toprule
\textbf{Subgroup} & \textbf{Cases} & \textbf{M2 M--U} & \textbf{M3 M--U} & \textbf{M4 M--U} \\
\midrule
\endhead
\midrule
\multicolumn{5}{r}{\footnotesize Continued on next page} \\
\endfoot
\bottomrule
\endlastfoot
Relevant record without a recorded death & 7 & 17.01 & 28.04 & 16.96 \\
\end{longtable}
\endgroup

This subgroup was identified from non-structured follow-up records referring to toxicity or intolerance, treatment cessation, functional or organ deterioration, or non-initiation of lung-cancer treatment. Values are model-derived directional point estimates without a confirmatory between-group test. No case-level mapping established whether these later records corresponded to decision-critical information, alternative pathways or safety constraints that were reasonably foreseeable at baseline.

\subsection{Supplementary Table 9i. Complete estimates for overall clinical anchoring and EASS rule re-rating}

\subsubsection{a. Overall clinical anchoring}

\begingroup
\footnotesize
\setlength{\tabcolsep}{3pt}
\renewcommand{\arraystretch}{1.12}
\begin{longtable}{@{}>{\raggedright\arraybackslash}p{0.253\linewidth}>{\raggedright\arraybackslash}p{0.427\linewidth}>{\raggedright\arraybackslash}p{0.130\linewidth}>{\raggedright\arraybackslash}p{0.130\linewidth}@{}}
\toprule
\textbf{Estimand} & \textbf{Unit} & \textbf{Result} & \textbf{95\% CI} \\
\midrule
\endfirsthead
\toprule
\textbf{Estimand} & \textbf{Unit} & \textbf{Result} & \textbf{95\% CI} \\
\midrule
\endhead
\midrule
\multicolumn{4}{r}{\footnotesize Continued on next page} \\
\endfoot
\bottomrule
\endlastfoot
Reviewer 1 four-category acceptability distribution & 120 outputs; unacceptable /\allowbreak{} major revision or verification /\allowbreak{} acceptable with minor revision /\allowbreak{} acceptable for clinical discussion & 12/\allowbreak{}34/\allowbreak{}19/\allowbreak{}55 & --- \\
Reviewer 2 four-category acceptability distribution & Same & 8/\allowbreak{}31/\allowbreak{}23/\allowbreak{}58 & --- \\
Reviewer 1 binary acceptable & 120 outputs & 74/\allowbreak{}120 (61.7\%) & --- \\
Reviewer 2 binary acceptable & Same & 81/\allowbreak{}120 (67.5\%) & --- \\
Acceptable to both reviewers & Same & 71/\allowbreak{}120 (59.2\%) & --- \\
Reviewer 1 major clinical defect & Same & 38/\allowbreak{}120 (31.7\%) & --- \\
Reviewer 2 major clinical defect & Same & 32/\allowbreak{}120 (26.7\%) & --- \\
Exact agreement, five-category acceptability & 120 outputs; 40 case clusters & 89/\allowbreak{}120 (74.2\%) & 66.7\%--80.8\% \\
Cohen \ensuremath{\kappa}, five-category acceptability & Same & 0.613 & 0.509--0.717 \\
Adjacent agreement, four-category ordinal outcome & Same & 117/\allowbreak{}120 (97.5\%) & 94.2\%--100.0\% \\
Linearly weighted Cohen \ensuremath{\kappa}, four-category ordinal outcome & Same & 0.745 & 0.671--0.815 \\
Exact agreement, binary acceptability & Same & 107/\allowbreak{}120 (89.2\%) & 84.2\%--94.2\% \\
Cohen \ensuremath{\kappa}, binary acceptability & Same & 0.764 & 0.665--0.862 \\
Exact agreement, major clinical defect & Same & 108/\allowbreak{}120 (90.0\%) & 84.2\%--95.0\% \\
Cohen \ensuremath{\kappa}, major clinical defect & Same & 0.759 & 0.624--0.877 \\
Spearman's \ensuremath{\rho} between APAS and mean four-category acceptability & 120 outputs; 40 case clusters & 0.671 & 0.581--0.752 \\
\end{longtable}
\endgroup

Clinical acceptability concerned suitability for clinical discussion and the extent of revision or additional verification required. The original five-category acceptability item comprised four ordered categories and a special not-assessable state; neither reviewer used the special state. Four-category ordinal and binary outcomes were prespecified sensitivity analyses. Intervals used case-cluster resampling. Source-specific analyses described the fixed enriched sample and were not confirmatory comparisons between sources.

\subsubsection{b. EASS rule re-rating}

\begingroup
\footnotesize
\setlength{\tabcolsep}{3pt}
\renewcommand{\arraystretch}{1.12}
\begin{longtable}{@{}>{\raggedright\arraybackslash}p{0.326\linewidth}>{\raggedright\arraybackslash}p{0.217\linewidth}>{\raggedright\arraybackslash}p{0.171\linewidth}>{\raggedright\arraybackslash}p{0.226\linewidth}@{}}
\toprule
\textbf{Estimand} & \textbf{Unit} & \textbf{Result} & \textbf{95\% CI or limits} \\
\midrule
\endfirsthead
\toprule
\textbf{Estimand} & \textbf{Unit} & \textbf{Result} & \textbf{95\% CI or limits} \\
\midrule
\endhead
\midrule
\multicolumn{4}{r}{\footnotesize Continued on next page} \\
\endfoot
\bottomrule
\endlastfoot
Rule-level exact agreement & 240 rules; 40 complete MCE reports & 188/\allowbreak{}240 (78.33\%) & --- \\
Rule-level adjacent agreement & Same & 238/\allowbreak{}240 (99.17\%) & --- \\
Linearly weighted Gwet AC2 & Same & 0.842 & 0.785--0.893 \\
Linearly weighted Cohen \ensuremath{\kappa} & Same & 0.411 & 0.257--0.558 \\
Case-level ICC(A,1) & 40 paired reports & 0.694 & 0.468--0.835 \\
Case-level Lin CCC & Same & 0.689 & --- \\
Mean case-level APAS difference, reviewer 1 minus reviewer 2 & Same & \ensuremath{-}1.10 & Limits of agreement, \ensuremath{-}15.74 to 13.53 \\
Mean absolute case-level APAS difference & Same & 5.22 & --- \\
\end{longtable}
\endgroup

Rule-level results compare two physicians using the fixed case-specific scoring rules; not-assessable states were not recoded. Case-level results aggregate rule scores within report. AC2, ICC and their intervals used case-cluster resampling. Limits of agreement describe the observed range of score differences and are not clinical-equivalence thresholds.

\begin{landscape}
\subsection{Supplementary Table 9j. Complete controlled-text perturbation estimates and target-domain mapping}

\subsubsection{a. Overall APAS change and direction counts}

\begingroup
\scriptsize
\setlength{\tabcolsep}{3pt}
\renewcommand{\arraystretch}{1.12}
\begin{longtable}{@{}>{\raggedright\arraybackslash}p{0.193\linewidth}>{\raggedright\arraybackslash}p{0.129\linewidth}>{\raggedright\arraybackslash}p{0.233\linewidth}>{\raggedright\arraybackslash}p{0.129\linewidth}>{\raggedright\arraybackslash}p{0.129\linewidth}>{\raggedright\arraybackslash}p{0.129\linewidth}@{}}
\toprule
\textbf{Variant type} & \textbf{Triplets} & \textbf{Overall APAS difference, variant minus original (95\% CI)} & \textbf{Decrease} & \textbf{No change} & \textbf{Increase} \\
\midrule
\endfirsthead
\toprule
\textbf{Variant type} & \textbf{Triplets} & \textbf{Overall APAS difference, variant minus original (95\% CI)} & \textbf{Decrease} & \textbf{No change} & \textbf{Increase} \\
\midrule
\endhead
\midrule
\multicolumn{6}{r}{\footnotesize Continued on next page} \\
\endfoot
\bottomrule
\endlastfoot
Prespecified section-order variant & 36 & +0.85 (\ensuremath{-}0.44 to 2.31) & 15/\allowbreak{}36 & 2/\allowbreak{}36 & 19/\allowbreak{}36 \\
Rule-linked target-content-deletion variant & 36 & \ensuremath{-}0.66 (\ensuremath{-}1.69 to 0.36) & 20/\allowbreak{}36 & 1/\allowbreak{}36 & 15/\allowbreak{}36 \\
\end{longtable}
\endgroup

\end{landscape}
\subsubsection{b. Within-domain APAS change for the deleted target content}

\begingroup
\small
\setlength{\tabcolsep}{3pt}
\renewcommand{\arraystretch}{1.12}
\begin{longtable}{@{}>{\raggedright\arraybackslash}p{0.306\linewidth}>{\raggedright\arraybackslash}p{0.215\linewidth}>{\raggedright\arraybackslash}p{0.419\linewidth}@{}}
\toprule
\textbf{Target domain} & \textbf{Triplets} & \textbf{Target-domain APAS difference, variant minus original (95\% CI)} \\
\midrule
\endfirsthead
\toprule
\textbf{Target domain} & \textbf{Triplets} & \textbf{Target-domain APAS difference, variant minus original (95\% CI)} \\
\midrule
\endhead
\midrule
\multicolumn{3}{r}{\footnotesize Continued on next page} \\
\endfoot
\bottomrule
\endlastfoot
M2 decision-critical information & 9 & \ensuremath{-}0.32 (\ensuremath{-}3.41 to 2.13) \\
M3 candidate clinical pathways & 9 & \ensuremath{-}3.70 (\ensuremath{-}8.02 to 0.00) \\
M4 safety boundaries and risk governance & 9 & \ensuremath{-}0.46 (\ensuremath{-}3.86 to 4.32) \\
M5 reassessment and governance & 9 & \ensuremath{-}1.85 (\ensuremath{-}4.94 to 1.23) \\
\end{longtable}
\endgroup

\subsubsection{c. Variant-target mapping}

\begingroup
\footnotesize
\setlength{\tabcolsep}{3pt}
\renewcommand{\arraystretch}{1.12}
\begin{longtable}{@{}>{\raggedright\arraybackslash}p{0.228\linewidth}>{\raggedright\arraybackslash}p{0.293\linewidth}>{\raggedright\arraybackslash}p{0.152\linewidth}>{\raggedright\arraybackslash}p{0.268\linewidth}@{}}
\toprule
\textbf{Variant type} & \textbf{Intended transformation} & \textbf{Triplets} & \textbf{Main comparison} \\
\midrule
\endfirsthead
\toprule
\textbf{Variant type} & \textbf{Intended transformation} & \textbf{Triplets} & \textbf{Main comparison} \\
\midrule
\endhead
\midrule
\multicolumn{4}{r}{\footnotesize Continued on next page} \\
\endfoot
\bottomrule
\endlastfoot
Prespecified section-order variant & Change section order while retaining strategy content & 36 & Overall APAS, variant minus original \\
Rule-linked target-content-deletion variant & Delete designated clinical content linked to M2, M3, M4 or M5 & 36 (9 per domain) & Overall and target-domain APAS, variant minus original \\
\end{longtable}
\endgroup

The triplet was the paired unit and 12 cases were clustering units. Three ratings from each of three judges per text were measurement repeats. CIs used 30,000 case-cluster percentile bootstrap samples. Physicians verified attainment of the intended transformation and recorded off-target changes before scoring; this construction check did not replace outcome scoring or establish a human reference standard.

\subsection{Supplementary Table 9k. Post hoc clinical-content estimates by case category and hospital background}

All scores and contrasts use the 0-100 domain scale. General denotes general hospitals; Specialty denotes cancer or thoracic/pulmonary-specialty hospitals. Category abbreviations refer to physician-reviewed complexity axes, not absence of the other disease. Confidence intervals are unadjusted. The primary 250-physician APAS analysis is separate.

\subsubsection{a. Complete adjusted means}

\textbf{Comorbidity only}

\begingroup
\scriptsize
\setlength{\tabcolsep}{3pt}
\renewcommand{\arraystretch}{1.12}
\begin{longtable}{@{}>{\raggedright\arraybackslash}p{0.188\linewidth}>{\raggedright\arraybackslash}p{0.188\linewidth}>{\raggedright\arraybackslash}p{0.188\linewidth}>{\raggedright\arraybackslash}p{0.188\linewidth}>{\raggedright\arraybackslash}p{0.188\linewidth}@{}}
\toprule
\textbf{Domain} & \textbf{Hospital} & \textbf{U (95\% CI)} & \textbf{R (95\% CI)} & \textbf{M (95\% CI)} \\
\midrule
\endfirsthead
\toprule
\textbf{Domain} & \textbf{Hospital} & \textbf{U (95\% CI)} & \textbf{R (95\% CI)} & \textbf{M (95\% CI)} \\
\midrule
\endhead
\midrule
\multicolumn{5}{r}{\footnotesize Continued on next page} \\
\endfoot
\bottomrule
\endlastfoot
M1 & General & 49.11 (44.82 to 53.41) & 59.15 (54.80 to 63.50) & 59.37 (55.89 to 62.84) \\
M1 & Specialty & 52.21 (45.97 to 58.46) & 53.53 (45.87 to 61.18) & 57.09 (52.01 to 62.18) \\
M2 & General & 62.86 (57.33 to 68.39) & 67.04 (62.69 to 71.39) & 77.69 (73.35 to 82.02) \\
M2 & Specialty & 62.23 (52.52 to 71.93) & 71.01 (65.16 to 76.85) & 67.77 (62.18 to 73.37) \\
M3 & General & 39.89 (35.08 to 44.71) & 44.92 (40.17 to 49.67) & 55.04 (49.82 to 60.26) \\
M3 & Specialty & 33.77 (22.64 to 44.89) & 43.19 (37.12 to 49.25) & 48.12 (40.21 to 56.03) \\
M4 & General & 56.17 (50.81 to 61.52) & 61.43 (56.79 to 66.07) & 67.49 (63.00 to 71.97) \\
M4 & Specialty & 53.64 (45.64 to 61.65) & 61.42 (55.91 to 66.93) & 65.05 (57.30 to 72.80) \\
M5 & General & 65.72 (61.76 to 69.68) & 62.84 (58.48 to 67.21) & 70.42 (66.57 to 74.26) \\
M5 & Specialty & 63.75 (57.47 to 70.02) & 60.28 (52.05 to 68.51) & 65.97 (60.53 to 71.41) \\
\end{longtable}
\endgroup

\textbf{Lung cancer only}

\begingroup
\scriptsize
\setlength{\tabcolsep}{3pt}
\renewcommand{\arraystretch}{1.12}
\begin{longtable}{@{}>{\raggedright\arraybackslash}p{0.188\linewidth}>{\raggedright\arraybackslash}p{0.188\linewidth}>{\raggedright\arraybackslash}p{0.188\linewidth}>{\raggedright\arraybackslash}p{0.188\linewidth}>{\raggedright\arraybackslash}p{0.188\linewidth}@{}}
\toprule
\textbf{Domain} & \textbf{Hospital} & \textbf{U (95\% CI)} & \textbf{R (95\% CI)} & \textbf{M (95\% CI)} \\
\midrule
\endfirsthead
\toprule
\textbf{Domain} & \textbf{Hospital} & \textbf{U (95\% CI)} & \textbf{R (95\% CI)} & \textbf{M (95\% CI)} \\
\midrule
\endhead
\midrule
\multicolumn{5}{r}{\footnotesize Continued on next page} \\
\endfoot
\bottomrule
\endlastfoot
M1 & General & 47.06 (43.37 to 50.75) & 57.07 (53.37 to 60.77) & 64.54 (59.88 to 69.21) \\
M1 & Specialty & 42.43 (36.43 to 48.43) & 55.60 (46.86 to 64.34) & 60.51 (52.55 to 68.47) \\
M2 & General & 73.89 (68.36 to 79.43) & 82.30 (78.48 to 86.13) & 89.23 (85.83 to 92.64) \\
M2 & Specialty & 79.65 (71.59 to 87.71) & 84.98 (76.30 to 93.67) & 87.09 (80.36 to 93.83) \\
M3 & General & 30.21 (25.84 to 34.59) & 46.95 (41.92 to 51.98) & 55.99 (50.76 to 61.21) \\
M3 & Specialty & 24.72 (17.77 to 31.67) & 43.43 (35.69 to 51.18) & 43.87 (31.15 to 56.60) \\
M4 & General & 50.59 (46.33 to 54.85) & 62.62 (58.67 to 66.58) & 68.20 (64.38 to 72.01) \\
M4 & Specialty & 52.35 (45.23 to 59.47) & 58.68 (51.12 to 66.24) & 62.53 (53.54 to 71.52) \\
M5 & General & 62.70 (58.66 to 66.74) & 64.44 (61.02 to 67.87) & 69.34 (66.46 to 72.23) \\
M5 & Specialty & 63.27 (57.16 to 69.39) & 64.69 (59.15 to 70.24) & 69.23 (62.91 to 75.56) \\
\end{longtable}
\endgroup

\textbf{Both axes}

\begingroup
\scriptsize
\setlength{\tabcolsep}{3pt}
\renewcommand{\arraystretch}{1.12}
\begin{longtable}{@{}>{\raggedright\arraybackslash}p{0.188\linewidth}>{\raggedright\arraybackslash}p{0.188\linewidth}>{\raggedright\arraybackslash}p{0.188\linewidth}>{\raggedright\arraybackslash}p{0.188\linewidth}>{\raggedright\arraybackslash}p{0.188\linewidth}@{}}
\toprule
\textbf{Domain} & \textbf{Hospital} & \textbf{U (95\% CI)} & \textbf{R (95\% CI)} & \textbf{M (95\% CI)} \\
\midrule
\endfirsthead
\toprule
\textbf{Domain} & \textbf{Hospital} & \textbf{U (95\% CI)} & \textbf{R (95\% CI)} & \textbf{M (95\% CI)} \\
\midrule
\endhead
\midrule
\multicolumn{5}{r}{\footnotesize Continued on next page} \\
\endfoot
\bottomrule
\endlastfoot
M1 & General & 53.52 (49.38 to 57.66) & 62.68 (58.29 to 67.07) & 67.12 (63.17 to 71.08) \\
M1 & Specialty & 55.08 (50.11 to 60.05) & 62.51 (57.34 to 67.68) & 63.80 (58.63 to 68.96) \\
M2 & General & 75.50 (71.29 to 79.71) & 76.18 (72.48 to 79.89) & 79.79 (76.06 to 83.52) \\
M2 & Specialty & 75.86 (71.11 to 80.62) & 77.03 (70.42 to 83.63) & 79.62 (74.08 to 85.16) \\
M3 & General & 33.44 (29.27 to 37.62) & 48.57 (43.62 to 53.52) & 57.36 (52.33 to 62.39) \\
M3 & Specialty & 40.26 (34.51 to 46.01) & 44.52 (37.66 to 51.39) & 53.59 (46.60 to 60.59) \\
M4 & General & 54.15 (49.31 to 58.98) & 64.01 (59.87 to 68.16) & 67.37 (62.62 to 72.11) \\
M4 & Specialty & 58.25 (52.82 to 63.68) & 60.33 (55.92 to 64.73) & 64.66 (59.15 to 70.17) \\
M5 & General & 65.48 (62.29 to 68.67) & 69.03 (66.54 to 71.53) & 67.92 (64.39 to 71.44) \\
M5 & Specialty & 67.95 (64.76 to 71.14) & 66.86 (61.98 to 71.73) & 69.40 (65.47 to 73.32) \\
\end{longtable}
\endgroup

\subsubsection{b. Complete hospital-specific support contrasts}

The 60 M-U/M-R comparisons below form one complete Holm family; a positive unadjusted interval alone is not a multiplicity-adjusted declaration.

\begin{landscape}
\textbf{Comorbidity only}

\begingroup
\scriptsize
\setlength{\tabcolsep}{3pt}
\renewcommand{\arraystretch}{1.12}
\begin{longtable}{@{}>{\raggedright\arraybackslash}p{0.157\linewidth}>{\raggedright\arraybackslash}p{0.157\linewidth}>{\raggedright\arraybackslash}p{0.157\linewidth}>{\raggedright\arraybackslash}p{0.157\linewidth}>{\raggedright\arraybackslash}p{0.157\linewidth}>{\raggedright\arraybackslash}p{0.157\linewidth}@{}}
\toprule
\textbf{Domain} & \textbf{Hospital} & \textbf{M-U (95\% CI)} & \textbf{Holm P} & \textbf{M-R (95\% CI)} & \textbf{Holm P} \\
\midrule
\endfirsthead
\toprule
\textbf{Domain} & \textbf{Hospital} & \textbf{M-U (95\% CI)} & \textbf{Holm P} & \textbf{M-R (95\% CI)} & \textbf{Holm P} \\
\midrule
\endhead
\midrule
\multicolumn{6}{r}{\footnotesize Continued on next page} \\
\endfoot
\bottomrule
\endlastfoot
M1 & General & 10.26 (4.94 to 15.57) & 0.020 & 0.22 (-3.99 to 4.43) & 1.000 \\
M1 & Specialty & 4.88 (-1.46 to 11.22) & 1.000 & 3.56 (-4.66 to 11.79) & 1.000 \\
M2 & General & 14.82 (8.83 to 20.82) & <0.001 & 10.65 (5.88 to 15.42) & 0.004 \\
M2 & Specialty & 5.54 (-6.12 to 17.21) & 1.000 & -3.23 (-10.39 to 3.92) & 1.000 \\
M3 & General & 15.14 (8.18 to 22.11) & 0.005 & 10.12 (4.42 to 15.81) & 0.045 \\
M3 & Specialty & 14.35 (2.84 to 25.86) & 0.599 & 4.93 (-3.05 to 12.91) & 1.000 \\
M4 & General & 11.32 (5.54 to 17.09) & 0.017 & 6.06 (1.15 to 10.97) & 0.585 \\
M4 & Specialty & 11.41 (2.43 to 20.39) & 0.573 & 3.63 (-2.91 to 10.17) & 1.000 \\
M5 & General & 4.70 (-0.55 to 9.94) & 1.000 & 7.57 (2.72 to 12.43) & 0.130 \\
M5 & Specialty & 2.22 (-5.22 to 9.67) & 1.000 & 5.69 (-2.21 to 13.60) & 1.000 \\
\end{longtable}
\endgroup

\textbf{Lung cancer only}

\begingroup
\scriptsize
\setlength{\tabcolsep}{3pt}
\renewcommand{\arraystretch}{1.12}
\begin{longtable}{@{}>{\raggedright\arraybackslash}p{0.157\linewidth}>{\raggedright\arraybackslash}p{0.157\linewidth}>{\raggedright\arraybackslash}p{0.157\linewidth}>{\raggedright\arraybackslash}p{0.157\linewidth}>{\raggedright\arraybackslash}p{0.157\linewidth}>{\raggedright\arraybackslash}p{0.157\linewidth}@{}}
\toprule
\textbf{Domain} & \textbf{Hospital} & \textbf{M-U (95\% CI)} & \textbf{Holm P} & \textbf{M-R (95\% CI)} & \textbf{Holm P} \\
\midrule
\endfirsthead
\toprule
\textbf{Domain} & \textbf{Hospital} & \textbf{M-U (95\% CI)} & \textbf{Holm P} & \textbf{M-R (95\% CI)} & \textbf{Holm P} \\
\midrule
\endhead
\midrule
\multicolumn{6}{r}{\footnotesize Continued on next page} \\
\endfoot
\bottomrule
\endlastfoot
M1 & General & 17.48 (13.15 to 21.81) & <0.001 & 7.47 (2.93 to 12.01) & 0.077 \\
M1 & Specialty & 18.07 (10.62 to 25.53) & 0.005 & 4.91 (-7.03 to 16.85) & 1.000 \\
M2 & General & 15.34 (8.70 to 21.98) & 0.002 & 6.93 (3.47 to 10.39) & 0.011 \\
M2 & Specialty & 7.44 (-1.67 to 16.56) & 1.000 & 2.11 (-8.32 to 12.54) & 1.000 \\
M3 & General & 25.77 (19.97 to 31.57) & <0.001 & 9.04 (3.40 to 14.67) & 0.096 \\
M3 & Specialty & 19.15 (8.88 to 29.42) & 0.051 & 0.44 (-13.22 to 14.10) & 1.000 \\
M4 & General & 17.61 (13.01 to 22.21) & <0.001 & 5.57 (1.65 to 9.50) & 0.242 \\
M4 & Specialty & 10.18 (1.04 to 19.33) & 0.972 & 3.85 (-5.04 to 12.75) & 1.000 \\
M5 & General & 6.65 (2.65 to 10.64) & 0.072 & 4.90 (1.63 to 8.17) & 0.164 \\
M5 & Specialty & 5.96 (-1.00 to 12.92) & 1.000 & 4.54 (-2.67 to 11.74) & 1.000 \\
\end{longtable}
\endgroup

\textbf{Both axes}

\begingroup
\scriptsize
\setlength{\tabcolsep}{3pt}
\renewcommand{\arraystretch}{1.12}
\begin{longtable}{@{}>{\raggedright\arraybackslash}p{0.157\linewidth}>{\raggedright\arraybackslash}p{0.157\linewidth}>{\raggedright\arraybackslash}p{0.157\linewidth}>{\raggedright\arraybackslash}p{0.157\linewidth}>{\raggedright\arraybackslash}p{0.157\linewidth}>{\raggedright\arraybackslash}p{0.157\linewidth}@{}}
\toprule
\textbf{Domain} & \textbf{Hospital} & \textbf{M-U (95\% CI)} & \textbf{Holm P} & \textbf{M-R (95\% CI)} & \textbf{Holm P} \\
\midrule
\endfirsthead
\toprule
\textbf{Domain} & \textbf{Hospital} & \textbf{M-U (95\% CI)} & \textbf{Holm P} & \textbf{M-R (95\% CI)} & \textbf{Holm P} \\
\midrule
\endhead
\midrule
\multicolumn{6}{r}{\footnotesize Continued on next page} \\
\endfoot
\bottomrule
\endlastfoot
M1 & General & 13.60 (9.38 to 17.82) & <0.001 & 4.45 (0.72 to 8.17) & 0.665 \\
M1 & Specialty & 8.72 (2.77 to 14.66) & 0.268 & 1.29 (-4.31 to 6.89) & 1.000 \\
M2 & General & 4.29 (-1.13 to 9.71) & 1.000 & 3.60 (-0.45 to 7.66) & 1.000 \\
M2 & Specialty & 3.76 (-3.17 to 10.68) & 1.000 & 2.59 (-4.84 to 10.02) & 1.000 \\
M3 & General & 23.91 (18.51 to 29.32) & <0.001 & 8.79 (4.20 to 13.37) & 0.020 \\
M3 & Specialty & 13.33 (6.76 to 19.90) & 0.034 & 9.07 (-0.00 to 18.14) & 1.000 \\
M4 & General & 13.22 (8.62 to 17.82) & <0.001 & 3.35 (-1.46 to 8.16) & 1.000 \\
M4 & Specialty & 6.41 (-0.01 to 12.84) & 1.000 & 4.34 (-1.26 to 9.94) & 1.000 \\
M5 & General & 2.43 (-1.05 to 5.91) & 1.000 & -1.12 (-4.16 to 1.92) & 1.000 \\
M5 & Specialty & 1.45 (-2.50 to 5.40) & 1.000 & 2.54 (-3.46 to 8.55) & 1.000 \\
\end{longtable}
\endgroup

\subsubsection{c. Direct hospital increment differences}

\begingroup
\scriptsize
\setlength{\tabcolsep}{3pt}
\renewcommand{\arraystretch}{1.12}
\begin{longtable}{@{}>{\raggedright\arraybackslash}p{0.157\linewidth}>{\raggedright\arraybackslash}p{0.157\linewidth}>{\raggedright\arraybackslash}p{0.157\linewidth}>{\raggedright\arraybackslash}p{0.157\linewidth}>{\raggedright\arraybackslash}p{0.157\linewidth}>{\raggedright\arraybackslash}p{0.157\linewidth}@{}}
\toprule
\textbf{Category} & \textbf{Domain} & \textbf{M-U gap (95\% CI)} & \textbf{Holm P} & \textbf{M-R gap (95\% CI)} & \textbf{Holm P} \\
\midrule
\endfirsthead
\toprule
\textbf{Category} & \textbf{Domain} & \textbf{M-U gap (95\% CI)} & \textbf{Holm P} & \textbf{M-R gap (95\% CI)} & \textbf{Holm P} \\
\midrule
\endhead
\midrule
\multicolumn{6}{r}{\footnotesize Continued on next page} \\
\endfoot
\bottomrule
\endlastfoot
Comorbidity only & M1 & -5.38 (-13.45 to 2.70) & 1.000 & 3.34 (-5.43 to 12.12) & 1.000 \\
Comorbidity only & M2 & -9.28 (-22.15 to 3.59) & 1.000 & -13.88 (-22.47 to -5.29) & 0.075 \\
Comorbidity only & M3 & -0.79 (-14.20 to 12.62) & 1.000 & -5.19 (-14.95 to 4.58) & 1.000 \\
Comorbidity only & M4 & 0.09 (-10.61 to 10.80) & 1.000 & -2.43 (-10.68 to 5.83) & 1.000 \\
Comorbidity only & M5 & -2.47 (-11.25 to 6.30) & 1.000 & -1.88 (-10.59 to 6.83) & 1.000 \\
Lung cancer only & M1 & 0.59 (-7.33 to 8.52) & 1.000 & -2.56 (-14.92 to 9.79) & 1.000 \\
Lung cancer only & M2 & -7.90 (-18.72 to 2.93) & 1.000 & -4.82 (-15.68 to 6.04) & 1.000 \\
Lung cancer only & M3 & -6.62 (-17.98 to 4.74) & 1.000 & -8.60 (-22.63 to 5.44) & 1.000 \\
Lung cancer only & M4 & -7.43 (-17.54 to 2.68) & 1.000 & -1.72 (-11.15 to 7.72) & 1.000 \\
Lung cancer only & M5 & -0.69 (-8.32 to 6.94) & 1.000 & -0.36 (-7.89 to 7.16) & 1.000 \\
Both axes & M1 & -4.88 (-11.94 to 2.17) & 1.000 & -3.16 (-9.71 to 3.40) & 1.000 \\
Both axes & M2 & -0.53 (-9.23 to 8.16) & 1.000 & -1.01 (-9.33 to 7.31) & 1.000 \\
Both axes & M3 & -10.58 (-18.49 to -2.67) & 0.316 & 0.28 (-9.52 to 10.08) & 1.000 \\
Both axes & M4 & -6.80 (-14.22 to 0.61) & 1.000 & 0.98 (-5.99 to 7.96) & 1.000 \\
Both axes & M5 & -0.98 (-6.00 to 4.03) & 1.000 & 3.66 (-2.67 to 9.99) & 1.000 \\
\end{longtable}
\endgroup

Gaps are specialty minus general, adjusted across all 30 direct comparisons. The M2 comorbidity-only M-R gap is -13.88 (-22.47 to -5.29), Holm P = 0.075.

\subsubsection{d. Principal hospital-by-category heterogeneity tests}

\begingroup
\scriptsize
\setlength{\tabcolsep}{3pt}
\renewcommand{\arraystretch}{1.12}
\begin{longtable}{@{}>{\raggedright\arraybackslash}p{0.154\linewidth}>{\raggedright\arraybackslash}p{0.154\linewidth}>{\raggedright\arraybackslash}p{0.154\linewidth}>{\raggedright\arraybackslash}p{0.170\linewidth}>{\raggedright\arraybackslash}p{0.154\linewidth}>{\raggedright\arraybackslash}p{0.154\linewidth}@{}}
\toprule
\textbf{Domain} & \textbf{Contrast} & \textbf{F} & \textbf{df (numerator /\allowbreak{} denominator)} & \textbf{Raw P} & \textbf{Holm P} \\
\midrule
\endfirsthead
\toprule
\textbf{Domain} & \textbf{Contrast} & \textbf{F} & \textbf{df (numerator /\allowbreak{} denominator)} & \textbf{Raw P} & \textbf{Holm P} \\
\midrule
\endhead
\midrule
\multicolumn{6}{r}{\footnotesize Continued on next page} \\
\endfoot
\bottomrule
\endlastfoot
M1 & M-U & 0.759 & 2 /\allowbreak{} 26.22 & 0.478 & 1.000 \\
M1 & M-R & 0.721 & 2 /\allowbreak{} 26.89 & 0.495 & 1.000 \\
M2 & M-U & 0.851 & 2 /\allowbreak{} 26.21 & 0.439 & 1.000 \\
M2 & M-R & 2.306 & 2 /\allowbreak{} 26.86 & 0.119 & 1.000 \\
M3 & M-U & 1.018 & 2 /\allowbreak{} 26.22 & 0.375 & 1.000 \\
M3 & M-R & 0.504 & 2 /\allowbreak{} 26.89 & 0.610 & 1.000 \\
M4 & M-U & 0.908 & 2 /\allowbreak{} 26.23 & 0.416 & 1.000 \\
M4 & M-R & 0.219 & 2 /\allowbreak{} 26.90 & 0.805 & 1.000 \\
M5 & M-U & 0.046 & 2 /\allowbreak{} 26.22 & 0.955 & 1.000 \\
M5 & M-R & 0.566 & 2 /\allowbreak{} 26.89 & 0.574 & 1.000 \\
\end{longtable}
\endgroup

The 10 joint tests ask whether hospital increment differences vary across categories; none retained Holm support. This is not evidence of equivalence.

\end{landscape}
\subsubsection{e. Category-pair comparisons and multiplicity inventory}

\begingroup
\small
\setlength{\tabcolsep}{3pt}
\renewcommand{\arraystretch}{1.12}
\begin{longtable}{@{}>{\raggedright\arraybackslash}p{0.313\linewidth}>{\raggedright\arraybackslash}p{0.313\linewidth}>{\raggedright\arraybackslash}p{0.313\linewidth}@{}}
\toprule
\textbf{Complete family} & \textbf{Comparisons} & \textbf{Holm P < 0.05} \\
\midrule
\endfirsthead
\toprule
\textbf{Complete family} & \textbf{Comparisons} & \textbf{Holm P < 0.05} \\
\midrule
\endhead
\midrule
\multicolumn{3}{r}{\footnotesize Continued on next page} \\
\endfoot
\bottomrule
\endlastfoot
category increment pairs & 30 & 0 \\
category U gap pairs & 15 & 1 \\
\end{longtable}
\endgroup

Complete raw P values, adjusted P values, effective degrees of freedom and all pairwise contrasts are in the accompanying aggregate Source Data. The U M3 hospital-gap change for both-complex minus lung-cancer-only cases was 12.31 (95\% CI, 5.16 to 19.46), Holm P = 0.023; individual category hospital gaps were -5.49 (-13.52 to 2.54) and 6.82 (-0.10 to 13.73), respectively. This secondary baseline finding was not retained after correction in several sensitivities.

\begin{landscape}
\subsubsection{f. Model diagnostics and sensitivity summaries}

\begingroup
\scriptsize
\setlength{\tabcolsep}{3pt}
\renewcommand{\arraystretch}{1.12}
\begin{longtable}{@{}>{\raggedright\arraybackslash}p{0.134\linewidth}>{\raggedright\arraybackslash}p{0.134\linewidth}>{\raggedright\arraybackslash}p{0.134\linewidth}>{\raggedright\arraybackslash}p{0.134\linewidth}>{\raggedright\arraybackslash}p{0.134\linewidth}>{\raggedright\arraybackslash}p{0.134\linewidth}>{\raggedright\arraybackslash}p{0.136\linewidth}@{}}
\toprule
\textbf{Domain} & \textbf{Responses} & \textbf{Physicians} & \textbf{Codes} & \textbf{Fixed rank} & \textbf{Singular} & \textbf{Predictions outside 0-100} \\
\midrule
\endfirsthead
\toprule
\textbf{Domain} & \textbf{Responses} & \textbf{Physicians} & \textbf{Codes} & \textbf{Fixed rank} & \textbf{Singular} & \textbf{Predictions outside 0-100} \\
\midrule
\endhead
\midrule
\multicolumn{7}{r}{\footnotesize Continued on next page} \\
\endfoot
\bottomrule
\endlastfoot
M1 & 2133 & 237 & 91 & 143/\allowbreak{}143 & 0 & 17 \\
M2 & 2133 & 237 & 91 & 143/\allowbreak{}143 & 0 & 285 \\
M3 & 2133 & 237 & 91 & 143/\allowbreak{}143 & 0 & 65 \\
M4 & 2133 & 237 & 91 & 143/\allowbreak{}143 & 0 & 52 \\
M5 & 2133 & 237 & 91 & 143/\allowbreak{}143 & 0 & 13 \\
\end{longtable}
\endgroup

\end{landscape}
\begingroup
\scriptsize
\setlength{\tabcolsep}{3pt}
\renewcommand{\arraystretch}{1.12}
\begin{longtable}{@{}>{\raggedright\arraybackslash}p{0.217\linewidth}>{\raggedright\arraybackslash}p{0.179\linewidth}>{\raggedright\arraybackslash}p{0.187\linewidth}>{\raggedright\arraybackslash}p{0.179\linewidth}>{\raggedright\arraybackslash}p{0.179\linewidth}@{}}
\toprule
\textbf{Sensitivity} & \textbf{Domain responses} & \textbf{Holm-supported joint tests} & \textbf{Minimum Holm P} & \textbf{Minimum denominator df} \\
\midrule
\endfirsthead
\toprule
\textbf{Sensitivity} & \textbf{Domain responses} & \textbf{Holm-supported joint tests} & \textbf{Minimum Holm P} & \textbf{Minimum denominator df} \\
\midrule
\endhead
\midrule
\multicolumn{5}{r}{\footnotesize Continued on next page} \\
\endfoot
\bottomrule
\endlastfoot
main & 2133-2133 & 0/\allowbreak{}10 & 1.000 & 26.21 \\
strict direct source & 2034-2034 & 0/\allowbreak{}10 & 1.000 & 24.88 \\
alias cluster & 2133-2133 & 0/\allowbreak{}10 & 1.000 & 26.02 \\
hospital equal & 2133-2133 & 0/\allowbreak{}10 & 0.338 & 29.23 \\
bounded & 2133-2133 & 0/\allowbreak{}10 & 1.000 & 24.18 \\
common39 & 2088-2088 & 0/\allowbreak{}10 & 0.949 & 25.94 \\
case hospital baseline & 2133-2133 & 0/\allowbreak{}10 & 1.000 & 24.25 \\
judge claude-opus-4.7 & 2132-2133 & 0/\allowbreak{}10 & 1.000 & 26.13 \\
leave out claude-opus-4.7 & 2133-2133 & 0/\allowbreak{}10 & 0.853 & 26.20 \\
judge gemini-3.1-pro-preview & 1898-2133 & 0/\allowbreak{}10 & 1.000 & 25.23 \\
leave out gemini-3.1-pro-preview & 2133-2133 & 0/\allowbreak{}10 & 0.720 & 26.21 \\
judge gpt-5.5 & 2129-2133 & 0/\allowbreak{}10 & 0.153 & 26.19 \\
leave out gpt-5.5 & 2133-2133 & 0/\allowbreak{}10 & 1.000 & 26.21 \\
\end{longtable}
\endgroup

All principal and sensitivity joint families are preserved, including negative contrasts and differing targets. Hospital-equal weighting changes the estimand; bounded models retain the response-level scale; single-judge branches retain actual applicable denominators. One empty specialty R case cell is retained in the main target. The 39-case sensitivity used 2,088 responses. Two hundred leave-one-case fits report influence ranges only. Full sensitivity contrasts and leave-one-case summaries are supplied as aggregate interfaces; they do not replace the main estimates.

\section{Supplementary Table 10. Operational criteria, complete counts and exploratory primary clinical-decision scenarios in the 100-case corpus}

\subsection{a. Operational definitions of gray-zone criteria}

\begingroup
\small
\setlength{\tabcolsep}{3pt}
\renewcommand{\arraystretch}{1.12}
\begin{longtable}{@{}>{\raggedright\arraybackslash}p{0.347\linewidth}>{\raggedright\arraybackslash}p{0.593\linewidth}@{}}
\toprule
\textbf{Gray-zone criterion} & \textbf{Operational definition} \\
\midrule
\endfirsthead
\toprule
\textbf{Gray-zone criterion} & \textbf{Operational definition} \\
\midrule
\endhead
\midrule
\multicolumn{2}{r}{\footnotesize Continued on next page} \\
\endfoot
\bottomrule
\endlastfoot
Guideline recommendation unclear & The guideline recommendation is unclear, or the combination of patient features does not map directly to a standard guideline scenario. \\
Major conflict from comorbidity or treatment constraints & An important non-cancer medical constraint creates conflict between otherwise plausible pathways. \\
Documented decision disagreement & The multidisciplinary record contains at least two distinct treatment recommendations or a documented disagreement between senior experts. \\
Pathway sequence or strategy choice may materially affect outcome & Treatment sequence, pathway combination or ordering may materially alter later options or outcomes. \\
Interacting factors limit direct application of a standard pathway & Molecular, pathological, imaging, comorbidity, treatment-history or other factors jointly prevent direct application of a standard pathway. \\
\end{longtable}
\endgroup

All five criteria were non-exclusive and recorded at case level.

\subsection{b. Counts of prespecified gray-zone attributes}

\begingroup
\footnotesize
\setlength{\tabcolsep}{3pt}
\renewcommand{\arraystretch}{1.12}
\begin{longtable}{@{}>{\raggedright\arraybackslash}p{0.342\linewidth}>{\raggedright\arraybackslash}p{0.167\linewidth}>{\raggedright\arraybackslash}p{0.180\linewidth}>{\raggedright\arraybackslash}p{0.251\linewidth}@{}}
\toprule
\textbf{Prespecified attribute (non-exclusive)} & \textbf{Overall (N=100)} & \textbf{Expert-reference set (N=40)} & \textbf{Source-masked content-evaluation set (N=60)} \\
\midrule
\endfirsthead
\toprule
\textbf{Prespecified attribute (non-exclusive)} & \textbf{Overall (N=100)} & \textbf{Expert-reference set (N=40)} & \textbf{Source-masked content-evaluation set (N=60)} \\
\midrule
\endhead
\midrule
\multicolumn{4}{r}{\footnotesize Continued on next page} \\
\endfoot
\bottomrule
\endlastfoot
Guideline recommendation unclear & 74 (74.0\%) & 26 (65.0\%) & 48 (80.0\%) \\
Major conflict from comorbidity or treatment constraints & 47 (47.0\%) & 26 (65.0\%) & 21 (35.0\%) \\
Documented decision disagreement & 1 (1.0\%) & 0 & 1 (1.7\%) \\
Pathway sequence or strategy choice may materially affect outcome & 45 (45.0\%) & 1 (2.5\%) & 44 (73.3\%) \\
Interacting factors limit direct application of a standard pathway & 63 (63.0\%) & 11 (27.5\%) & 52 (86.7\%) \\
At least two prespecified gray-zone criteria & 78 (78.0\%) & 24 (60.0\%) & 54 (90.0\%) \\
\end{longtable}
\endgroup

The attributes describe the purposively assembled case materials and could co-occur. The expert-reference and source-masked content-evaluation sets served different study purposes; no baseline-balance test was performed. Adding documented decision disagreement did not change the count of 78 cases meeting at least two criteria. Pathway-sequence attributes were not emphasized in main-text Table 1 because the two sets differed in construction purpose and source composition.

\subsection{c. Exploratory primary clinical-decision scenarios}

\begingroup
\footnotesize
\setlength{\tabcolsep}{3pt}
\renewcommand{\arraystretch}{1.12}
\begin{longtable}{@{}>{\raggedright\arraybackslash}p{0.335\linewidth}>{\raggedright\arraybackslash}p{0.169\linewidth}>{\raggedright\arraybackslash}p{0.182\linewidth}>{\raggedright\arraybackslash}p{0.254\linewidth}@{}}
\toprule
\textbf{Primary clinical-decision scenario} & \textbf{Overall (N=100)} & \textbf{Expert-reference set (N=40)} & \textbf{Source-masked content-evaluation set (N=60)} \\
\midrule
\endfirsthead
\toprule
\textbf{Primary clinical-decision scenario} & \textbf{Overall (N=100)} & \textbf{Expert-reference set (N=40)} & \textbf{Source-masked content-evaluation set (N=60)} \\
\midrule
\endhead
\midrule
\multicolumn{4}{r}{\footnotesize Continued on next page} \\
\endfoot
\bottomrule
\endlastfoot
Diagnosis, staging, tumor origin or biological boundary & 19 (19.0\%) & 11 (27.5\%) & 8 (13.3\%) \\
Local-treatment resectability and local control & 12 (12.0\%) & 7 (17.5\%) & 5 (8.3\%) \\
Systemic-treatment selection, resistance and sequence & 22 (22.0\%) & 5 (12.5\%) & 17 (28.3\%) \\
Treatment toxicity, comorbidity and tolerance constraints & 32 (32.0\%) & 16 (40.0\%) & 16 (26.7\%) \\
Prioritization of coexisting disease or multiple primary tumors & 6 (6.0\%) & 1 (2.5\%) & 5 (8.3\%) \\
Acute critical illness and initial symptom stabilization & 9 (9.0\%) & 0 & 9 (15.0\%) \\
\end{longtable}
\endgroup

This mutually exclusive exploratory classification was AI assisted and used only to describe the principal decision scenario. It did not determine inclusion, grouping or performance evaluation and was not presented as a baseline characteristic in main-text Table 1. Five low-confidence classifications retained their original assignments and remained in the denominator.

\subsection{d. Physician-reviewed categories for the 40-case expert-reference set}

\begingroup
\footnotesize
\setlength{\tabcolsep}{3pt}
\renewcommand{\arraystretch}{1.12}
\begin{longtable}{@{}>{\raggedright\arraybackslash}p{0.235\linewidth}>{\raggedright\arraybackslash}p{0.235\linewidth}>{\raggedright\arraybackslash}p{0.235\linewidth}>{\raggedright\arraybackslash}p{0.235\linewidth}@{}}
\toprule
\textbf{Comorbidity-complex axis} & \textbf{Lung-cancer-complex axis} & \textbf{Category} & \textbf{Cases} \\
\midrule
\endfirsthead
\toprule
\textbf{Comorbidity-complex axis} & \textbf{Lung-cancer-complex axis} & \textbf{Category} & \textbf{Cases} \\
\midrule
\endhead
\midrule
\multicolumn{4}{r}{\footnotesize Continued on next page} \\
\endfoot
\bottomrule
\endlastfoot
Marked & Not marked & Comorbidity-complex only & 9 \\
Not marked & Marked & Lung-cancer-complex only & 13 \\
Marked & Marked & Complex on both axes & 18 \\
\end{longtable}
\endgroup

These post hoc labels are separate from the original five non-exclusive gray-zone inclusion attributes above. The comorbidity axis includes other tumors and accompanying disease, organ, nutritional or functional constraints; the lung-cancer axis concerns lung cancer and its treatment. Not marked means the reviewer did not label that axis complex, not that the disease or constraint was absent. Categories are unordered, with no neither-complex cases, and do not form a validated severity scale.

\section{Supplementary Table 11. Audit of structural relations in complete MCE outputs}

This table reports the five-relation assessment of 40 complete MCE Strategy Review Packs. The unit was one complete output. The three judges are reported separately without voting or averaging. Verification tasks were judge--task observations nested within outputs and were not independent clinical samples. The audit evaluated connections among candidate pathways, their premises, verification results and subsequent actions in the text. Part c lists outputs with a conflict, conditional misconnection or omission flagged by at least one AI judge; targeted human checking is reported separately in part d.

\begin{landscape}
\subsection{a. Judge-specific dimensions and overall closure}

\begingroup
\scriptsize
\setlength{\tabcolsep}{3pt}
\renewcommand{\arraystretch}{1.12}
\begin{longtable}{@{}>{\raggedright\arraybackslash}p{0.104\linewidth}>{\raggedright\arraybackslash}p{0.137\linewidth}>{\raggedright\arraybackslash}p{0.108\linewidth}>{\raggedright\arraybackslash}p{0.114\linewidth}>{\raggedright\arraybackslash}p{0.131\linewidth}>{\raggedright\arraybackslash}p{0.128\linewidth}>{\raggedright\arraybackslash}p{0.104\linewidth}>{\raggedright\arraybackslash}p{0.114\linewidth}@{}}
\toprule
\textbf{Judge} & \textbf{D1 candidate-pathway differentiation} & \textbf{D2 action-premise relation} & \textbf{D3 decision-changing unknown} & \textbf{D4 verification-candidate relation} & \textbf{D5 reassessment/\allowbreak{}fallback relation} & \textbf{All five complete} & \textbf{Overall relationship closure} \\
\midrule
\endfirsthead
\toprule
\textbf{Judge} & \textbf{D1 candidate-pathway differentiation} & \textbf{D2 action-premise relation} & \textbf{D3 decision-changing unknown} & \textbf{D4 verification-candidate relation} & \textbf{D5 reassessment/\allowbreak{}fallback relation} & \textbf{All five complete} & \textbf{Overall relationship closure} \\
\midrule
\endhead
\midrule
\multicolumn{8}{r}{\footnotesize Continued on next page} \\
\endfoot
\bottomrule
\endlastfoot
Gemini 3.1 Pro Preview & 40/\allowbreak{}40 & 37/\allowbreak{}40 & 40/\allowbreak{}40 & 38/\allowbreak{}40 & 40/\allowbreak{}40 & 36/\allowbreak{}40 & 36/\allowbreak{}40 \\
GPT-5.5 & 40/\allowbreak{}40 & 31/\allowbreak{}40 & 40/\allowbreak{}40 & 37/\allowbreak{}40 & 40/\allowbreak{}40 & 30/\allowbreak{}40 & 30/\allowbreak{}40 \\
Claude Opus 4.7 & 40/\allowbreak{}40 & 40/\allowbreak{}40 & 40/\allowbreak{}40 & 40/\allowbreak{}40 & 40/\allowbreak{}40 & 40/\allowbreak{}40 & 40/\allowbreak{}40 \\
\end{longtable}
\endgroup

\end{landscape}
\subsection{b. Three-state subsequent actions for verification tasks}

\begingroup
\scriptsize
\setlength{\tabcolsep}{3pt}
\renewcommand{\arraystretch}{1.12}
\begin{longtable}{@{}>{\raggedright\arraybackslash}p{0.184\linewidth}>{\raggedright\arraybackslash}p{0.163\linewidth}>{\raggedright\arraybackslash}p{0.193\linewidth}>{\raggedright\arraybackslash}p{0.206\linewidth}>{\raggedright\arraybackslash}p{0.193\linewidth}@{}}
\toprule
\textbf{Judge} & \textbf{Verification tasks} & \textbf{Subsequent action after support} & \textbf{Subsequent action after refutation} & \textbf{Subsequent action if unresolved} \\
\midrule
\endfirsthead
\toprule
\textbf{Judge} & \textbf{Verification tasks} & \textbf{Subsequent action after support} & \textbf{Subsequent action after refutation} & \textbf{Subsequent action if unresolved} \\
\midrule
\endhead
\midrule
\multicolumn{5}{r}{\footnotesize Continued on next page} \\
\endfoot
\bottomrule
\endlastfoot
Gemini 3.1 Pro Preview & 137 & 137/\allowbreak{}137 & 137/\allowbreak{}137 & 137/\allowbreak{}137 \\
GPT-5.5 & 145 & 145/\allowbreak{}145 & 145/\allowbreak{}145 & 144/\allowbreak{}145 \\
Claude Opus 4.7 & 135 & 135/\allowbreak{}135 & 135/\allowbreak{}135 & 135/\allowbreak{}135 \\
Total judge--task observations & 417 & 417/\allowbreak{}417 & 417/\allowbreak{}417 & 416/\allowbreak{}417 \\
\end{longtable}
\endgroup

\subsection{c. Outputs with judge-coded defects}

\begingroup
\footnotesize
\setlength{\tabcolsep}{3pt}
\renewcommand{\arraystretch}{1.12}
\begin{longtable}{@{}>{\raggedright\arraybackslash}p{0.235\linewidth}>{\raggedright\arraybackslash}p{0.235\linewidth}>{\raggedright\arraybackslash}p{0.235\linewidth}>{\raggedright\arraybackslash}p{0.235\linewidth}@{}}
\toprule
\textbf{Anonymized output} & \textbf{Gemini 3.1 Pro Preview} & \textbf{GPT-5.5} & \textbf{Claude Opus 4.7} \\
\midrule
\endfirsthead
\toprule
\textbf{Anonymized output} & \textbf{Gemini 3.1 Pro Preview} & \textbf{GPT-5.5} & \textbf{Claude Opus 4.7} \\
\midrule
\endhead
\midrule
\multicolumn{4}{r}{\footnotesize Continued on next page} \\
\endfoot
\bottomrule
\endlastfoot
Output 1 & CONFLICT & CONFLICT & NONE \\
Output 2 & NONE & CONFLICT & NONE \\
Output 3 & NONE & CONFLICT & NONE \\
Output 4 & NONE & CONFLICT & NONE \\
Output 5 & MISCONN\-ECTION & CONFLICT/\allowbreak{}MISCONN\-ECTION & NONE \\
Output 6 & NONE & CONFLICT & NONE \\
Output 7 & NONE & CONFLICT & NONE \\
Output 8 & CONFLICT & NONE & NONE \\
Output 9 & NONE & OMISSION & NONE \\
Output 10 & NONE & CONFLICT & NONE \\
Output 11 & CONFLICT/\allowbreak{}OMISSION & NONE & NONE \\
Output 12 & NONE & CONFLICT & NONE \\
\end{longtable}
\endgroup

Atomic defect labels totaled 12 CONFLICT, 2 MISCONNECTION, 2 OMISSION and 0 NOT\_EVALUABLE calls.

\subsection{d. Targeted human checking}

\begingroup
\small
\setlength{\tabcolsep}{3pt}
\renewcommand{\arraystretch}{1.12}
\begin{longtable}{@{}>{\raggedright\arraybackslash}p{0.235\linewidth}>{\raggedright\arraybackslash}p{0.208\linewidth}>{\raggedright\arraybackslash}p{0.497\linewidth}@{}}
\toprule
\textbf{Review target} & \textbf{Items} & \textbf{Result} \\
\midrule
\endfirsthead
\toprule
\textbf{Review target} & \textbf{Items} & \textbf{Result} \\
\midrule
\endhead
\midrule
\multicolumn{3}{r}{\footnotesize Continued on next page} \\
\endfoot
\bottomrule
\endlastfoot
Still-unresolved branches & 10 & Subsequent action confirmed in 9; one omission confirmed \\
Additional judgment rationale & 8 & Text evidence and location recorded for supportive, refuting and unresolved states \\
Total reviewed objects & 18 & All checked against the corresponding output \\
\end{longtable}
\endgroup

Judge-specific relationship and validation-state summaries are provided in Source Data. Object-level ratings and review records remain governed materials and are not included in the distributable package.

\section{Supplementary Table 12. Case provenance, fixed input snapshot and published-source mapping}

\begingroup
\small
\setlength{\tabcolsep}{3pt}
\renewcommand{\arraystretch}{1.12}
\begin{longtable}{@{}>{\raggedright\arraybackslash}p{0.179\linewidth}>{\raggedright\arraybackslash}p{0.549\linewidth}>{\raggedright\arraybackslash}p{0.212\linewidth}@{}}
\toprule
\textbf{Study characteristic} & \textbf{Description} & \textbf{Supporting documentation} \\
\midrule
\endfirsthead
\toprule
\textbf{Study characteristic} & \textbf{Description} & \textbf{Supporting documentation} \\
\midrule
\endhead
\midrule
\multicolumn{3}{r}{\footnotesize Continued on next page} \\
\endfoot
\bottomrule
\endlastfoot
Cases and study sets & 100 cases: 40 in the expert-reference set and 60 in the source-masked content-evaluation set & Case identifier and study-set membership \\
Case sources & 47 restricted internal clinical cases and 53 published cases & Source type and public-source identifier \\
Inclusion and grouping & Purposively assembled using prespecified gray-zone attributes; 40 internal cases formed the expert-reference set and the remaining 60 cases formed the source-masked content-evaluation set; 0 actively excluded & Analysis-set and source-set labels \\
Fixed case input & Six fields: basic information, chief complaint, present illness, past and other medical history, investigations, and diagnosis; excluded management, treating-physician recommendations, subsequent plans and outcomes & Input structure version and cutoff \\
De-identification and release boundary & Internal cases were limited, de-identified restricted data; submission materials exclude case narratives and access logs & De-identification category and access conditions \\
Identifier and integrity & 100 unique case identifiers and unique decision-input and full-record integrity signatures; no identifier or input overlap between study sets & Case identity and content integrity \\
Development exposure and reuse & All cases had entered development or evaluation; reuse across experimental arms, generation or scoring did not increase the number of independent cases & Development exposure and case-reuse unit \\
Published sources & 49 publications mapped to 53 published cases; 48 identified by DOI and one by PMID and a stable PubMed record; four publications contributed two cases each & Publication identifier and case mapping \\
\end{longtable}
\endgroup

Case-level provenance and public-source crosswalks support verification of source, version and analysis unit. Internal clinical narratives are not publicly released; their ethical basis and access conditions belong in the ethics and data-availability statements.

\section{Supplementary Table 13. Clinical-content interpretation of existing EASS requirements}

\subsection{a. Complete non-exclusive marker coverage}

\begingroup
\footnotesize
\setlength{\tabcolsep}{3pt}
\renewcommand{\arraystretch}{1.12}
\begin{longtable}{@{}>{\raggedright\arraybackslash}p{0.223\linewidth}>{\raggedright\arraybackslash}p{0.271\linewidth}>{\raggedright\arraybackslash}p{0.223\linewidth}>{\raggedright\arraybackslash}p{0.223\linewidth}@{}}
\toprule
\textbf{Category} & \textbf{Requirement marker} & \textbf{Cases} & \textbf{Rules} \\
\midrule
\endfirsthead
\toprule
\textbf{Category} & \textbf{Requirement marker} & \textbf{Cases} & \textbf{Rules} \\
\midrule
\endhead
\midrule
\multicolumn{4}{r}{\footnotesize Continued on next page} \\
\endfoot
\bottomrule
\endlastfoot
Comorbidity only & Lung-cancer decision & 9/\allowbreak{}9 & 24/\allowbreak{}54 (44.4\%) \\
Comorbidity only & Accompanying constraint & 9/\allowbreak{}9 & 23/\allowbreak{}54 (42.6\%) \\
Comorbidity only & Explicit constraint-pathway link & 8/\allowbreak{}9 & 15/\allowbreak{}54 (27.8\%) \\
Lung cancer only & Lung-cancer decision & 13/\allowbreak{}13 & 33/\allowbreak{}78 (42.3\%) \\
Lung cancer only & Accompanying constraint & 6/\allowbreak{}13 & 6/\allowbreak{}78 (7.7\%) \\
Lung cancer only & Explicit constraint-pathway link & 4/\allowbreak{}13 & 4/\allowbreak{}78 (5.1\%) \\
Both axes & Lung-cancer decision & 18/\allowbreak{}18 & 47/\allowbreak{}108 (43.5\%) \\
Both axes & Accompanying constraint & 15/\allowbreak{}18 & 33/\allowbreak{}108 (30.6\%) \\
Both axes & Explicit constraint-pathway link & 14/\allowbreak{}18 & 26/\allowbreak{}108 (24.1\%) \\
\end{longtable}
\endgroup

\subsection{b. Representative requirement content}

\begingroup
\small
\setlength{\tabcolsep}{3pt}
\renewcommand{\arraystretch}{1.12}
\begin{longtable}{@{}>{\raggedright\arraybackslash}p{0.181\linewidth}>{\raggedright\arraybackslash}p{0.759\linewidth}@{}}
\toprule
\textbf{Category} & \textbf{Decision requirement (English paraphrase)} \\
\midrule
\endfirsthead
\toprule
\textbf{Category} & \textbf{Decision requirement (English paraphrase)} \\
\midrule
\endhead
\midrule
\multicolumn{2}{r}{\footnotesize Continued on next page} \\
\endfoot
\bottomrule
\endlastfoot
Comorbidity only & Determine whether accompanying effusion and a bladder lesion need characterization before systemic therapy. \\
Lung cancer only & If active infection is indicated, first adjust the sequence of local treatment. \\
Both axes & If pathological-source findings or infection and prior severe targeted-therapy pulmonary toxicity alter eligibility, redirect to the corresponding pathology-specific or safety-prioritized pathway. \\
\end{longtable}
\endgroup

Examples paraphrase existing explicit requirements; they are not new treatment recommendations or newly adjudicated clinical outcomes.

\begin{landscape}
\subsection{c. Complete descriptive composite-rule scores}

\textbf{Lung-cancer decision}

\begingroup
\scriptsize
\setlength{\tabcolsep}{3pt}
\renewcommand{\arraystretch}{1.12}
\begin{longtable}{@{}>{\raggedright\arraybackslash}p{0.157\linewidth}>{\raggedright\arraybackslash}p{0.157\linewidth}>{\raggedright\arraybackslash}p{0.157\linewidth}>{\raggedright\arraybackslash}p{0.157\linewidth}>{\raggedright\arraybackslash}p{0.157\linewidth}>{\raggedright\arraybackslash}p{0.157\linewidth}@{}}
\toprule
\textbf{Category} & \textbf{Hospital} & \textbf{U} & \textbf{R} & \textbf{M} & \textbf{Observed cases U/\allowbreak{}R/\allowbreak{}M} \\
\midrule
\endfirsthead
\toprule
\textbf{Category} & \textbf{Hospital} & \textbf{U} & \textbf{R} & \textbf{M} & \textbf{Observed cases U/\allowbreak{}R/\allowbreak{}M} \\
\midrule
\endhead
\midrule
\multicolumn{6}{r}{\footnotesize Continued on next page} \\
\endfoot
\bottomrule
\endlastfoot
Comorbidity only & General & 44.69 & 54.99 & 58.34 & 9/\allowbreak{}9/\allowbreak{}9 \\
Comorbidity only & Specialty & 42.13 & 51.57 & 54.91 & 9/\allowbreak{}9/\allowbreak{}9 \\
Lung cancer only & General & 38.52 & 55.80 & 63.36 & 13/\allowbreak{}13/\allowbreak{}13 \\
Lung cancer only & Specialty & 39.22 & 55.84 & 57.50 & 13/\allowbreak{}12/\allowbreak{}13 \\
Both axes & General & 45.16 & 53.36 & 62.67 & 18/\allowbreak{}18/\allowbreak{}18 \\
Both axes & Specialty & 47.79 & 53.17 & 58.50 & 18/\allowbreak{}18/\allowbreak{}18 \\
\end{longtable}
\endgroup

\textbf{Accompanying constraint}

\begingroup
\scriptsize
\setlength{\tabcolsep}{3pt}
\renewcommand{\arraystretch}{1.12}
\begin{longtable}{@{}>{\raggedright\arraybackslash}p{0.157\linewidth}>{\raggedright\arraybackslash}p{0.157\linewidth}>{\raggedright\arraybackslash}p{0.157\linewidth}>{\raggedright\arraybackslash}p{0.157\linewidth}>{\raggedright\arraybackslash}p{0.157\linewidth}>{\raggedright\arraybackslash}p{0.157\linewidth}@{}}
\toprule
\textbf{Category} & \textbf{Hospital} & \textbf{U} & \textbf{R} & \textbf{M} & \textbf{Observed cases U/\allowbreak{}R/\allowbreak{}M} \\
\midrule
\endfirsthead
\toprule
\textbf{Category} & \textbf{Hospital} & \textbf{U} & \textbf{R} & \textbf{M} & \textbf{Observed cases U/\allowbreak{}R/\allowbreak{}M} \\
\midrule
\endhead
\midrule
\multicolumn{6}{r}{\footnotesize Continued on next page} \\
\endfoot
\bottomrule
\endlastfoot
Comorbidity only & General & 43.02 & 52.24 & 59.04 & 9/\allowbreak{}9/\allowbreak{}9 \\
Comorbidity only & Specialty & 39.90 & 52.18 & 54.53 & 9/\allowbreak{}9/\allowbreak{}9 \\
Lung cancer only & General & 20.98 & 41.36 & 58.99 & 6/\allowbreak{}6/\allowbreak{}6 \\
Lung cancer only & Specialty & 17.04 & 36.30 & 36.10 & 6/\allowbreak{}5/\allowbreak{}6 \\
Both axes & General & 44.79 & 53.80 & 63.86 & 15/\allowbreak{}15/\allowbreak{}15 \\
Both axes & Specialty & 48.17 & 51.31 & 59.08 & 15/\allowbreak{}15/\allowbreak{}15 \\
\end{longtable}
\endgroup

\textbf{Explicit constraint-pathway link}

\begingroup
\scriptsize
\setlength{\tabcolsep}{3pt}
\renewcommand{\arraystretch}{1.12}
\begin{longtable}{@{}>{\raggedright\arraybackslash}p{0.157\linewidth}>{\raggedright\arraybackslash}p{0.157\linewidth}>{\raggedright\arraybackslash}p{0.157\linewidth}>{\raggedright\arraybackslash}p{0.157\linewidth}>{\raggedright\arraybackslash}p{0.157\linewidth}>{\raggedright\arraybackslash}p{0.157\linewidth}@{}}
\toprule
\textbf{Category} & \textbf{Hospital} & \textbf{U} & \textbf{R} & \textbf{M} & \textbf{Observed cases U/\allowbreak{}R/\allowbreak{}M} \\
\midrule
\endfirsthead
\toprule
\textbf{Category} & \textbf{Hospital} & \textbf{U} & \textbf{R} & \textbf{M} & \textbf{Observed cases U/\allowbreak{}R/\allowbreak{}M} \\
\midrule
\endhead
\midrule
\multicolumn{6}{r}{\footnotesize Continued on next page} \\
\endfoot
\bottomrule
\endlastfoot
Comorbidity only & General & 41.50 & 51.45 & 57.33 & 8/\allowbreak{}8/\allowbreak{}8 \\
Comorbidity only & Specialty & 39.60 & 48.56 & 54.75 & 8/\allowbreak{}8/\allowbreak{}8 \\
Lung cancer only & General & 18.01 & 46.24 & 57.38 & 4/\allowbreak{}4/\allowbreak{}4 \\
Lung cancer only & Specialty & 20.17 & 40.93 & 25.81 & 4/\allowbreak{}4/\allowbreak{}4 \\
Both axes & General & 43.55 & 54.59 & 63.34 & 14/\allowbreak{}14/\allowbreak{}14 \\
Both axes & Specialty & 47.15 & 49.45 & 57.79 & 14/\allowbreak{}14/\allowbreak{}14 \\
\end{longtable}
\endgroup

Scores descriptively regroup original ratings for composite rules containing each marker. They are not isolated linkage scores or cross-specialty coordination rates. Markers overlap and have different rule/case sets. The both-complex linkage group contains 14 eligible cases; U-to-M means are 43.55 to 63.34 in general hospitals and 47.15 to 57.79 in specialty hospitals. Complete answers, scorable-rule counts and eligibility denominators are in the aggregate marker-score interface. No new clinical review or response scoring was performed.

\end{landscape}

\clearpage
\section*{Extended Data Figures}

\phantomsection
\label{fig:extended-1}
\begin{center}
\centering
\includegraphics[width=\textwidth,height=0.68\textheight,keepaspectratio]{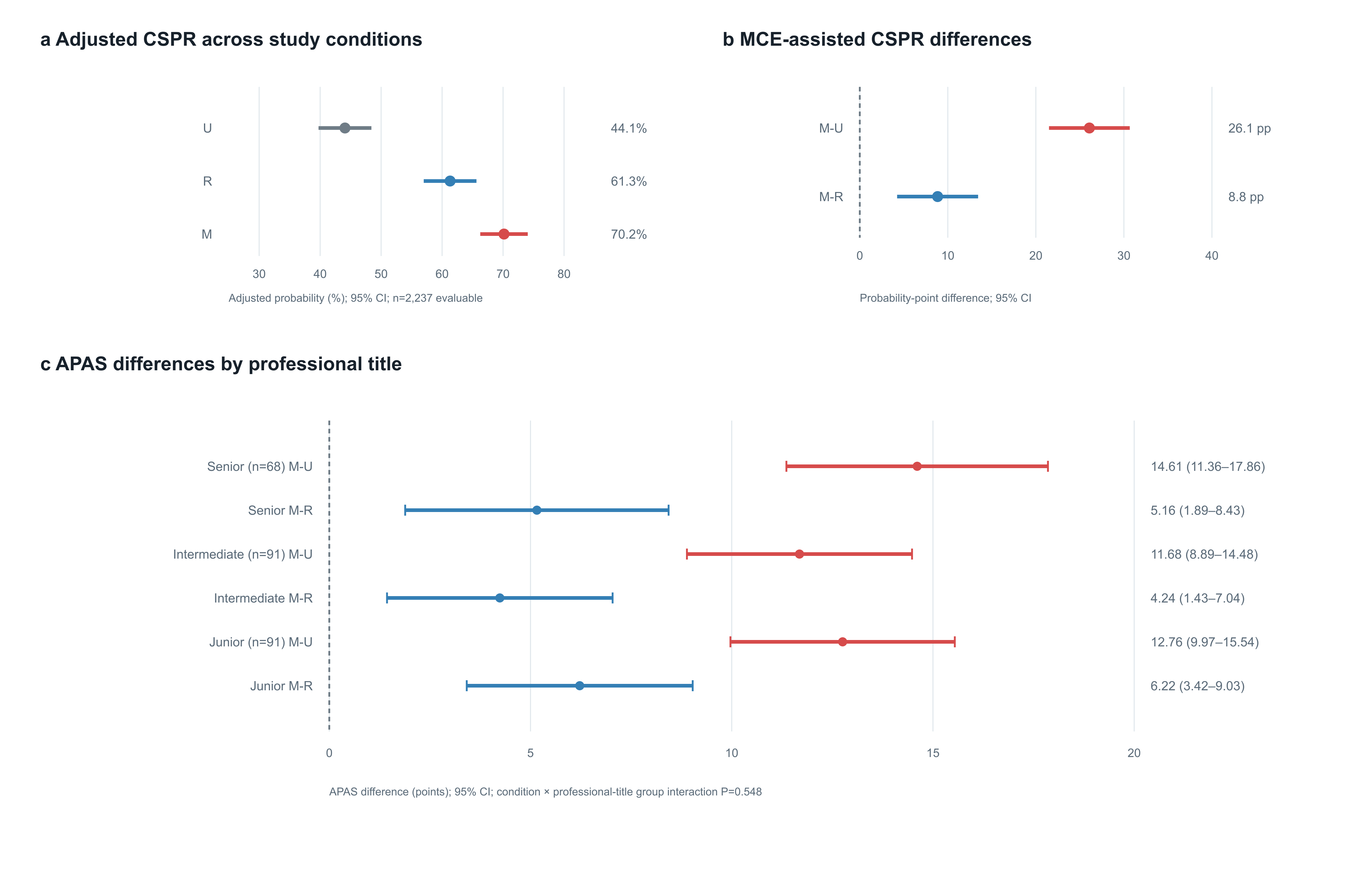}
\end{center}
{\small\noindent\textbf{Extended Data Figure 1 | Secondary CSPR gate and analysis by professional title} \textbf{a.} Adjusted CSPR estimates and 95\% CIs under the unaided (U), retrieval-reference (R) and MCE-assisted (M) conditions. \textbf{b.} Point differences in CSPR probability for M--U and M--R, with 95\% CIs. \textbf{c.} Adjusted M--U and M--R APAS differences, with 95\% CIs, in the senior, intermediate and junior professional-title groups; all six differences were positive and none of the six 95\% CIs crossed zero. CSPR was a secondary or exploratory rule-based gate; 2,237 responses were classifiable and 13 were indeterminate. The professional-title analysis included 68, 91 and 91 physicians in the three groups, respectively; the joint condition-by-professional-title-group interaction test gave P = 0.548 and was treated as exploratory. Repeated judge runs were measurement repeats, and professional title was treated only as a coarse indicator of clinical experience. Overall adjusted APAS estimates and primary contrasts are in Figure 3a; panel c contains the complete professional-title-group estimates, and physician-task M1--M5 clinical-content results are in Figure 3b.\par}

\clearpage
\phantomsection
\label{fig:extended-2}
\begin{center}
\centering
\includegraphics[width=\textwidth,height=0.68\textheight,keepaspectratio]{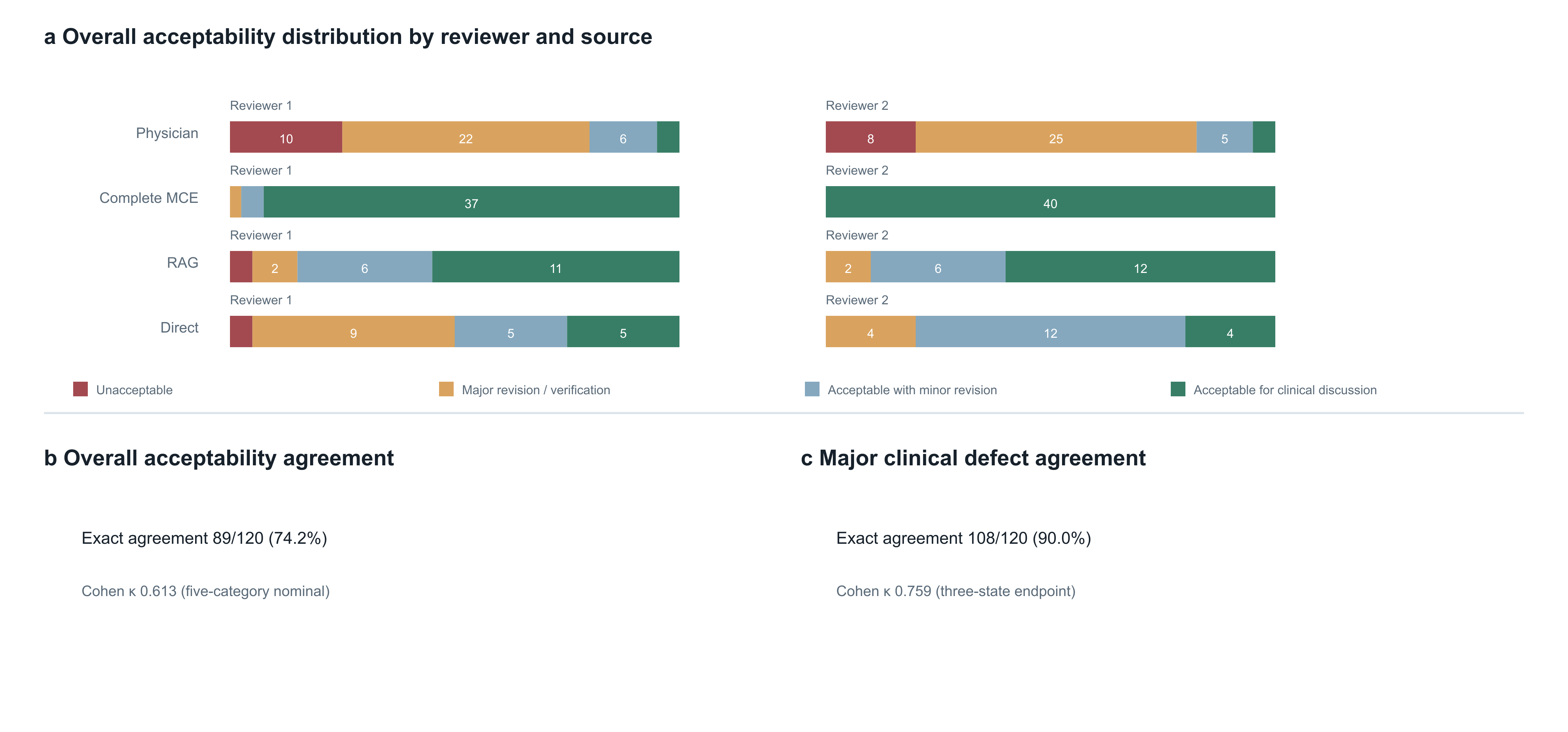}
\end{center}
{\small\noindent\textbf{Extended Data Figure 2 | Overall clinical anchoring distributions and agreement} Overall clinical anchoring covered 40 case clusters and 120 outputs. Clinical acceptability concerned suitability for clinical discussion and the extent of revision or additional verification required. The two reviewers' overall-acceptability distributions by output source are reported together with full-sample exact agreement and Cohen kappa for the five-category overall-acceptability item and the major-clinical-defect endpoint. Major clinical defects and overall clinical acceptability were treated as separate endpoints. Four-category ordinal and binary sensitivity estimates, with 95\% CIs, are reported in Supplementary Table 9i; neither reviewer used the \emph{not assessable} category in this sample. Source proportions describe the fixed enriched sample and are not confirmatory comparisons between sources. Figure 5a pools all 120 outputs for the correlation analysis.\par}

\clearpage
\phantomsection
\label{fig:extended-3}
\begin{center}
\centering
\includegraphics[width=\textwidth,height=0.68\textheight,keepaspectratio]{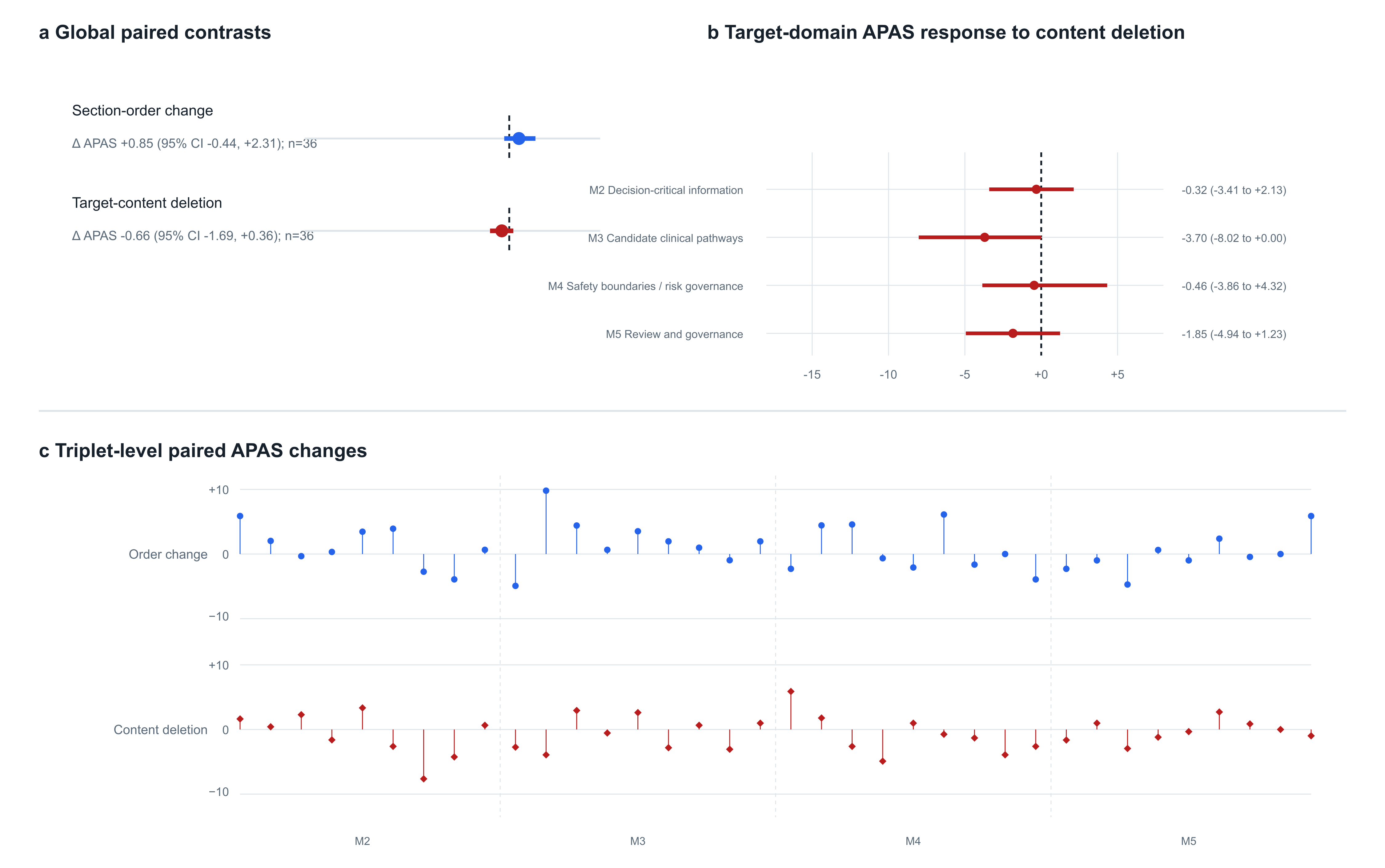}
\end{center}
{\small\noindent\textbf{Extended Data Figure 3 | Controlled-text perturbation diagnostics} The controlled-text perturbation analysis comprised 36 original--variant triplets and 108 texts. Section-order changes and target clinical-content deletion are reported as paired overall APAS differences, target-domain APAS differences for M2 decision-critical information, M3 candidate clinical pathways, M4 safety boundaries and risk governance, and M5 reassessment and governance, and paired changes for every triplet. Overall APAS and target-domain APAS were distinct estimands. The triplet was the paired unit and the 12 cases were the clustering units; partial gate states were supplementary descriptions and were not equivalent to complete CSPR. Limited section reordering produced no clear systematic score shift, while responses to targeted deletion were heterogeneous and domain-selective sensitivity was not established. Complete estimates are in Supplementary Table 9j.\par}

\clearpage
\phantomsection
\label{fig:extended-4}
\begin{center}
\centering
\includegraphics[width=\textwidth,height=0.68\textheight,keepaspectratio]{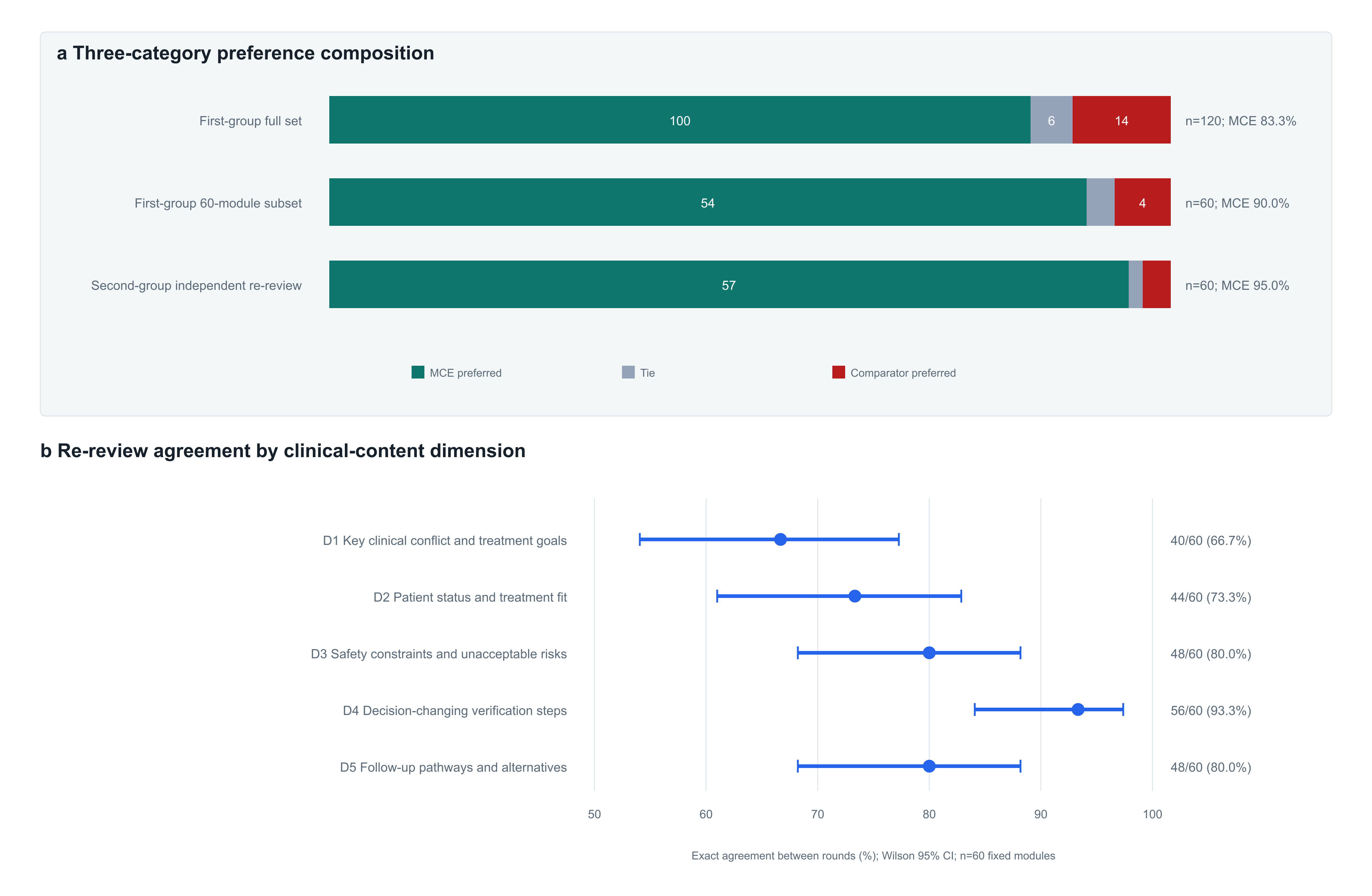}
\end{center}
{\small\noindent\textbf{Extended Data Figure 4 | Source-masked content evaluation} \textbf{a.} Three-category preference distributions for source-masked A/B content evaluation in the first-round set of 120 modules, the 60 modules selected for independent re-review and the 60 modules reviewed independently in the second round. \textbf{b.} Exact agreement and Wilson 95\% CIs across the two rounds for the 60 fixed text modules, for D1 key clinical conflicts and treatment goals, D2 patient state and treatment fit, D3 safety constraints and unacceptable risk, D4 decision-changing verification steps and D5 subsequent pathways and alternatives. The corresponding counts were 40/60, 44/60, 48/60, 56/60 and 48/60. The analysis unit was the within-case comparison module. Overall Gwet AC1 and the complete contingency tables are in Supplementary Table 6 and Source Data. Dimension-specific agreement describes reproducibility of judgments on the same fixed modules, not preference direction for the corresponding dimension. Source recognition was not directly measured; length, structure or style could still provide cues, and the results are not interpreted as clinical superiority.\par}

\clearpage
\phantomsection
\label{fig:extended-5}
\begin{center}
\centering
\includegraphics[width=\textwidth,height=0.68\textheight,keepaspectratio]{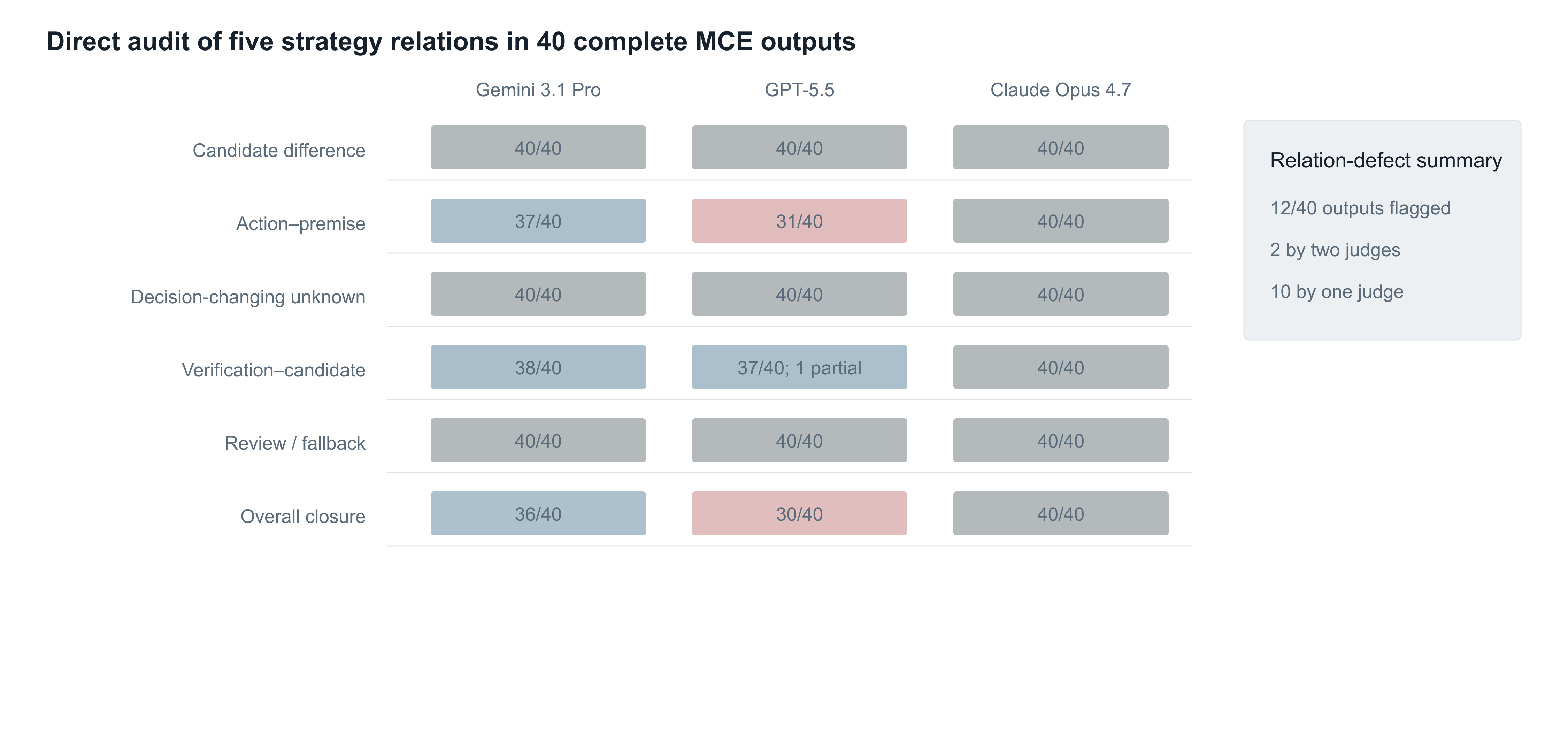}
\end{center}
{\small\noindent\textbf{Extended Data Figure 5 | Direct audit of prespecified strategy relations} In 40 fixed complete MCE outputs, Gemini 3.1 Pro Preview, GPT-5.5 and Claude Opus 4.7 independently assessed candidate-pathway differentiation, action--premise links, decision-changing unknowns, verification--candidate relations, reassessment/fallback and overall relationship closure. Cells show the number of outputs judged complete by each judge out of 40 while retaining partly complete states. In the right-hand summary, at least one AI judge flagged a conflict, conditional misconnection or omission in 12/40 outputs; two were flagged by two judges and ten by one judge. Judges are reported separately without voting or performance ranking. The 417 judge--verification-task observations, flag types and targeted source-text review are reported in Supplementary Table 11.\par}

\clearpage
\phantomsection
\label{fig:extended-6}
\begin{center}
\centering
\includegraphics[width=\textwidth,height=0.68\textheight,keepaspectratio]{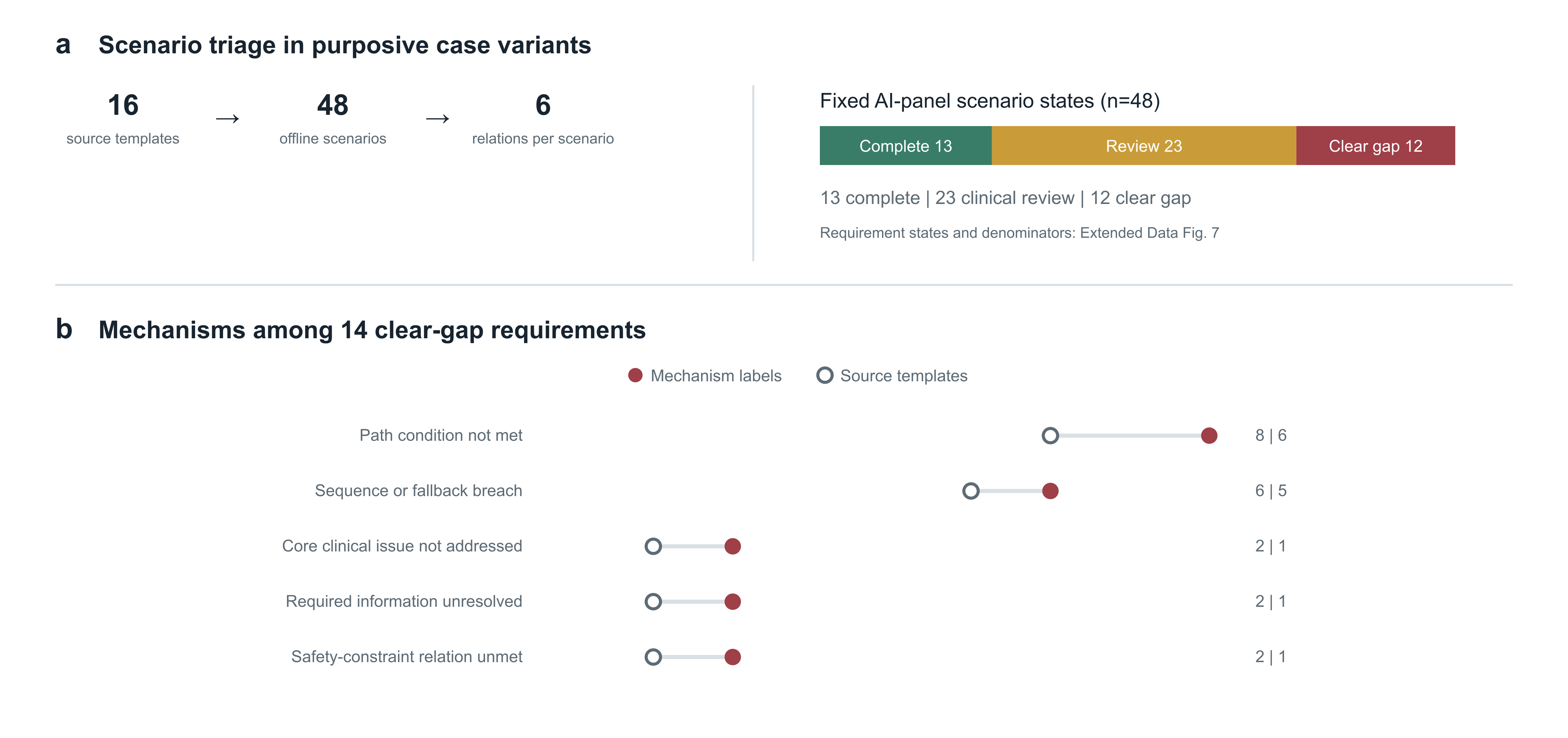}
\end{center}
{\small\noindent\textbf{Extended Data Figure 6 | Purposive case-variant triage and relationship-gap mechanisms} \textbf{a.} Sixteen source-case templates generated 48 purposive offline scenarios, each with six prespecified relationship requirements. The fixed AI-judge panel classified 13 scenarios as completely closed, 23 as requiring clinical review and 12 as containing at least one clear relationship gap. \textbf{b.} Fourteen clear gaps carried five non-exclusive mechanism labels; filled red points show label counts and open grey points the number of involved source templates. Complete requirement states and denominators are in Extended Data Figure 7.\par}

\clearpage
\phantomsection
\label{fig:extended-7}
\begin{center}
\centering
\includegraphics[width=\textwidth,height=0.68\textheight,keepaspectratio]{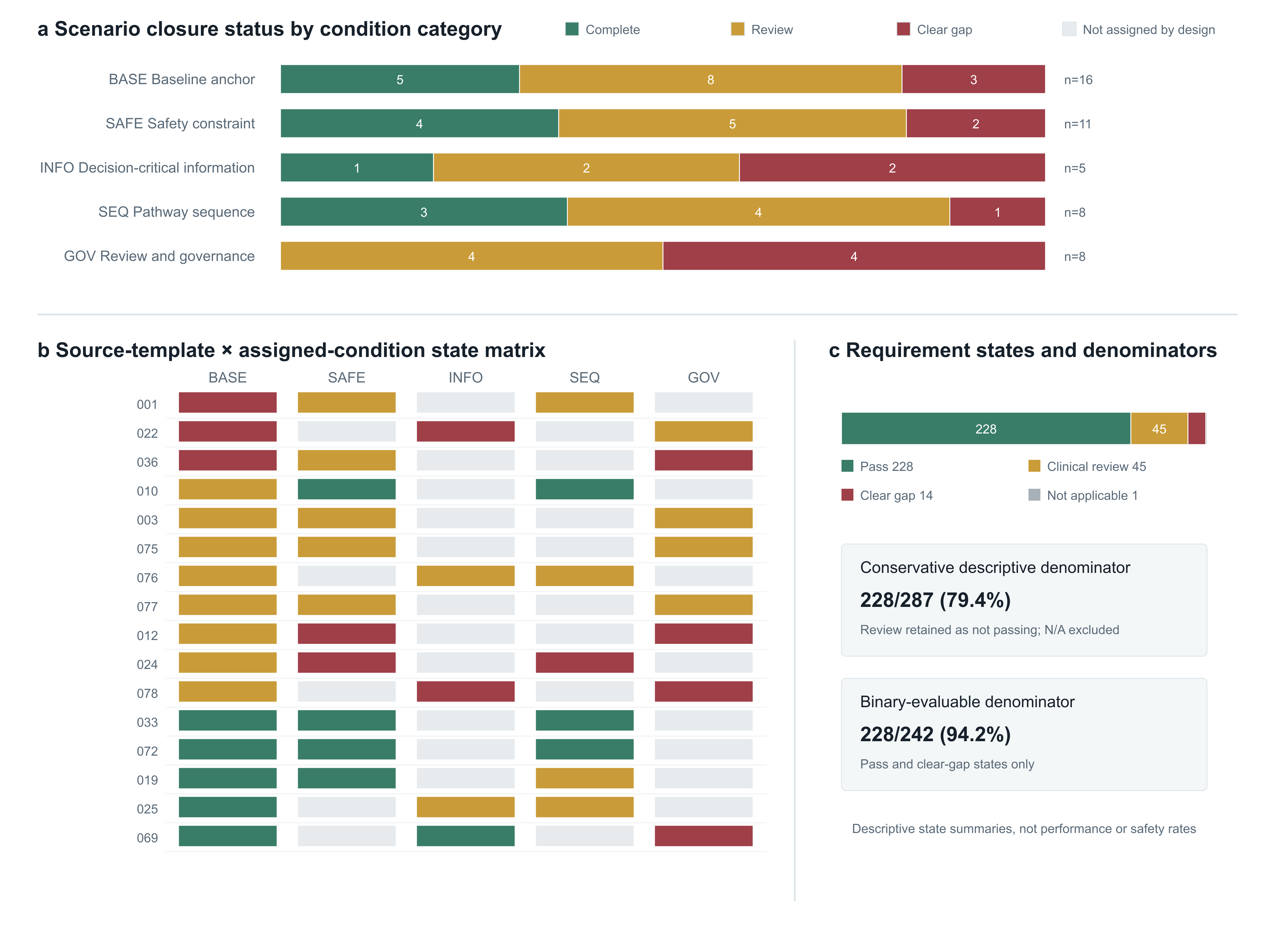}
\end{center}
{\small\noindent\textbf{Extended Data Figure 7 | Purposive case-variant states and requirement denominators} Scenario-state composition across the BASE baseline anchor, SAFE safety-constraint, INFO decision-critical-information, SEQ pathway-sequence and GOV reassessment-and-governance conditions for 16 source-case templates, together with a template-by-condition matrix ordered by BASE and modified-scenario states. Gray cells indicate conditions not assigned by design rather than missing data. The requirement-level panel shows the complete states assigned by the fixed AI-judge panel to 288 prespecified relationship requirements and the denominator composition for the conservative descriptive proportion, 228/287 (79.4\%), and the binary-evaluable proportion, 228/242 (94.2\%). These proportions describe states and denominators only and are not interpreted as overall MCE performance, error or clinical-safety rates.\par}

\clearpage
\phantomsection
\label{fig:extended-8}
\begin{center}
\centering
\includegraphics[width=\textwidth,height=0.68\textheight,keepaspectratio]{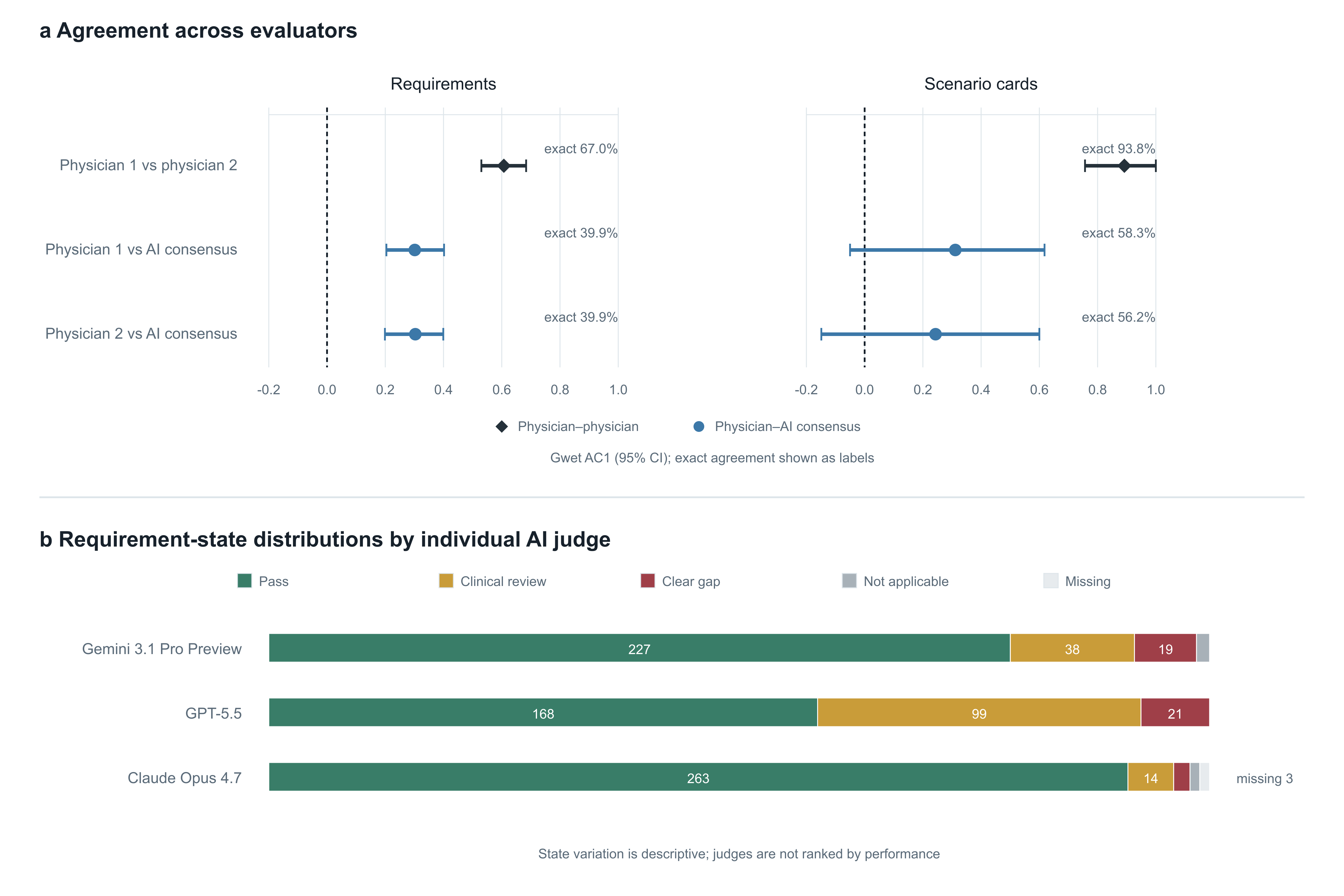}
\end{center}
{\small\noindent\textbf{Extended Data Figure 8 | Evaluator dependence in purposive case-variant judgments} Requirement-level and scenario-card-level Gwet AC1 estimates and 95\% CIs are shown for agreement between the two physicians and between each physician and the fixed AI consensus. Black diamonds denote physician--physician comparisons and blue circles physician--AI comparisons; exact agreement is labeled beside each estimate. The supporting panel gives requirement-level distributions of pass, clinical review, clear gap, not applicable and missing states across planned calls for each AI judge; judges are reported separately without performance ranking. Physician agreement was estimated from independent, pre-adjudication review of all 48 scenarios and 288 CCRRs. Requirement-level and scenario-card-level units were retained separately; physician--AI comparisons diagnose evaluator dependence and do not define either side as a reference standard. Requirement states by prespecified safety label are reported in Supplementary Table 8.\par}

\clearpage
\phantomsection
\label{fig:extended-9}
\begin{center}
\centering
\includegraphics[width=\textwidth,height=0.68\textheight,keepaspectratio]{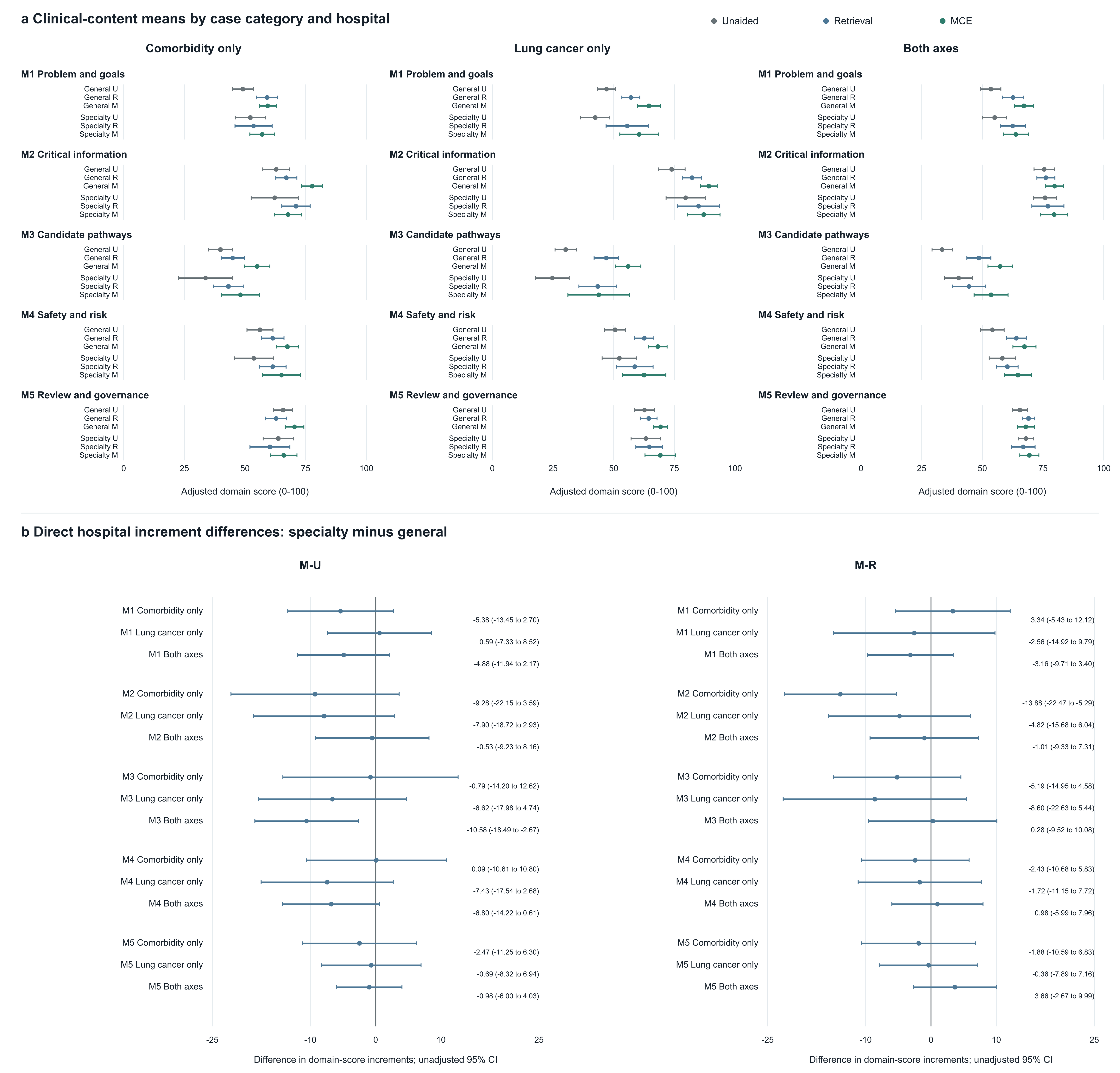}
\end{center}
{\small\noindent\textbf{Extended Data Figure 9 | Clinical-content patterns by case category and hospital background} \textbf{a.} Adjusted U, R and M means and unadjusted 95\% CIs for all five clinical-content domains in each of the three case categories and two hospital backgrounds. \textbf{b.} Direct differences in M-U and M-R increments, consistently specialty minus general, for all 15 domain-by-category combinations. The analysis used 2,133 responses from 237 physicians at 91 registered institutional codes. Models used physician random intercepts, condition-by-case and hospital/category contrasts, equal weighting of cases within category and task positions, common professional-title standardization and institutional-code CR2 inference. Comorbidity-complex only and lung-cancer-complex only mean only that axis was marked complex, not absence of the other disease. Categories are unordered; no neither-complex group was present. The main target retained 40 cases and one empty specialty-hospital R case cell without imputation; the common-support sensitivity used 39 cases. CIs are unadjusted; the complete 60-test within-category and 30-test direct-hospital families, ten principal heterogeneity tests and sensitivities are in Supplementary Table 9k. The displayed differences are exploratory content-score contrasts, not hospital capability rankings.\par}

\clearpage
\newgeometry{margin=15mm}
\noindent\textbf{Extended Data Figure 9 | Panel a: Comorbidity only}\par
\vspace{0.5em}
\begin{center}
\includegraphics[width=\textwidth,height=0.9\textheight,keepaspectratio]{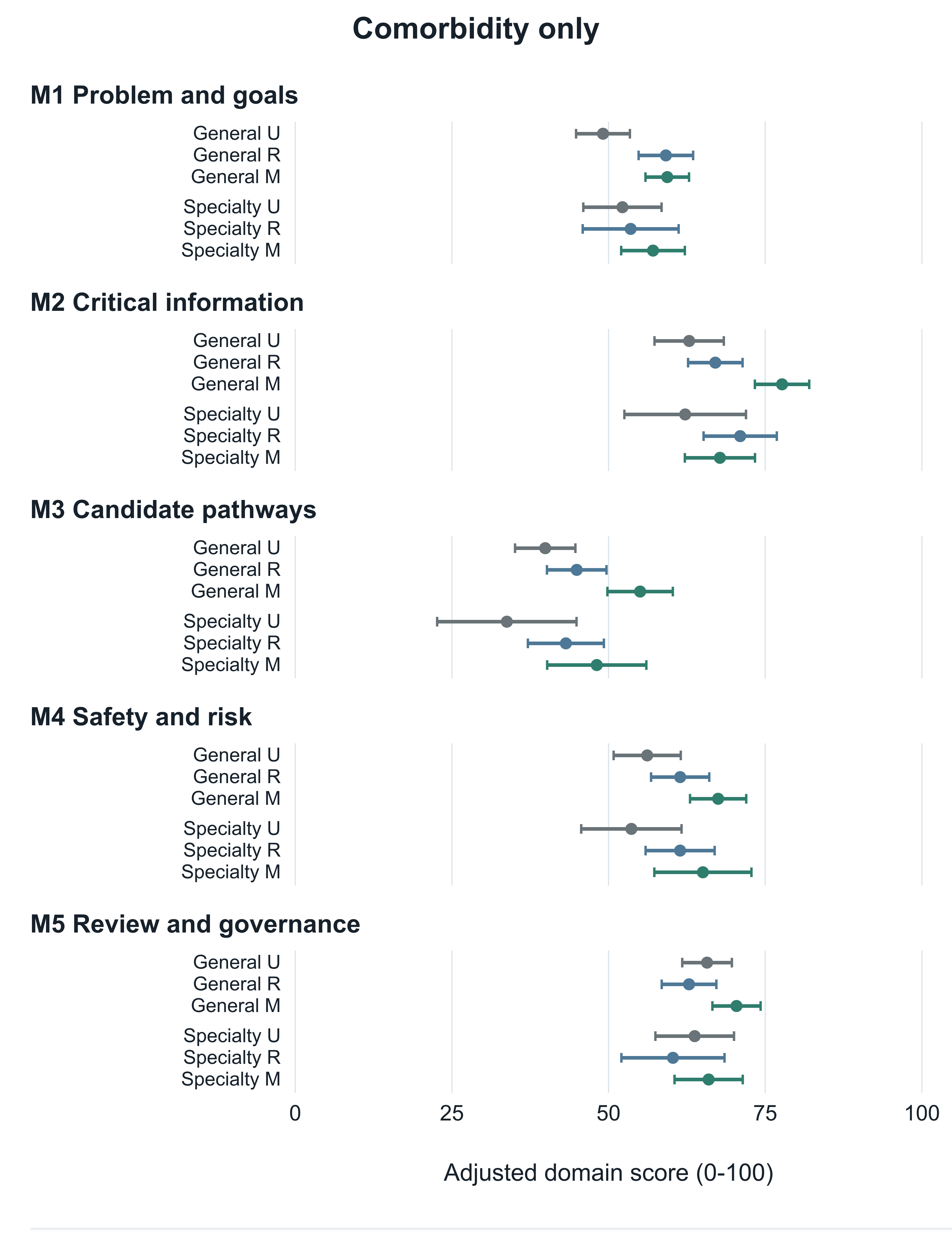}
\end{center}
\clearpage
\noindent\textbf{Extended Data Figure 9 | Panel a: Lung cancer only}\par
\vspace{0.5em}
\begin{center}
\includegraphics[width=\textwidth,height=0.9\textheight,keepaspectratio]{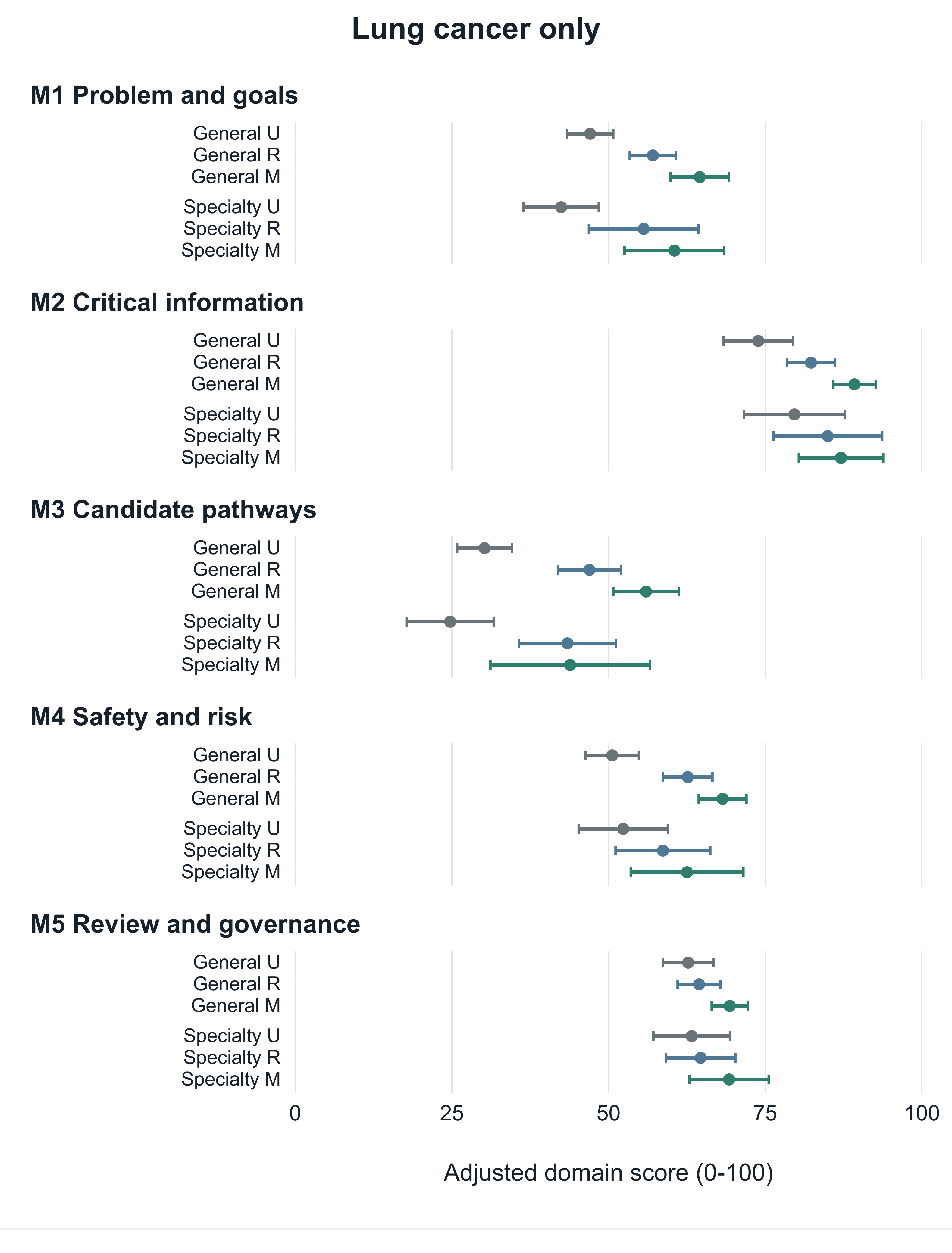}
\end{center}
\clearpage
\noindent\textbf{Extended Data Figure 9 | Panel a: Both axes}\par
\vspace{0.5em}
\begin{center}
\includegraphics[width=\textwidth,height=0.9\textheight,keepaspectratio]{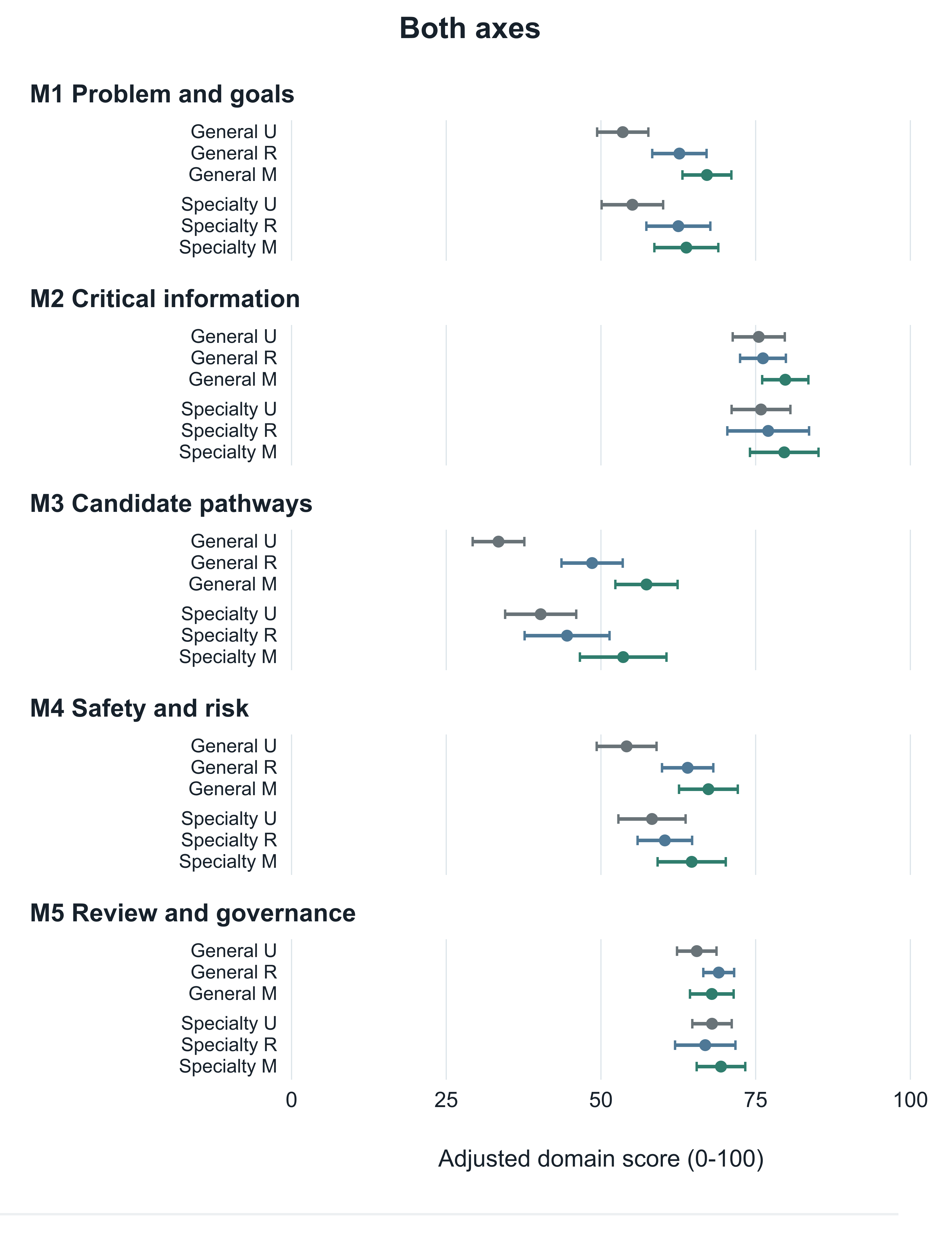}
\end{center}
\clearpage
\noindent\textbf{Extended Data Figure 9 | Panel b: M-U}\par
\vspace{0.5em}
\begin{center}
\includegraphics[width=\textwidth,height=0.9\textheight,keepaspectratio]{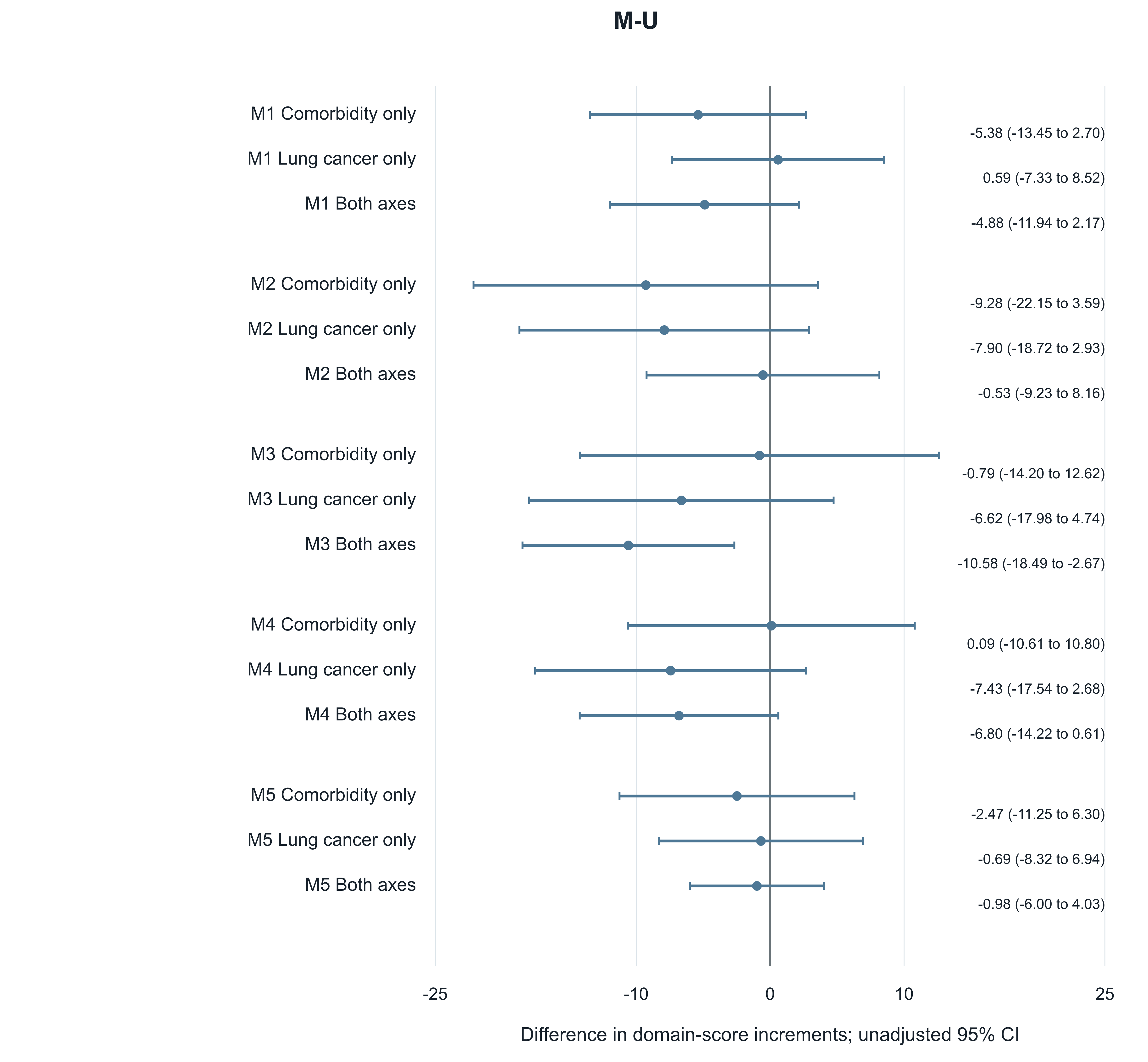}
\end{center}
\clearpage
\noindent\textbf{Extended Data Figure 9 | Panel b: M-R}\par
\vspace{0.5em}
\begin{center}
\includegraphics[width=\textwidth,height=0.9\textheight,keepaspectratio]{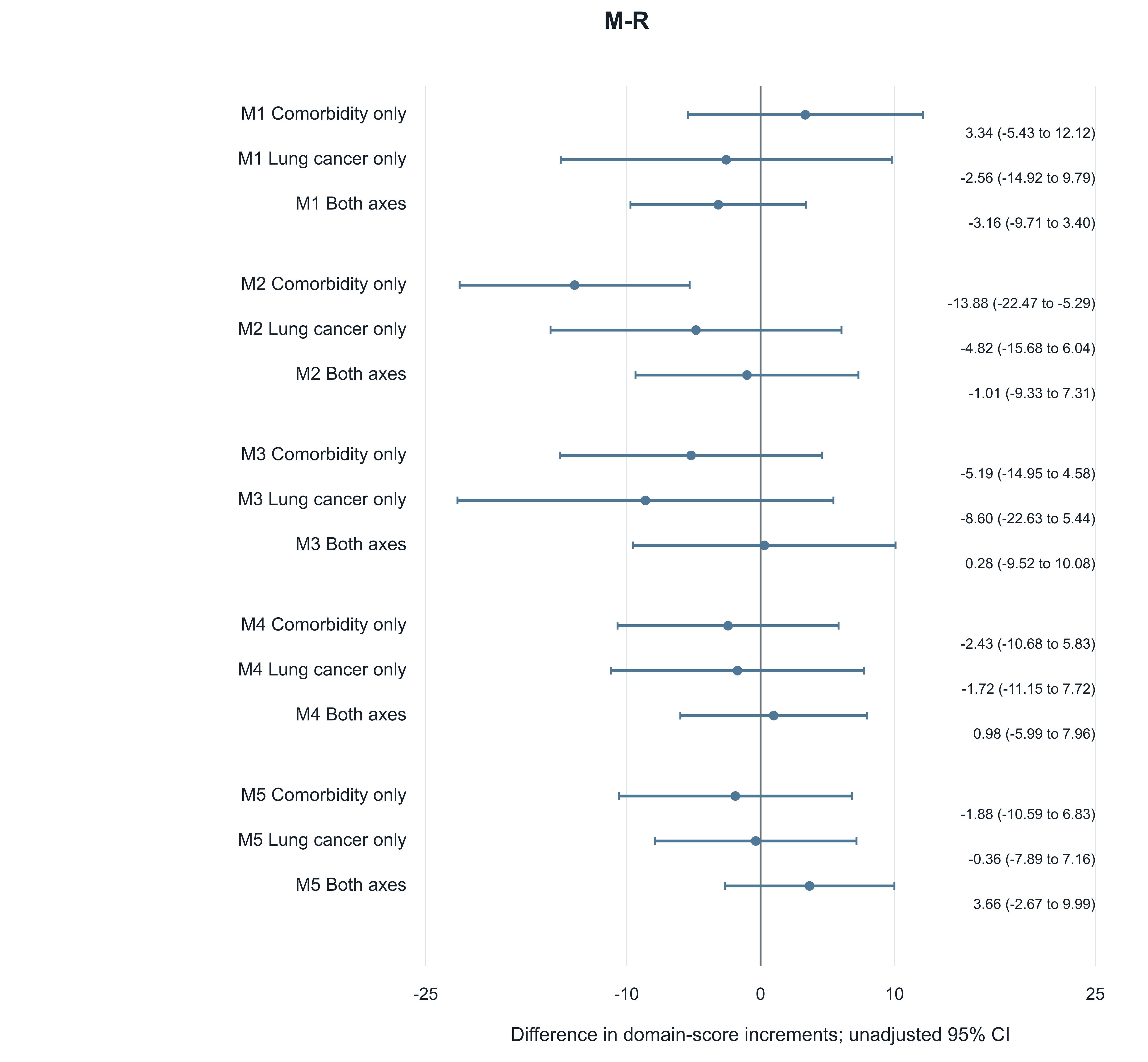}
\end{center}
\restoregeometry

\end{document}